\documentclass{article} 
\usepackage{iclr2026_conference,times}

\usepackage{amsmath,amsfonts,bm}

\newcommand{\reach}{\operatorname{Reach}}
\newcommand{\inv}{\operatorname{Inv}}

\def\eqref#1{equation~\ref{#1}}

\def\1{\bm{1}}

\DeclareMathAlphabet{\mathsfit}{\encodingdefault}{\sfdefault}{m}{sl}
\SetMathAlphabet{\mathsfit}{bold}{\encodingdefault}{\sfdefault}{bx}{n}

\usepackage{hyperref}
\usepackage{url}

\usepackage{bbm}
\usepackage{graphicx}
\usepackage{amsmath}
\usepackage{algorithm}
\usepackage{algpseudocode}

\usepackage{booktabs} 
\usepackage{amssymb}  

\usepackage{subcaption}

\usepackage[dvipsnames]{xcolor}
\usepackage{tikz}
\usetikzlibrary{shapes.geometric, arrows.meta, positioning, decorations.pathreplacing}
\tikzset{
    tableNode/.style={
        rectangle,           
        draw,                
        thick,               
        align=center,          
        inner sep=2pt,       
        rounded corners=4pt, 
        font=\small,         
        minimum width=2cm,
        minimum height=0.6cm,
    },
    lineNode/.style={
        align=center,          
        font=\small,         
    }
}
\newcommand{\colora}[1]{\textcolor{RoyalBlue}{\textbf{#1}}}
\newcommand{\colorb}[1]{\textcolor{BrickRed}{\textbf{#1}}}
\newcommand{\scolora}[1]{\textcolor{RoyalBlue}{\boldsymbol{#1}}}
\newcommand{\scolorb}[1]{\textcolor{BrickRed}{\boldsymbol{#1}}}

\title{TGPO: Temporal Grounded Policy\\ Optimization for Signal Temporal Logic Tasks}

\author{Yue Meng \quad\quad Fei Chen \quad\quad Chuchu Fan\\
Massachusetts Institute of Technology\\
\texttt{\{mengyue,feic,chuchu\}@mit.edu}
}

\newcommand{\partitle}[1]{\noindent{\textbf{#1.}}}

\iclrfinalcopy 
\begin{document}
\maketitle
\begin{abstract}
Learning control policies for complex, long-horizon tasks is a central challenge in robotics and autonomous systems. Signal Temporal Logic (STL) offers a powerful and expressive language for specifying such tasks, but its non-Markovian nature and inherent sparse reward make it difficult to be solved via standard Reinforcement Learning (RL) algorithms. Prior RL approaches focus only on limited STL fragments or use STL robustness scores as sparse terminal rewards. In this paper, we propose TGPO, Temporal Grounded Policy Optimization, to solve general STL tasks. TGPO decomposes STL into timed subgoals and invariant constraints and provides a hierarchical framework to tackle the problem. The high-level component of TGPO proposes concrete time allocations for these subgoals, and the low-level time-conditioned policy learns to achieve the sequenced subgoals using a dense, stage-wise reward signal. During inference, we sample various time allocations and select the most promising assignment for the policy network to rollout the solution trajectory. To foster efficient policy learning for complex STL with multiple subgoals, we leverage the learned critic to guide the high-level temporal search via Metropolis-Hastings sampling, focusing exploration on temporally feasible solutions. We conduct experiments on five environments, ranging from low-dimensional navigation to manipulation, drone, and quadrupedal locomotion. Under a wide range of STL tasks, TGPO significantly outperforms state-of-the-art baselines (especially for high-dimensional and long-horizon cases), with an average of 31.6\% improvement in task success rate compared to the best baseline. The code will be available at \url{https://github.com/mengyuest/TGPO}
\end{abstract}
\section{Introduction}
\label{sec:introduction}

Signal Temporal Logic (STL) is a powerful framework for specifying tasks with temporal and spatial constraints in real-world robotic applications. However, designing controllers to satisfy these specifications is difficult, especially for systems with complex dynamics and a long task horizon. While Reinforcement Learning (RL) excels in handling these dynamical systems, directly deploying RL for STL specifications poses significant challenges. The history-dependent nature of STL breaks the Markovian assumption for the common RL algorithms. Furthermore, the reward based on the STL satisfaction is extremely sparse for long-horizon tasks, making RL struggle to learn effectively. 

Existing model-free RL approaches for STL tasks typically leverage state augmentation with reward shaping. $\tau$-MDP~\citep{aksaray2016q} encodes histories explicitly in the augmented spaces and F-MDP~\citep{venkataraman2020tractable} designs flags to bookkeep the satisfaction of STL subformulas. However, these techniques only work on limited STL fragments with up to two temporal layers. While model-based RL~\citep{kapoor2020model,he2024scalable} has fewer restrictions on the STL formulas, learning the system (latent space) dynamics can be challenging, and the estimation error accumulates over long horizons. Additionally, the planning often relies on Monte Carlo Tree Search or sampling action sequences, which may not be tractable for high-dimensional systems.

We argue that the primary barrier for RL to efficiently solve STL tasks is the difficulty of designing a dense, stage-wise reward function. This challenge stems directly from the unspecified temporal variables governing the ``reach"-type tasks in STL formulas, which prevents a direct decomposition of STL into a sequence of executable subgoals. For example, for an STL $F_{[0,160]}A \land F_{[0,160]}B$ (``Eventually reach $A$ and eventually reach $B$ within the time interval $[0,160]$"), the time assignments for reaching $A$ and reaching $B$ determine the order of visiting these regions. If we can ground the variables into concrete values (e.g., reach $A$ at 35, and reach $B$ at 120), the problem can be cast into a sequence of goal-reaching problems, which is much easier to solve by RL.

Inspired by this observation, we propose a hierarchical RL framework to solve STL tasks by iteratively conducting \textbf{T}emporal \textbf{G}rounding and \textbf{P}olicy \textbf{O}ptimization (TGPO). The high-level component assigns values for the time variables to form the sequenced subgoals, and the low-level time-conditioned policy learns to achieve the task guided by the dense, stage-wise rewards derived from these subgoals. To efficiently bind values for multiple time variables, we carry out a high-level temporal search with a critic that predicts STL satisfaction. A Metropolis–Hastings sampling is used to guide exploration toward more ``promising" time allocations. During inference, we sample time variable assignments and evaluate them using the critic. The most promising schedule is then executed by the low-level policy to generate the final solution trajectory for the STL specification.

We conduct extensive experiments over five simulation environments, ranging from 2D linear dynamics to 29D Ant navigation tasks. Compared to other baselines, TGPO\textsuperscript{*} (with Bayesian time variable sampling) achieves the highest overall task success rate. The performance gains are significant, especially in high-dimensional systems and long-horizon tasks. Furthermore, our time-conditioned design offers key benefits: our critic offers interpretability by identifying promising temporal plans, and the policy can generate diverse, multi-modal behaviors to satisfy a single STL specification.

Our main contributions are summarized as follows: (1) \textbf{Hierarchical RL-STL framework}: To the best of our knowledge, we are the first to develop a hierarchical model-free RL algorithm capable of solving general, nested STL tasks over long horizons. (2) \textbf{Critic-guided Bayesian sampling}: We introduce a critic-guided temporal grounding mechanism that, together with STL decomposition, yields subgoals and invariant constraints. This mechanism constructs an augmented MDP with dense, stage-wise rewards and thus overcomes the sparse reward challenges that have hindered existing RL approaches. (3) \textbf{Interpretability}: By explicitly grounding subgoals and invariant constraints in the STL structure using critic-guided Bayesian sampling, our approach offers a more interpretable learning process, where progress can be directly traced to logical task components. (4) \textbf{Complex dynamics and reproducibility}: TGPO demonstrates strong performance over other baselines and fits for complex dynamics, which supports the effectiveness of the design. All the code (the algorithm, the simulations and STL tasks) will be open-sourced to advance STL planning.

\section{Related work}
\label{sec:related-work}

\subsection{Signal Temporal Logic tasks}
Signal Temporal Logic (STL) offers a powerful framework for specifying robotics tasks~\citep{donze2013signal}. Unlike Linear Temporal Logic (LTL), STL operates over continuous signals with time intervals and lacks an automaton representation, making it challenging to conduct planning ~\citep{finucane2010ltlmop}. Traditional approaches for STL include sampling-based methods~\citep{vasile2017sampling,karlsson2020sampling,linard2023real,sewlia2023cooperative}, Mixed-integer Programming~\citep{sun2022multi,kurtz2022mixed} and trajectory optimization~\citep{leung2023backpropagation}. More recently, learning-based methods emerged, such as differentiable policy learning~\citep{liu2021recurrent,liu2023learning,meng2023signal}, imitation learning~\citep{puranic2021learning,leung2022semi,meng2024diverse,mengtelograf}, and reinforcement learning (RL)~\citep{liao2020survey}. 

\subsection{Reinforcement learning for temporal logic tasks}
Temporal logic RL has been extensively studied in Linear Temporal Logic (LTL) and some Signal Temporal Logic (STL) fragments~\citep{liao2020survey}, where the key challenge is designing suitable rewards. For LTL, existing methods~\citep{sadigh2014learning,li2017reinforcement,hasanbeig2018logically,hasanbeig2020deep} typically convert the formula into Limit-Deterministic Büchi Automata (LDBA)~\citep{sickert2016limit} or reward machines~\citep{icarte2018using}, while LTL2Action~\citep{vaezipoor2021ltl2action} uses progression~\citep{bacchus2000using} to assign dense reward, and SpectRL~\citep{jothimurugan2019composable} devises a composable specification language for complex objectives. In contrast, STL poses additional challenges due to its explicit time constraints and real-value predicates. Early approaches augment the state space via temporal abstractions using history segments~\citep{aksaray2016q,ikemoto2022deep} or flags~\citep{venkataraman2020tractable,wang2024tractable}, while bounded horizon nominal robustness (BHNR)~\citep{balakrishnan2019structured} offers intermediate reward approximations. Recent work uses model-based learning to solve STL tasks with evolutionary strategies~\citep{kapoor2020model} and Monte-Carlo Tree Search in value function space~\citep{he2024scalable}. However, most of these methods are restricted to STL structures and systems (limited temporal nesting, fixed-size time windows, or grid-like environments). Instead, our method can handle more general STLs and efficiently designs augmented states along with dense, stage-wise rewards.

\section{Preliminaries}
\label{sec:preliminaries}

\subsection{Signal Temporal Logic (STL)}
\label{sec:prelim3-1}
Consider a discrete-time system $x_{t+1}=f(x_t,u_t)$ where $x_t\in\mathcal{X}\subseteq \mathbb{R}^n$ and $u_t\in\mathcal{U}\subseteq \mathbb{R}^m$ denote the state and control at time $t$. Starting from an initial state $x_0$, a signal $\sigma=x_0,...,x_T$ is generated via controls $u_0,...,u_{T-1}$. STL specifies properties via the following rules~\citep{donze2013efficient}:
\begin{equation}
    \phi::= \top \ | \ \mu(x)\geq 0 \ | \ \neg \phi \ | \ \phi_1 \land \phi_2 \ | \ \phi_1 U_{[a,b]}\phi_2
    \label{eq:stl_formula}.
\end{equation}
Here the boolean-type operators split by ``$|$" are the building blocks to compose an STL: $\top$ means ``true", $\mu$ denotes a predicate function $\mathbb{R}^n\to\mathbb{R}$, and $\neg$, $\land$, $U$, ${}_{[a,b]}$ are ``negation", ``conjuction", ``until'' and the time interval from $a$ to $b$. 
Other operators are ``disjunction": $\phi_1\lor\phi_2=\neg(\neg\phi_1 \land \neg\phi_2)$,  ``eventually": $F_{[a,b]} \phi=\top U_{[a,b]}\phi$ and ``always": $G_{[a,b]} \phi = \neg F_{[a,b]}\neg\phi$. We denote $\sigma,t\models \phi$ if the signal $\sigma$ from time $t$   satisfies the STL formula (the evaluation of $\phi$ returns True). In particular, we simply write $\sigma\models \phi$ if the signal is evaluated from $t=0$.
For operators $\top, \mu\geq 0, \neg, \land$ and $\lor$, the evaluation checks for the signal state at time $t$. As for temporal operators~\citep{maler2004monitoring}: $\sigma,t \models F_{[a,b]}\phi  \Leftrightarrow \,\, \exists t' \in [t+a, t+b],\, \sigma,t'\models \phi$; and $\sigma,t \models G_{[a,b]}\phi  \Leftrightarrow \,\, \forall t' \in [t+a, t+b],\, \sigma,t'\models \phi$; and $\sigma,t\models \phi_1 U_{[a,b]} \phi_2 \Leftrightarrow \exists t'\in [t+a,t+b], \sigma,t'\models \phi_2, \forall t''\in[0,t'], \sigma, t''\models \phi_1$. In plain words, $\phi_1 U_{[a,b]} \phi_2$ means ``$\phi_1$ holds until $\phi_2$ happens in $[a, b]$." Robustness score~\citep{donze2010robust} $\rho(\sigma,t,\phi)$ measures how well a signal $\sigma$ satisfies $\phi$. We have $\rho\geq 0 \text{ iff } \sigma,t\models \phi$. The score $\rho$ is:
\begin{equation}
    \begin{cases}
        & \rho(\sigma,t, \top)= 1,\quad\quad\quad  \rho(\sigma,t, \mu)= \mu(\sigma(t)),\quad\quad\quad \rho(\sigma,t, \neg \phi)= -\rho(\sigma,t,\phi),\\
        & \rho(\sigma,t, \phi_1 \land \phi_2)= \min\{\rho(\sigma,t,\phi_1),\rho(\sigma,t,\phi_2) \},\\
        & \rho(\sigma,t, F_{[a,b]} \phi)= \sup\limits_{r\in[a,b]}\rho(\sigma,t+r,\phi),
        \quad\quad  \rho(\sigma,t, G_{[a,b]} \phi)= \inf\limits_{r\in[a,b]}\rho(\sigma,t+r,\phi),\\
        & \rho(\sigma,t, \phi_1 U_{[a,b]} \phi_2)= \sup\limits_{t'\in[t+a,t+b]}\min\left\{\rho(\sigma,t',\phi_2), \inf\limits_{t''\in[t,t']}\rho(\sigma,t'',\phi_1) \right\}.
    \end{cases}
    \label{eq:robustness_score}
\end{equation}

\subsection{Markov Decision Process}
A Markov Decision Process (MDP) is defined by the tuple $\mathcal{M}=(\mathcal{S}, \mathcal{A}, P, R, \gamma)$ where: $\mathcal{S}$ and $\mathcal{A}$ represent the sets of states and actions, respectively, $P:\mathcal{S}\times \mathcal{A}\times \mathcal{S}\to[0,1]$ is the probabilistic transition function where $P(s'|s,a)$ denotes the probability of the next state $s'$ given current state $s$ and action $a$, $R:\mathcal{S}\times \mathcal{A}\to \mathbb{R}$ is the reward function, and $\gamma\in [0,1)$ is the discount factor. The agent decision is made by a policy $\pi:\mathcal{S}\to \mathcal{A}$ which maps states to a probability distribution over actions. The objective is to find an optimal policy $\pi^*$ that maximizes the expected discounted cumulative reward from a starting state $s_0$: $\mathop{\mathbb{E}}_{\pi}\left[ \sum\limits_{t=0}^\infty\gamma^t R(s_t, a_t) \big| s_0 \right]$ with $a_t\sim \pi(\cdot|s_t)$ and $s_{t+1}\sim P(\cdot |s_t, a_t)$.

\subsection{Problem formulation}
Consider a discrete-time system with state space $\mathcal{X}$, control space $\mathcal{U}$ and the initial state set $\mathcal{X}_0$. Given an STL formula $\phi$ defined in Eq.~\ref{eq:stl_formula}, our objective is to first formulate an MDP $(\mathcal{S}, \mathcal{A}, P, R, \gamma)$ and then learn a policy: $\pi: \mathcal{S}\to\mathcal{A}$ to maximize the satisfaction probability, $\max\limits_{\pi}\mathop{\mathbb{P}}\limits_{x_0\in \mathcal{X}_0}(\sigma\models\phi)$.

\partitle{Remarks}
It is tempting to treat the control system state $\mathcal{X}$ as the MDP state $\mathcal{S}$, and the control input $\mathcal{U}$ as the actions $\mathcal{A}$. However, for STL tasks, the policy also depends on the history\footnote{E.g., if an STL task is to ``Eventually reach region A and then reach B", the policy needs to ``remember" whether it has already visited the region A in order to proceed to reach B.}, making the problem non-Markovian. Thus, we need to augment the state to keep history data. Besides, the satisfaction of an STL is checked over the full trajectory, making it difficult to define dense rewards (unlike LTL, where stage-wise rewards~\citep{camacho2017decision,vaezipoor2021ltl2action} can be defined). Thus, we need to design dense rewards under the augmented state space to learn efficiently.

\section{Methodology}
\label{sec:methodology}
\begin{figure}[htbp]
    \centering
    \includegraphics[width=0.99\textwidth]{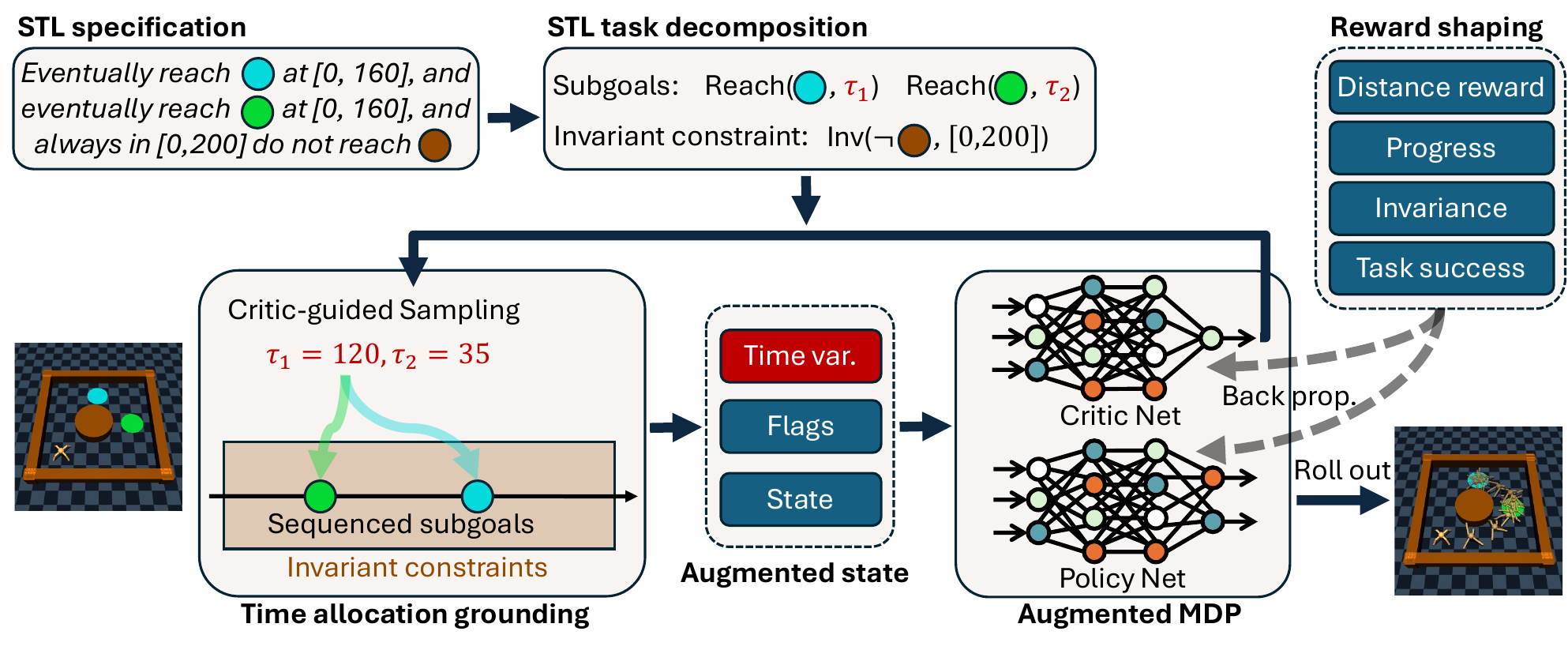}
    \caption{Framework: STL decomposition and critic-guided temporal grounding yield subgoals and invariant constraints that guide an augmented MDP with dense rewards for policy optimization.}
    \label{fig:framework-curve}
\end{figure}

We propose TGPO, Temporal Grounded Policy Optimization, to address the problem considered. The entire framework is illustrated in Fig.~\ref{fig:framework-curve}, and we explain each component in detail below.

\subsection{STL subgoal decomposition}
Our method of decomposing STL into subgoals with invariant constraints is inspired by~\citet{kapoor2024safe,liu2025zero}. The essence is to first translate the STL into a set of subtasks, where each subtask has a checker $\mu$ on the trace $\sigma$ and belongs to one of the following types:
\begin{itemize}
    \item \textbf{Reachability task}: achieve $\mu(\sigma(\tau))\geq 0$ at a time instant $\tau$, denoted as $\reach(\mu,\tau)$.
    \item \textbf{Invariance task}: keep $\mu(\sigma(\tau))\geq 0$ for all time $\tau$ in an interval $W$, denoted as $\inv(\mu, W)$.
\end{itemize}

\begin{figure}[!ht]
\centering
\begin{tikzpicture}[node distance=0.0cm and 0.0cm]
    \node[tableNode] (root) {
        $\text{Reach}(\phi,0)$
    };

    \node[tableNode, on grid, below left=0.9cm and 7.45cm of root] (side) {
        $\textbf{Time variables } \mathbf{t}=(\scolora{\tau}, \scolorb{\tau'})$ \\
        \setlength{\tabcolsep}{1.5pt}
            \begin{tabular}{c|c|c|c}
                 Task & AP & Starting time $\underline{t}$ & Ending time $\bar{t}$ \\
                 \hline
                 \textbf{Subgoal 1} & $\mu_1$ & $\scolora{\tau}$ & $\scolora{\tau}$ \\
                 \hline
                 \textbf{Subgoal 2} & $\mu_2$ & $\scolora{\tau}+a_2$ & $\scolora{\tau}+b_2$ \\
                 \hline
                 \textbf{Subgoal 3} & $\mu_3$ & $\scolora{\tau}+\scolorb{\tau'}$ & $\scolora{\tau}+\scolorb{\tau'}$\\
                 \hline
                 \textbf{Invariant} & $\neg\mu_0$  & $c$  & $d$\\
            \end{tabular}
    };

    \node[tableNode, on grid, below left=1.3cm and 2cm of root] (childA) {
        $\text{Reach}(\phi_0, \scolora{\tau})$
    };
    
    \node[tableNode, fill=yellow!15, on grid, below right=1.2cm and 1.8cm of root] (childB) {
        $\text{Inv}(\neg\mu_0,[c,d])$\\
        \textbf{Invariant constraint}
    };
    
    \node[tableNode, fill=yellow!15, on grid, below right=1.3cm and -2.9cm of childA] (grandchildA1) {
        $\text{Reach}(\mu_1, \scolora{\tau})$\\
        \textbf{Subgoal 1}
    };

    \node[tableNode, fill=yellow!15, on grid, below right=1.3cm and 0cm of childA] (grandchildA2) {
        $\text{Inv}(\mu_2, [\scolora{\tau}+a_2, \scolora{\tau}+b_2])$\\
        \textbf{Subgoal 2}
    };

    \node[tableNode, fill=yellow!15, on grid, below right=1.3cm and 3.2cm of childA] (grandchildA3) {
        $\text{Reach}(\mu_3, \scolora{\tau}+\scolorb{\tau'})$\\
        \textbf{Subgoal 3}
    };

    \begin{scope}[-Latex, thick] 
        \draw (root) -- node[lineNode,xshift=-46pt, yshift=5pt]{\colora{\text{Time variable}} $\scolora{\tau\in [a,b]}$} (childA);
        \draw (root) -- (childB);
        \draw (childA) -- (grandchildA1);
        \draw (childA) -- (grandchildA2);
        \draw (childA) --node[lineNode,xshift=55pt, yshift=2pt]{\colorb{\text{Time variable}} $\scolorb{\tau'\in [a_3,b_3]}$} (grandchildA3);
    \end{scope}

\end{tikzpicture}
\caption{STL decomposition of $\phi=F_{[a,b]}(\mu_1 \land G_{[a_2,b_2]}\mu_2\land F_{[a_3,b_3]}\mu_3) \land G_{[c,d]}\neg \mu_0$.}
\label{fig:tikz-stl-decompose}
\end{figure}
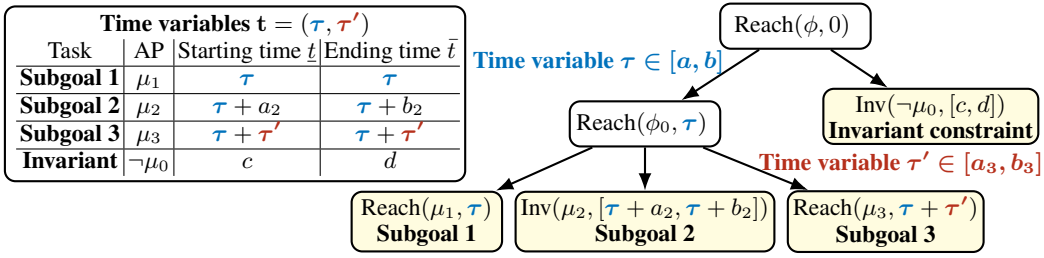

For basic STL formulas, the time instants and the time intervals can be concrete values or variables: e.g., the formula $G_{[a,b]}\mu$ can be written as $\inv(\mu,[a,b])$ with concrete $[a,b]$, whereas the formula $F_{[a,b]}\mu$ can be written as $\reach(\mu,\tau)$ with the time variable $\tau\in [a,b]$, 
and $\mu_1 U_{[a,b]} \mu_2$ can be written as $\{\reach(\mu_2, \tau), \inv(\mu_1, [a,\tau])\}$ with the time variable $\tau \in [a,b]$. 
For a nested STL, we follow a top-down approach to ``flatten" it into reachability and invariance tasks governed by time variables. We denote $\reach(\phi,\tau)$ for $\rho(\sigma,\tau,\phi)\geq 0$ and use $\inv(\phi,W)$ to represent $\rho(\sigma,\tau,\phi)\geq0, \forall \tau \in W$. 
For any STL $\phi$ we can write it as $\reach(\phi,0)$ and then we rewrite with tasks using its subformulas. The subformula will always carry time variables from its ancestor operators, and we repeat the process until all the tasks are represented as atomic propositions (APs) corresponding to $\mu$ or its negation $\neg\mu$. For example, for $\phi=F_{[a,b]}\phi_0 \land G_{[c,d]}\neg \mu_0$ where $\phi_0=\mu_1 \land G_{[a_2,b_2]}\mu_2\land F_{[a_3,b_3]}\mu_3$ is a subformula, we can represent $\phi$ as $\{\reach(\phi_0,\tau), \inv(\neg\mu_0, [c,d])\}$ with domain $\{ \tau\in[a,b]\}$, then we can pass $\tau$ into $\phi_0$ to represent the STL as $\{\reach(\mu_1, \tau), \inv(\mu_2, [\tau+a_2,\tau+b_2]), \reach(\mu_3, \tau+\tau'), \inv(\neg\mu_0,[c,d])\}$ with domains $\{\tau\in[a,b],\tau'\in[a_3,b_3]\}$. An illustration of the decomposition is depicted in Fig. \ref{fig:tikz-stl-decompose}. 
In this work, we do not consider disjunctions or temporal structures of the form ``$G(F\ldots)$." Such STLs can be represented by introducing additional binary variables to select the disjunction branch and more time variables for each instant in the time domain of the $G$ operator.

From the reachability and invariance tasks, we further denote \textbf{subgoals} (reach or stay) as tasks that are either a reachability task (e.g., \textbf{Subgoal 1} in Fig. \ref{fig:tikz-stl-decompose}) or an invariance task (e.g., \textbf{Subgoal 2} in Fig. \ref{fig:tikz-stl-decompose}) with atomic proposition $\mu$ (we assume all the APs are for reaching certain regions). The remaining invariance tasks associated with negation of APs (e.g., $\inv(\neg\mu_0,[c,d])$) are treated as \textbf{invariant constraints} (avoidance).  
Through this decomposition, a complex STL formula $\phi$ reduces to $N_g$ subgoals $\phi^g_{i}$ with $\reach (\mu^g_{i},\tau_i)$ or $\inv (\mu^g_{i},W_i)$, $i\in \Theta_g:=\{1,\cdots,N_g\}$ and $N_c$ invariant constraints $\phi^c_{j}$ with $\inv (\neg\mu^c_{j},W_j)$, $j\in \Theta_c:=\{1,\cdots,N_c\}$. Each subgoal / constraint has a starting time and an ending time $[\underline{t}, \bar{t}]$ which is $[\tau,\tau]$ (or $W$). Denote all the time variables in this STL as $\mathbf{t}$. Next, we will show how this decomposition guides our state augmentation and reward shaping.

\subsection{Temporal grounded state augmentation and reward design} 
Given a concrete time variables assignment $\mathbf{t}$, the problem is now structured as reaching a sequence of subgoals sorted by their starting time with invariant constraints satisfied during execution. For brevity, we assume the subgoal indices are already sorted. We augment our state as: 
\begin{equation}
    s=(x,\tau, p_{prev}, p, r, \chi)
\end{equation}
Here $x\in\mathbb{R}^{n}$ stands for the original state, $\tau\in \{0,1,\cdots,T\}$ represents the time index, $p\in \{0,1,\cdots,N_g\}$ represents the progress index and $p_{prev}$ records the previous progress, $r$ records the certificate to proceed to the next subgoal, $\chi\in \{0,1\}^{N_c}$ maintains the satisfaction status for the invariant constraints. For the $k$-th subgoal (or invariant constraint), denote the starting time $\underline{t}^g_k$ (or $\underline{t}^c_k$) and the ending time $\bar{t}^g_k$ (or $\bar{t}^c_k$). The augmented state transition can be written as:
\begin{equation}
\begin{cases}
    x'=f(x, u), \quad
    \tau' = \tau+1, \quad
    p_{prev}'=p, \quad
    r'=h(r,x',\tau',p'), \quad
    p' = p + \mathbbm{1}(r'=2) \\
    \chi'_{k} = \chi_{k} \times \mathbbm{1}(\neg (\underline{t}^c_{k}\leq \tau' \leq \bar{t}^c_{k} \land \neg \mu^c_{k}(x')< 0))\quad k=0,1,...,N_c \\
\end{cases}
\end{equation}

where:
\begin{equation}
h(r,x',\tau',p')=\begin{cases}
0, \quad \text{if } r=2\\
1, \quad \text{if } \underline{t}^g_{{p'}} \neq \bar{t}^g_{{p'}} \land \tau'=\underline{t}^g_{{p'}} \land \mu^g_{{p'}}(x')\geq 0 \\
2, \quad \text{if } (r=1 \lor \underline{t}^g_{{p'}}=\bar{t}^g_{{p'}}) \land (\tau'=\bar{t}^g_{{p'}} \land \mu^g_{{p'}}(x')\geq 0)\\
r, \quad \text{otherwise}\\
\end{cases}
\end{equation}
The variable $r$ acts as a certificate (or flag) that keeps track of whether the reach-and-stay ($FG$) condition has been satisfied. It encodes the progress toward establishing that the predicate holds both at the entry time and the exit time of the required interval. To guide the agent to achieve these subgoals in a proper time window while satisfying the invariant constraints, we design the reward:
\begin{equation}
    R(s)=\lambda_1 R_{dist} + \lambda_2 R_{progress} + \lambda_3 R_{success} + \lambda_4 R_{inv}
\end{equation}
where $R_{dist}=\mu^g_{p}(x)$ is a distance-based reward shaping to encourage the agent to reach the current subgoal (and stay at the current subgoal within the time window $[\underline{t}_p^g, \bar{t}_p^g]$), $R_{progress} = \mathbbm{1}(p^{prev}\neq p)$ encourages the agent to achieve more subgoals, $R_{success} = \mathbbm{1}(p=N_g \land \chi=\mathbf{1})$ encourages the agent to finish all subgoals without violating any invariant constraints, and $\quad R_{inv} = \mathbbm{1}(\chi_k=0)$ penalizes for violating invariant constraints. The robustness score is also used at the final time step to encourage the agent to satisfy the STL. In this way, the agent is incentivized to reach all the subgoals while obeying the invariant constraints. We use Proximal Policy Optimization (PPO)~\citep{schulman2017proximal} to train the agent. The policy network and the critic receive the augmented state and the time variable assignment as the input, and output the action and the critic value correspondingly. At the beginning of each training epoch, we sample the time variables and collect episodes to update the network parameters. During inference, we sample time variables and use the trained critic to find the most effective assignment. The most naive way to sample these time variables will be randomly sampling from their feasible intervals, but we will present a better solution in the following section.

\subsection{Critic-guided Bayesian Time Allocation}
The key challenge in our framework is efficiently searching for time variable assignments. A naive uniform sampling strategy might waste huge effort on assignments that lead to infeasible or low-reward trajectories. To address this, we propose a Bayesian sampling strategy to find promising time assignments. We do not need to learn an extra surrogate function, as the value function learned by the PPO agent already provides a powerful heuristic. We employ a Metropolis-Hastings (MH) algorithm to sample time variables from $\exp(V_\psi(s_0,\mathbf{t}))$ for initial state $s_0$. The MH performs a guided random walk over the discrete time variable space and prefers to stay in regions that yield high critic values. To mitigate the risk of the sampler converging to a local optima and the fact that the initial critic might not be accurate, we adopt a hybrid approach: In each epoch, we use an MH sampler to obtain a ratio $\eta_{\text{mcmc}}$ of the time variables and sample a ratio $\eta_{\text{uniform}}$ through uniform sampling. To further leverage knowledge across training epochs, we maintain a replay buffer containing the top $\eta_{\text{elite}}$ ratio of “elite” time variable assignments that yield the highest STL robustness scores. This combination creates a robust and efficient mechanism for discovering effective temporal plans. The full training procedure is detailed in Algo.~\ref{alg:tgpo_hybrid}, and the ablation study for each component is shown in Sec.~\ref{sec:exp-abl}.

\begin{algorithm}[H]
\caption{TGPO with Hybrid Time Variable Sampling}
\label{alg:tgpo_hybrid}
\begin{algorithmic}[1]
\State \textbf{Input:} STL formula $\phi$ (subgoals and invariant constraints), elite buffer size $K$, batch size $N_{B}$
\State Initialize policy $\pi_\theta(a|s, \mathbf{t})$, critic $V_\psi(s, \mathbf{t})$, and elite time variable buffer $\mathcal{B}$
\For{iteration $i = 1, \dots, N$}
    \State \Comment{\textit{1. High-level Temporal Grounding}}
    \State $\mathbf{T}_{\text{uniform}} \gets$ Sample $\eta_{\text{uniform}}N_{B}$ time variables uniformly from the valid domain $\mathcal{T}$
    \State $\mathbf{T}_{\text{mcmc}} \gets$ Run Metropolis-Hastings guided by $V_\psi$ to generate $\eta_{mcmc} N_{B}$ time variables.  
    \State $\mathbf{T}_{\text{elite}} \gets$ Top $\eta_{elite} N_{B}$ time variables from elite buffer $\mathcal{B}$
    \State $\mathbf{T}_{\text{batch}} \gets \mathbf{T}_{\text{uniform}} \cup \mathbf{T}_{\text{mcmc}} \cup \mathbf{T}_{\text{elite}}$
    \State \Comment{\textit{2. Low-level Policy Optimization}}
    \State Collect trajectories $\mathcal{D}_i = \{(\sigma_j, \rho^\phi_j, \mathbf{t}_j)\}$ by executing $\pi_\theta$ with time variables from $\mathbf{T}_{\text{batch}}$
    \State Update $\pi_\theta$ and $V_\psi$ using the PPO algorithm on $\mathcal{D}_i$
    \State Update $\mathcal{B}$ with time variables from $\mathcal{D}_i\cup \mathcal{B}$ corresponding to top-$K$ STL robustness score
\EndFor
\State \Return Trained policy $\pi_\theta$, critic $V_\psi$, and elite buffer $\mathcal{B}$.
\end{algorithmic}
\end{algorithm}


\section{Experiments}
\label{sec:experiments}

\subsection{Implementation details}
\partitle{Baselines} We consider the following approaches. \textbf{RNN}: Train RL with a recurrent neural network (RNN) to handle history and use the STL robustness score as the rewards. \textbf{CEM}: Cross-Entropy Method~\citep{de2005tutorial} that optimizes the policy network with the STL robustness score as the fitness score. \textbf{Grad}: A gradient-based method~\citep{meng2023signal} that trains the policy with a differentiable STL robustness score. \textbf{$\tau$-MDP}: An RL method~\citep{aksaray2016q} which augments the state space with a trajectory segment to handle history data.
\textbf{F-MDP}: An RL approach~\citep{venkataraman2020tractable} that augments the state space with flags. We denote our base algorithm as \textbf{TGPO} and the enhanced version with Bayesian time sampling as \textbf{TGPO\textsuperscript{*}}.

\begin{figure}[!htbp]
    \centering
    \begin{subfigure}[b]{0.19\textwidth}
        \centering
        \includegraphics[width=\textwidth]{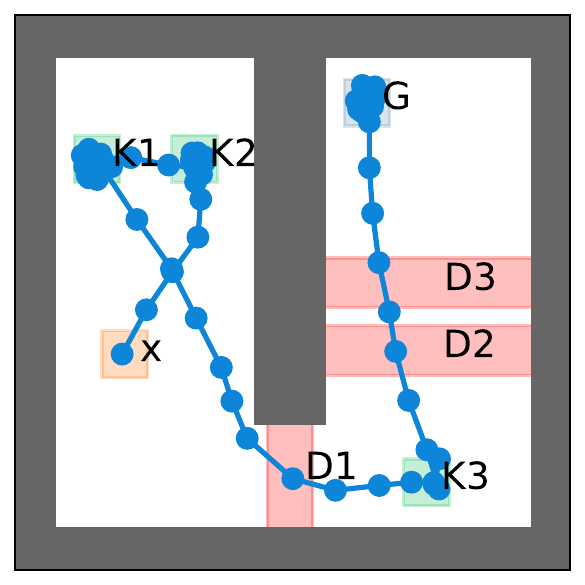}
        \caption{Linear}
        \label{fig:linear}
    \end{subfigure}
    \begin{subfigure}[b]{0.19\textwidth}
        \centering
        \includegraphics[width=\textwidth]{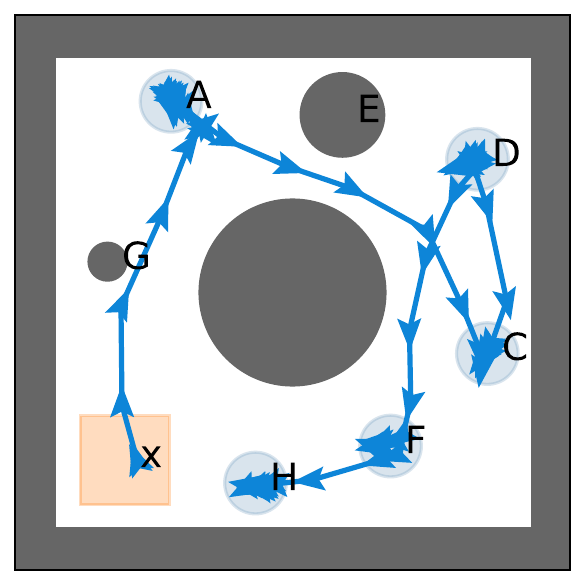}
        \caption{Unicycle}
        \label{fig:unicycle}
    \end{subfigure}
    \begin{subfigure}[b]{0.19\textwidth}
        \centering
        \includegraphics[width=\textwidth]{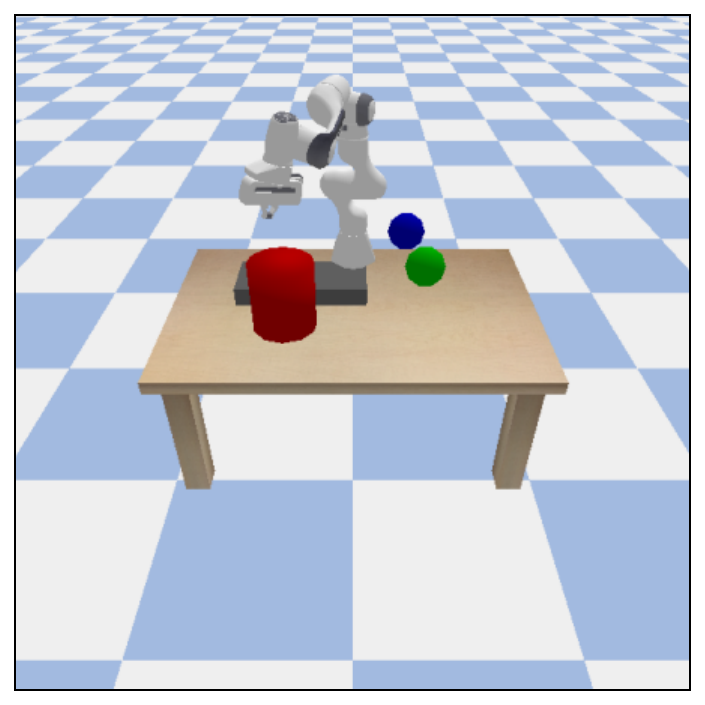}
        \caption{Franka Panda}
        \label{fig:panda}
    \end{subfigure}
    \begin{subfigure}[b]{0.19\textwidth}
        \centering
        \includegraphics[width=\textwidth]{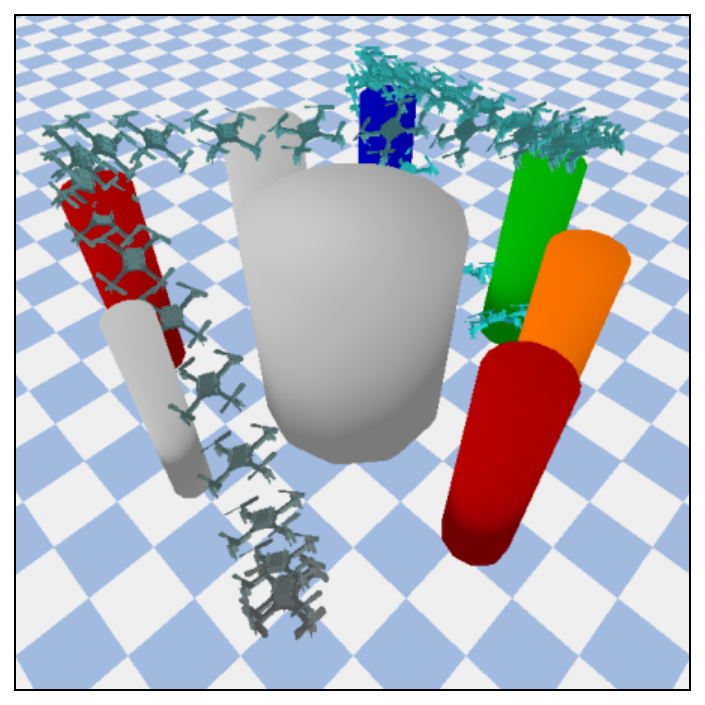}
        \caption{Quadrotor}
        \label{fig:drone}
    \end{subfigure}
    \begin{subfigure}[b]{0.19\textwidth}
        \centering
        \includegraphics[width=\textwidth]{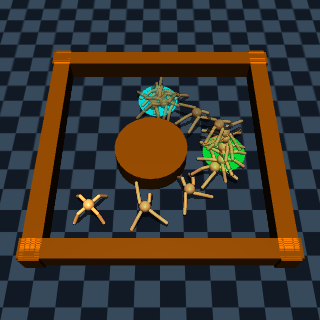}
        \caption{Ant}
        \label{fig:ant}
    \end{subfigure}
    \caption{Simulation benchmarks.}
    \label{fig:benchmarks}
\end{figure}

\partitle{Benchmarks} We evaluate TGPO across five environments shown in Fig.~\ref{fig:benchmarks} with varying dynamics and dimensionality:
(1) \textbf{Linear}: A 2D point-mass linear system.
(2) \textbf{Unicycle}: A non-holonomic 4D system for a wheeled robot.
(3) \textbf{Franka Panda}: A 7-DoF robot arm.
(4) \textbf{Quadrotor}: A 12D, full dynamic model of a quadrotor.
(5) \textbf{Ant}: a 29D quadruped robot for locomotion tasks. The agent starts from an initial set, and we specify the regions that the agent needs to reach, stay, or avoid using STL. For each benchmark, we designed 10 STL tasks of varying difficulty. Five of these STLs are two-layered (e.g., $F_{[0,T]}G_{[0,5]}(\text{Reach A})$), solvable by all the methods. The rest are multi-layer STLs with deeper nesting, which cannot be solved by F-MDP. Details can be found in App.~\ref{appendix:stls}.

\partitle{Training and evaluation}
For the main comparisons, the task horizon is fixed at $T$=100 except for ``Ant" ($T$=200). We trained each model with 7 random seeds to ensure statistical significance. All the methods are implemented in JAX~\citep{bradbury2018jax} and trained with 512 parallel environments for 1000$\sim$4000 epochs. All experiments were conducted on Amazon Web Services (AWS) g6e.2xlarge instances. A single experiment (a specific set of environment, method, STL, and random seed) took 5 to 90 minutes, depending on the environment and method complexity.
In the testing stage, we sample $512$ initial states. For each initial state, each baseline is given 10 attempts to generate the solution, and the trajectory with the highest STL robustness score is selected. For our approach, we attempt to select the best time assignment only once, based on the critic value, and then roll out the trajectory (we avoid the use of the STL score as feedback to choose the trajectory). The \textbf{Success rate} is the average performance over all the initial states and the STLs. We also measured \textbf{Training time}, as shown in App.~\ref{appendix:training-time}, which is the time to train each model (averaged over STLs).

\subsection{Main results}
\begin{figure}[!htbp]
    \centering
    \includegraphics[width=0.99\textwidth]{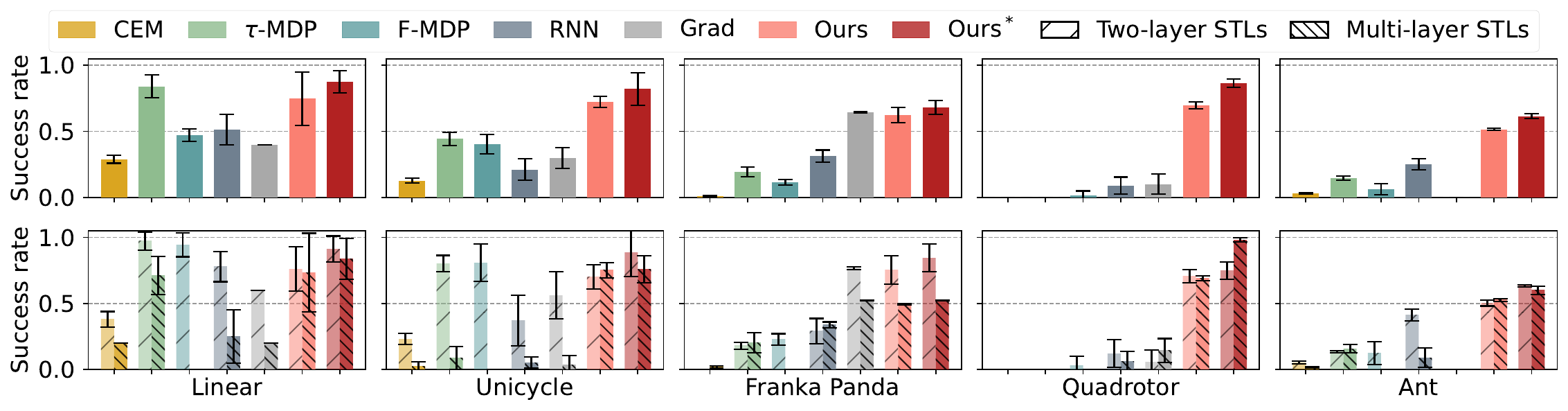}
    \caption{Main comparison. Our method has higher task success rate compared to other baselines.}
    \label{fig:main-result}
\end{figure}

\begin{figure}[!htbp]
    \centering
    \includegraphics[width=0.99\textwidth]{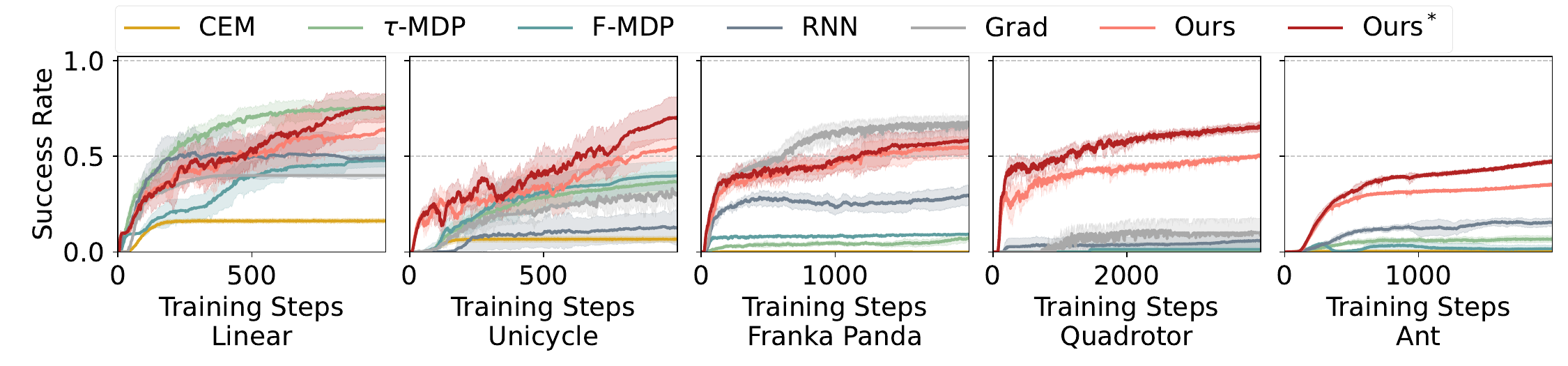}
    \caption{Main comparison for the STL success rate evaluation along the training process.}
    \label{fig:reward-curve}
\end{figure}

As shown in Fig.~\ref{fig:main-result} (top row), TGPO achieves the leading performance in most benchmarks, and with Bayesian time variable sampling, TGPO\textsuperscript{*} achieves the highest overall success rate across all benchmarks, indicating the strong empirical performance. Our advantage becomes clearer as the system dimension and the planning difficulty increase, especially in ``Quadrotor" and ``Ant", where most of the baselines achieve less than 10\% success rate, whereas TGPO\textsuperscript{*} can achieve 86.46\% and 61.57\% success rate, respectively. Under ``Linear" system, the best baseline $\tau$-MDP (84.11\%) performs competitively compared to TGPO\textsuperscript{*} (87.53\%), but $\tau$-MDP's performance drops drastically on the other benchmarks. The ``Grad" method is a strong baseline on ``Franka Panda", however, its success rate decreases by a large margin on ``Quadrotor" due to its complex nonlinear dynamics, and it cannot work at all on ``Ant", which is likely caused by the discrepancy between the simulator's approximated gradients and the true non-differentiable dynamics. These findings showcase TGPO's strong performance and great adaptation to high-dimensional and non-differentiable environments. If we look at different types of STLs (Fig.~\ref{fig:main-result}, bottom row), on low-dimensional cases (``Linear" and ``Unicycle"), most baselines work well under the simple STL tasks (``two-layer STLs") but they struggle on the harder STLs (``multi-layer STLs", note that F-MDP can only handle ``two-layer STLs"). Whereas our approaches (both TGPO and TGPO\textsuperscript{*}) excel at working on these complex STLs and perform consistently well. This shows our approach's strength in handling complex STLs. In Fig.~\ref{fig:reward-curve}, we show the task success rate in training. Our approach can achieve a high task success rate eventually, whereas other baselines show plateauing early in the training.

\subsection{Solving STL with Different horizon-lengths}
\begin{figure}[!htbp]
    \centering
    \includegraphics[width=0.8\textwidth]{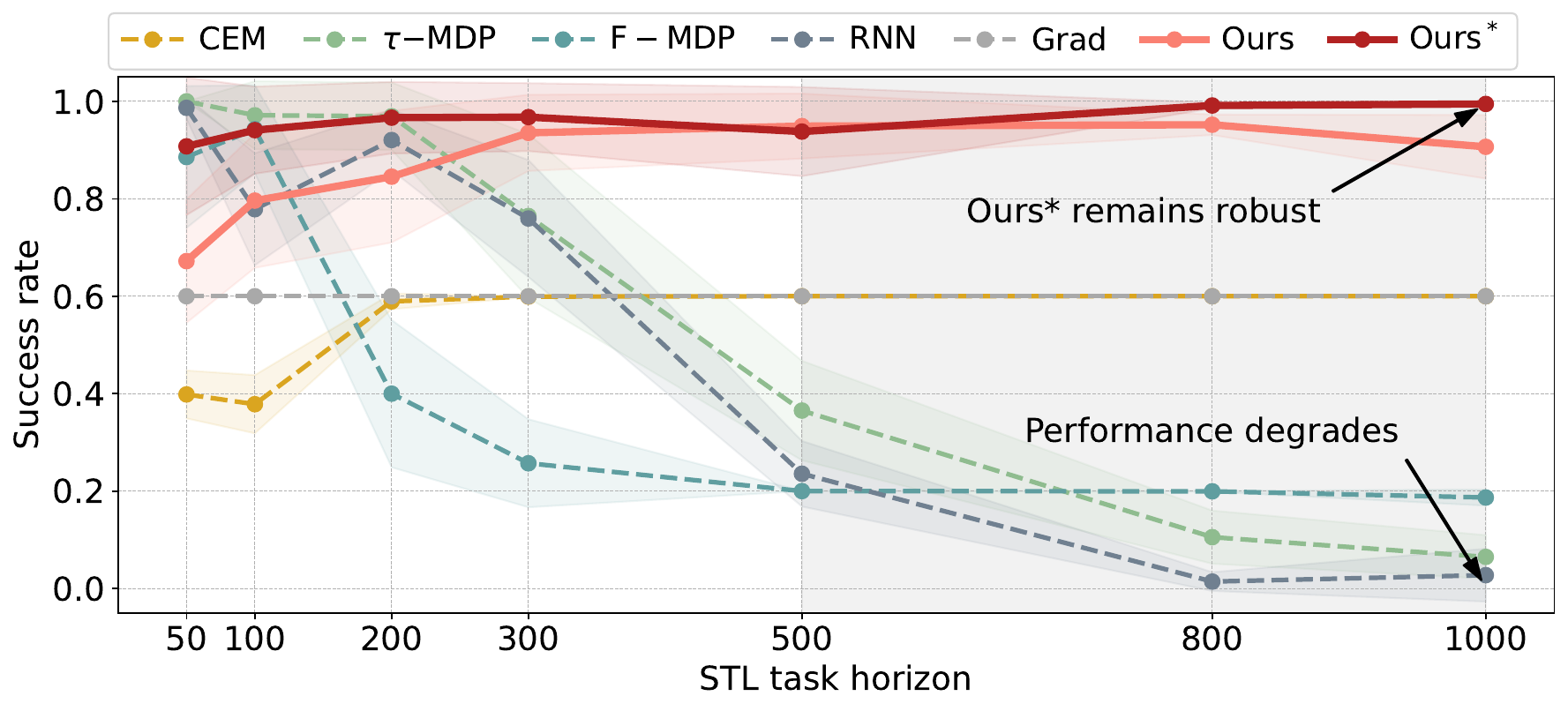}
    \caption{Solving STL in ``Linear" environment over varied task horizons. Our method performs the best and maintains high success rate in long horizons where the RL baselines performance degrades.}
    \label{fig:horizon-curve}
\end{figure}

Beyond system complexity and task difficulty, our methods also show resilient adaptivity for long-horizon tasks. Here, we consider only the two-level STLs and we scale the task horizon to different lengths (50, 200, 300, 800 and 1000). As shown in Fig.~\ref{fig:horizon-curve}, our methods (TGPO and TGPO\textsuperscript{*}) keep a high success rate over varied time lengths, whereas for RL methods $\tau$-MDP, F-MDP and RNN, which are strong baselines for shorter horizons ($T$=50 and 100), experience a huge drop in success rate as the horizon increases. It is interesting that CEM and Grad can maintain their performance as the horizon expands 10 times, which may be attributed to their trajectory optimization formulation.

\subsection{Ablation studies}
\label{sec:exp-abl}
\begin{table}[htbp]
\centering
\caption{Ablation studies for TGPO on the linear dynamics environment.}
\label{tab:ablations_combined}
\setlength{\tabcolsep}{4pt}
\begin{subtable}[t]{0.53\textwidth}
    \centering
    \caption{Different time variables sampling strategies.}
    \label{tab:ablation_sampling}
    \begin{tabular}{lcccl}
    \toprule
    \textbf{Method} & Rand. & Bay. & Elite & Test(\%) \\
    \midrule
    Ours & \checkmark & & & 80.33 $\pm$ 8.84\\
    Ours\textsubscript{Bay} & & \checkmark & & 53.79 $\pm$ 7.99 \\
    Ours\textsubscript{Elite} & & & \checkmark & 61.49 $\pm$ 10.02 \\
    Ours\textsubscript{mixBay} & \checkmark & \checkmark & & 81.18 $\pm$ 9.72 \\
    Ours\textsubscript{mixElite} & \checkmark & & \checkmark & 86.62 $\pm$ 8.67 \\
    Ours\textsubscript{BayElite} & & \checkmark & \checkmark & 81.04 $\pm$ 11.00 \\
    \midrule
    \textbf{Ours\textsuperscript{*}}& \checkmark & \checkmark & \checkmark &\textbf{88.99 $\pm$ 9.60} \\
    \bottomrule
    \end{tabular}
\end{subtable}%
\hfill 
\begin{subtable}[t]{0.45\textwidth}
    \centering
    \caption{Different state augmentation and rewards.}
    \label{tab:ablation_s_r}
    \begin{tabular}{ll}
    \toprule
    State aug. / Reward & Test(\%) \\
    \midrule
    t+flags / STL & 11.73 $\pm$ 2.67 \\
    t+flags / STL+Inv & 46.85 $\pm$ 10.53 \\
    t+flags / STL+Inv+Prog & 49.80 $\pm$ 7.72 \\
    t+flags / STL+Inv+Dist & 84.59 $\pm$ 7.88 \\
    $\emptyset$ / STL+Inv+Dist & 11.43 $\pm$ 3.48 \\
    t / STL+Inv+Dist & 47.51 $\pm$ 7.86 \\
    \midrule
    \textbf{Ours\textsuperscript{*}} (all / all) & \textbf{88.99 $\pm$ 9.60} \\
    \bottomrule
    \end{tabular}
\end{subtable}
\end{table}

We conduct a thorough ablation study under ``Linear" (all 10 STLs) for the analysis. We first study different sampling strategies. As shown in Tbl.~\ref{tab:ablation_sampling}, our base model with random sampling (Ours) can already achieve 80.33\% success rate ($\pm$ indicates the standard deviation over 7 random seeds). However, naively using Bayesian sampling (Ours\textsubscript{Bay}) or Elite variable replay buffer (Ours\textsubscript{Elite}) will hurt the performance, likely due to the myopic exploration at the beginning of the training, which restricts the agent from seeking more promising assignments. Hence, we mix the two sources of the time variables together and witness certain improvement (Ours\textsubscript{mixBay}, Ours\textsubscript{mixElite}, and Ours\textsubscript{BayElite}) compared to Ours. Finally, by combining all these together, Ours\textsuperscript{*} achieves the best performance.

In Tbl.~\ref{tab:ablation_s_r} we study how state augmentation and reward shaping foster an efficient multi-stage RL. For the reward design, we consider to just using parts of the reward terms introduced before, and the results (the first 4 rows) show that, just using STL robustness score will only result in 11.73\% success rate, whereas by gradually adding invariance penalty, progress reward and the distance reward, the performance will get improved (the most improvement comes from using the distance reward term) and finally becomes 88.99\% for Ours\textsuperscript{*}. Regarding the state augmentation, removing the flags in the augmented state will result in a 41.48\% drop in success rate, and if further removing the time index counter, the performance will drop to 11.43\%. The combined findings validate our design.
 
\subsection{Visualization for interpretability and multi-modal behavior}
\begin{figure}[!htbp]
    \centering
    \includegraphics[width=0.9\textwidth]{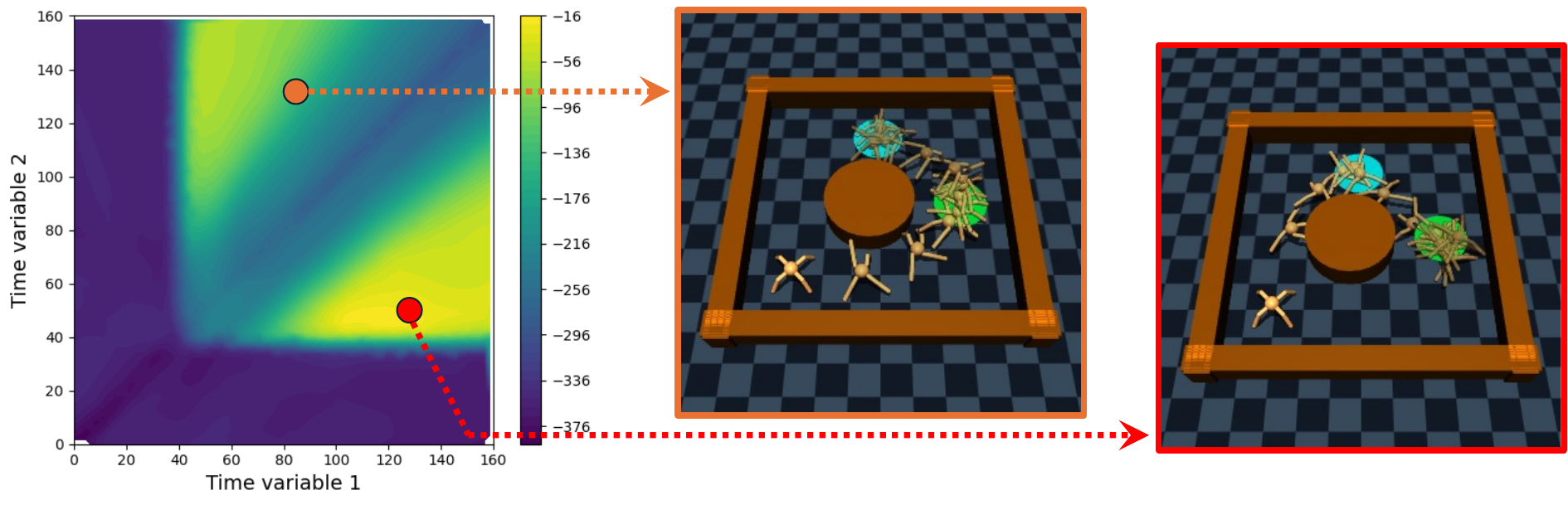}
    \caption{Critic value visualization and simulation for the ant environment under STL $F_{[0,160]}(A_1) \land F_{[0,160]}(A_2) \land G_{[0,200]}\neg B$. From the two feasible regions on the critic heatmap, we can see that the corresponding conditioned policy generates two behaviors to fulfill the task specification.}
    \label{fig:critic-heatmap}
\end{figure}
TGPO can generate diverse behaviors to fulfill the STL specifications, which can also be reflected from the critic values. We consider an example under the Ant environment for the STL task $F_{[0,160]}(A_1) \land F_{[0,160]}(A_2) \land G_{[0,200]}\neg B$. The ant starts from the lower left, and there is an obstacle in the middle of the scene. The time variables here correspond to ``Reach $A_1$" (the cyan region in the scene) and ``Reach $A_2$" (green). After the training, we plot the critic value heatmap across different time variable assignments for the initial state. As shown in Fig.~\ref{fig:critic-heatmap}, the lower-left L-shape region is in low critic value as it is dynamically infeasible to reach the first subgoal in a short time (0$\leq\tau\leq$40). The diagonal line region also receives low critic value, because the two subgoal regions cannot be visited in such a short time. The diagonal line splits the promising time variable regions (yellow) into two parts, from which we can generate two different ways to fulfill the STL task (as shown from the time-elapsed simulation plot on the right). This shows that we can leverage the time variables as the condition to generate multi-modal solutions to solve the STL problem.

\subsection{Limitations}
\label{sec:limitations}
While our method achieves strong empirical performance, it lacks formal guarantees on convergence to a global optimum. TGPO is effective on STLs with conjunctions and temporal operators, but it might not efficiently handle STLs with disjunctions or infinite-horizon task requirements like ``Always-Eventually (G(F))". In our paper, we have tested TGPO with 5 time variables; its scalability towards more complex STLs remains an open question. We aim to address these in future work.
\section{Conclusion}
\label{sec:conclusion}
In this paper, we introduce Temporal Grounded Policy Optimization (TGPO), a novel reinforcement learning framework for solving long-horizon Signal Temporal Logic tasks. By using STL decomposition, time variable sampling, state augmentation and reward design, TGPO can effectively handle general and complex STL tasks. Our experiments demonstrate that TGPO significantly outperforms existing baselines across various robotic environments and STL formulas. Future work will focus on extending TGPO to handle a broader class of STL formulas and improving its scalability.

\bibliography{z7_references}
\bibliographystyle{iclr2026_conference}
\newpage
\appendix
\section{Appendix}

\subsection{Algorithm hyperparameters}
All the main hyperparameters used during training are shown in Table~\ref{tab:hyperparams}. 

\begin{table}[!ht]
\caption{Hyperparameters assignments used for training TGPO\textsuperscript{*}.}
\label{tab:hyperparams}
\centering
\begin{tabular}{lc}
\toprule
\textbf{Hyperparameter} & \textbf{L}inear; \textbf{U}nicycle; \textbf{F}rankaPanda; \textbf{Q}uadrotor; \textbf{A}nt\\
\midrule 
Network hidden units & (512, 512, 512) \\
Optimizer & Adam \\
Learning rate & $3 \times 10^{-4}$ \\
Weight decay &$0.1$ \\
Grad norm clip &$0.5$ \\
Random seeds & 1007,1008,1009,1010,1011,1012,1013 \\
Batch size & 512 \\
Epochs & 1000 (L, U); 2000 (F, A); 4000(Q) \\
Time steps $T$ & 100 (L, U, F, Q); 200 (A) \\
Time duration $\Delta t$ & 0.2 (L, U); 0.05 (F, A); 0.1 (Q) \\
Distance reward $\lambda_1$ & 0.5 \\
Progress reward $\lambda_2$ & 20.0\\
Success reward $\lambda_3$ & 20.0\\
Invariance penalty $\lambda_4$ & -3.0 (L, F, Q); -3.5 (U); -1.5 (A)\\
Number of MCMC steps $N_{MCMC}$ & 500 \\
Number of warmup steps $N_{warmup}$ & 200 \\
Number of MCMC chains $M_{MCMC}$ & 512 \\
Ratio of Randomly-sampled time variables $\eta_{mcmc}$ & 0.5 \\
Ratio of MCMC-sampled time variables $\eta_{uniform}$ & 0.4 \\
Ratio of Elite time variables $\eta_{elite}$ & 0.1 \\
Elite buffer size $|\mathcal{B}|$ & 512 \\
\bottomrule
\end{tabular}
\end{table}

\subsection{Simulation environment details}
\label{appendix:env}
In this paper, we conduct experiments on five simulation environments (Linear, Unicycle, Franka Panda, Quadrotor, and Ant). The first four environments were implemented in plain JAX code by writing out the system dynamics, whereas the last one was adopted from the Mujoco JAX implementation. Detailed implementations are listed as follows.

\subsubsection{Linear}
We use a single-integrator dynamics model. The 2D state $(x, y)^T$ represents the 2D coordinates on a xy-plane, and the 2D control input $(v, w)^T$ reflects the velocities in these two directions. The system dynamics is described as:
\begin{equation}
    \begin{cases}
        x_{t+1}=x_t+v_t \Delta t\\
        y_{t+1}=y_t+w_t \Delta t\\
    \end{cases}
\end{equation}
We set the time step duration $\Delta t=0.2s$.

\subsubsection{Unicycle}
We use a car-like dynamics model. The 4D state $(x,y,\theta,v)^T$ represents the 2D coordinates on the xy-plane, the heading angle of the robot and the velocity of the robot, respectively. The 2D input $(\omega, a)^T$ represents the angular velocity and the acceleration. The system dynamics can be described as:
\begin{equation}
    \begin{cases}
        x_{t+1}=x_t+v_t\cos(\theta_t) \Delta t\\
        y_{t+1}=y_t+v_t\sin(\theta_t) \Delta t\\
        \theta_{t+1}=\theta_t + \omega_t \Delta t\\
        v_{t+1}=v_t + a_t \Delta t\\
    \end{cases}
\end{equation}
We set the time step duration $\Delta t=0.2s$. The control actuation is limited at $[-1 rad/s, +1 rad/s] \times [-4m/s^2, +4m/s^2]$. The scene layout is $[-5m, +5m] \times  [-5m, +5m]$ on the xy-plane.

\subsubsection{Franka Panda}
We use a 7 DoF Franka Panda robot arm model to conduct the simulation. The 7D state $(\theta_1, \theta_2, ..., \theta_7)^T$ represents the angle for all the joints where $\theta_7$ is for the end-effector joint. The 7D control input $(\omega_1, \omega_2,..., \omega_7)^T$ represents the angular velocity for all the joints. The dynamics follows a simple single-integrator case: $\theta_{i,t+1}=\theta_{i,t} + \omega_{i,t} \Delta t, \text{ for } i=1,2,...,7$. We set the time step duration $\Delta t=0.05s$.

\subsubsection{Quadrotor}
We use a full quadrotor dynamics model~\citep{tayebi2006attitude} to conduct the simulation. The 12D state $(x, y, z, v_x, v_y, v_z, \phi, \theta, \psi,\omega_x,\omega_y,\omega_z)^T$ represents the 3D coordinate $\mathbf{p}=(x,y,z)^T$, the velocity vector $\mathbf{v}=(v_x,v_y,v_z)^T$, the orientation vector $\boldsymbol{\Theta}=(\phi,\theta,\psi)^T$, and the angular velocity $\boldsymbol{\omega}=(\omega_x,\omega_y,\omega_z)^T$, respectively. The 4D control input $(f_1, f_2, f_3, f_4)^T$ represents the lifting force from the four motors. The full dynamics are:

\begin{equation}
\begin{cases}
\mathbf{p}_{t+1} = \mathbf{p}_t + \mathbf{v}_t \Delta t \\
\mathbf{v}_{t+1} = \mathbf{v}_t + (g\mathbf{e}_3 - \frac{T}{m} R_z(\psi)R_y(\theta)R_x(\phi)\mathbf{e}_3) \Delta t \\
\boldsymbol{\Theta}_{t+1} = \boldsymbol{\Theta}_t + \boldsymbol{\omega}_t \Delta t \\
\boldsymbol{\omega}_{t+1} = \boldsymbol{\omega}_t + I^{-1}(\boldsymbol{\tau} - \boldsymbol{\omega}_t \times (I\boldsymbol{\omega}_t)) \Delta t
\end{cases}
\end{equation}
with the rotation matrices $R_x(\phi)=\begin{bmatrix}
    1 & 0 & 0 \\
    0 & \cos(\phi) & -\sin(\phi)\\
    0 & \sin(\phi) & \cos(\phi)
\end{bmatrix}$, 
$R_y(\theta)=\begin{bmatrix}
    \cos(\theta) & 0 & \sin(\theta)\\
    0 & 1 & 0 \\
    -\sin(\theta) & 0 & \cos(\theta)
\end{bmatrix}$,
and $R_z(\psi)=\begin{bmatrix}
    \cos(\psi) & -\sin(\psi) & 0\\
    \sin(\psi) & \cos(\psi) & 0\\
    0 & 0 & 1\\
\end{bmatrix}$ and $T$ and $\boldsymbol{\tau}$ are the total thrust and the torques derived from the motor input $u$ with the Coriolis effect considered to the angular velocity vector. We set the time step duration $\Delta t=0.10s$, adapt the gravity coefficient $g=9.81m/s^2$ with the corresponding gravity vector $\mathbf{e}_3=(0, 0, 1)^T$, set the total mass of the quadrotor $m=0.2kg$ and set the diagonal line of the quadrotor inertia matrix $I$ as $(0.01kg\cdot m^2, 0.01kg\cdot m^2, 0.02kg\cdot m^2)^T$.

\subsubsection{Ant}
In this case, the agent is a 8-DoF quadruped robot with the complex dynamics implemented in Brax~\citep{freeman2021brax}. The observation space is 29-dimension (3-dimension for xyz coordinates, 4-dimension for the torso orientation (in Quaternion representation), 3-dimension velocity vector and 3-dimension angular velocity for the torso, 8-dimension for the joints' angles and another 8-dimension for the joints' angular velocities). The original control input is 8-dimension for the torques applied to each of the 8 joints. To ease the RL training, we first train a goal-reaching policy, enabling the ant to learn and move to a specified target location. Then, for the baselines and our methods, the problem becomes planning the waypoints so that the ant can satisfy the STL tasks specified.

\subsection{Baseline implementation details}
\label{appendix:baselines}
\subsubsection{CEM}
We use the Cross Entropy Method baseline mentioned in~\citep{meng2023signal}, which belongs to the evolutionary search algorithm mentioned in~\citep{salimans2017evolution}. We denote the initial neural network policy parameters as $\theta^{(0)}$. At $j$-th iteration, we draw $N$ samples $\theta_1,...\theta_N$ from $\mathcal{N}(\theta^{(j)}, {\sigma^{(j)}}^2)$ where $\sigma^{(j)}$ is the preset standard deviation, then we rollout the trajectories and compute their robustness score. We pick the top-$K$ candidates parameters $\theta_{E_1},...\theta_{E_k}$. Then we update the estimate for the neural network parameters $\theta^{(j+1)}=\frac{1}{k}\sum\limits_{i=1}^k \theta_{E_i}$ and $\sigma^{(j+1)}=\sqrt{\frac{1}{k-1}\sum\limits_{i=1}^k(\theta_{E_i}-\theta^{(j+1)})^2}$. We repeat this process for $L$ iterations to get the final parameters. We set the size for the elite pool to be $K=32$ and set the population sample size to be $N=512$. The number of iteration steps $L$ is the same as our method ($L=1000$ for ``Linear" and ``Unicycle", $L=2000$ for ``Franka Panda" and ``Ant", and $L=4000$ for ``Quadrotor".)

\subsubsection{\texorpdfstring{$\tau$-MDP}{tau-MDP}}
$\tau$-MDP is an RL method introduced in~\citep{aksaray2016q} to solve STL tasks under the discrete state space. The original method appends history to the state space, and uses Q-learning to solve short-horizon tasks with 2-layer STL specifications. Here, we extend it to handle general STL formulas by augmenting the entire trajectory into the state space with STL robustness score as the terminal reward to guide the agent to satisfy STL tasks. We also changed the RL backbone from Q-learning to PPO for better scalability to longer-horizon tasks (The original Q-learning tabular formulation will not work on continuous space for $T=100$).

\subsubsection{\texorpdfstring{$F$-MDP}{F-MDP}}
$F$-MDP is an improved RL method introduced in~\citep{venkataraman2020tractable} to solve STL tasks under the discrete state space more efficiently. This approach considers the 2-layer STL specifications, and introduces a flag for each of the subformulas in the STL. They defined the state transition rules and reward mechanism for ``F" and ``G"-based subformulas based on these flags and show that the Q-learning under this augmentation can learn more efficiently than the Q-learning under $\tau$-MDP~\citep{aksaray2016q}. We re-implemented $F$-MDP in PPO for our comparison.

\subsubsection{RNN}
In this case, similar to~\citep{liu2021recurrent}, we use an RNN to encode the history data and then use the robustness score as the final reward to guide the agent to satisfy the tasks. The issue of this implementation is that it is much more time-consuming compared to the other baselines.

\subsubsection{Grad}
In this case, similar to~\citep{leung2022semi} and~\citep{meng2023signal}, we use a neural network policy to roll out the trajectory (in a deterministic manner, rather than sampling from the learned Gaussian distribution). At each time step, the network receives the state (and the time index) and generates the action, which is then sent to the environment to derive the next state. We repeat this process $T$ times to roll out the full trajectory, which preserves the gradient through the differentiable system dynamics. We use the approximated robustness score mentioned in~\citep{pant2017smooth} to ensure the score is differentiable. We then conduct backpropagation-through-time (BPTT) to update the neural network parameters.

\subsection{Temporal sampling algorithm details}
\label{appendix:mcmc}
The Metropolis-Hastings algorithm~\citep{chib1995understanding} is a Markov Chain Monte Carlo (MCMC) method for sampling from a probability distribution, commonly used when directly sampling from the distribution is hard. In our approach TGPO\textsuperscript{*}, we use a discrete version of the M-H algorithm to sample time variables $\mathbf{t}$ that are likely to yield high critic values $V_\psi(s_0, \mathbf{t})$, where $s_0$ is the initial state. We use $\exp(V_\psi(s_0,\mathbf{t}))$ as a proxy for the unnormalized probability of the promising temporal variables. The algorithm proceeds by starting with an initial set of temporal variables $\mathbf{t}_0$ and iteratively proposing to move on grids to a new set $\mathbf{t}'$ based on a proposal distribution $g(\mathbf{t}'|\mathbf{t})$. The move is then accepted or rejected based on the acceptance ratio $\alpha$, which compares the critic value exponentials of the new and the current variables. The process is detailed in Algorithm.~\ref{alg:mh-sampler}.

\begin{algorithm}[H]
\caption{Metropolis-Hastings for time variable sampling (with multiple chains and warm-up)}
\label{alg:mh-sampler}
\begin{algorithmic}[1]
\State \textbf{Input:} Initial state $s_0$, Critic network $V_\psi(s, \mathbf{t})$, Proposal distribution $g(\mathbf{t}'|\mathbf{t})$
\State \textbf{Input:} Iterations $N_{mcmc}$, Number of chains $M_{chain}$, Number of warm-up steps $N_{warmup}$
\ForAll{chain $m \in \{1, \dots, M_{chain}\}$}
    \State Initialize temporal variables $\mathbf{t}_{m,0}$ randomly
    \State Initialize samples list $S_m \gets []$
\EndFor
\For{$i = 1$ \textbf{to} $N_{mcmc}$}
    \ForAll{chain $m \in \{1, \dots, M_{chain}\}$}
        \State \Comment{Propose new temporal variables for chain $m$}
        \State $\mathbf{t}' \gets \text{Sample from } g(\mathbf{t}'|\mathbf{t}_{m, i-1})$
        \State \Comment{Calculate the acceptance ratio $\alpha$}
        \State $Q_{\text{current}} \gets V_\psi(s_0, \mathbf{t}_{m, i-1})$
        \State $Q_{\text{new}} \gets V_\psi(s_0, \mathbf{t}')$
        \State $\alpha \gets \min(1, \exp(Q_{\text{new}} - Q_{\text{current}}))$
        \State \Comment{Accept or reject the new sample for chain $m$}
        \State $u \gets \text{Sample from Uniform}(0, 1)$
        \If{$u < \alpha$}
            \State $\mathbf{t}_{m,i} \gets \mathbf{t}'$ \Comment{Accept the new sample}
        \Else
            \State $\mathbf{t}_{m,i} \gets \mathbf{t}_{m, i-1}$ \Comment{Reject and keep the old sample}
        \EndIf
    \EndFor
\EndFor
\State \Comment{Collect samples after the warm-up period}
\For{$i = N_{warmup} + 1$ \textbf{to} $N_{mcmc}$}
    \ForAll{chain $m \in \{1, \dots, M_{chain}\}$}
        \State Add $\mathbf{t}_{m,i}$ to $S_m$
    \EndFor
\EndFor
\State \textbf{Return:} Sampled time variables $\{\mathbf{t} | \mathbf{t} \in S_i, i=1,2, \dots, M\}$
\end{algorithmic}
\end{algorithm}
In our approach, we set $N_{mcmc}=500$, $N_{warmup}=200$, $M_{chain}=512$ and pick the time variable from each $S_i$ with the highest critic value to form the time variable set $\mathbf{T}_{mcmc}$ used in Alg.~\ref{alg:tgpo_hybrid}. For the proposal distribution $g(\mathbf{t}'|\mathbf{t})$, we use a uniform distribution over the local neighborhood of the current temporal variables $\mathbf{t}$: we first uniformly sample an index $j$ from the dimensions of $\mathbf{t}$ and then uniformly sample a move direction $\Delta \in \{-1, +1\}$. The proposed new set of variables $\mathbf{t}'$ is generated by applying this change to the selected index but also ensure that the new value $\mathbf{t}'$ is within the valid range (otherwise we keep $\mathbf{t}$ unchanged).

\subsection{Training time comparison}
\label{appendix:training-time}
\begin{figure}[!htbp]
    \centering
    \includegraphics[width=0.95\textwidth]{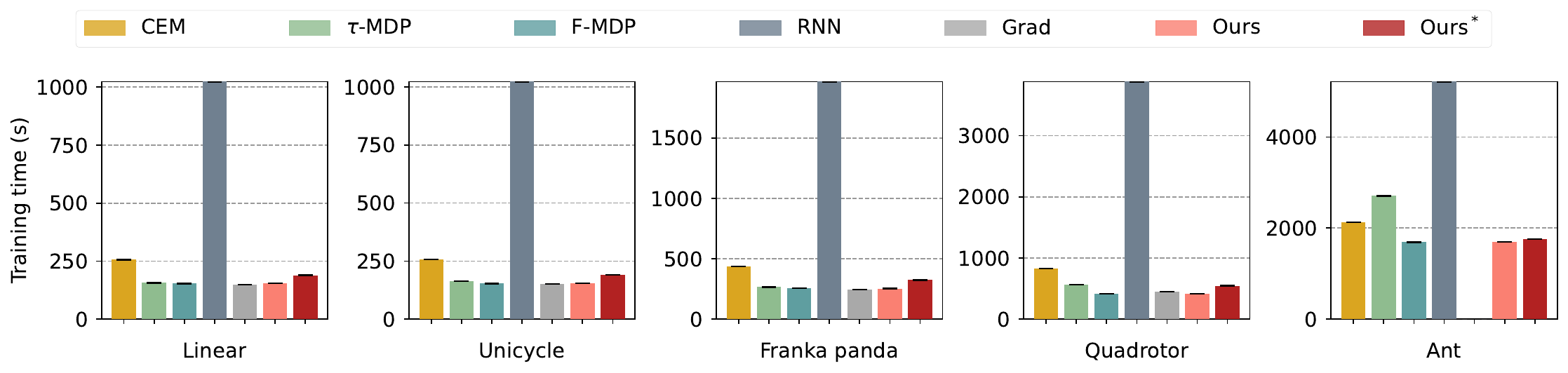}
    \caption{Comparison for training time. For each method under each environment, the result is averaged over 10 STLs and 7 random seeds. TGPO's training time is on par with leading baselines.}
    \label{fig:runtime}
\end{figure}

As shown in Fig.~\ref{fig:runtime}, TGPO and TGPO\textsuperscript{*} have a similar runtime compared to $\tau$-MDP, $F$-MDP and Grad baselines, whereas the CEM baseline is normally 20.8\%$\sim$35.8\% higher than TGPO\textsuperscript{*}. The most time-consuming baseline is RNN, where TGPO\textsuperscript{*} is 1.96X$\sim$6.11X faster in training speed. This shows that our approach is as scalable as other top RL baselines in training time, but our method can achieve higher task success rate.

\subsection{Correlation between the Critic and the STL robustness score}
\begin{figure}[!htbp]
  \centering
  \begin{subfigure}[b]{0.19\textwidth}
      \centering \includegraphics[width=\textwidth]{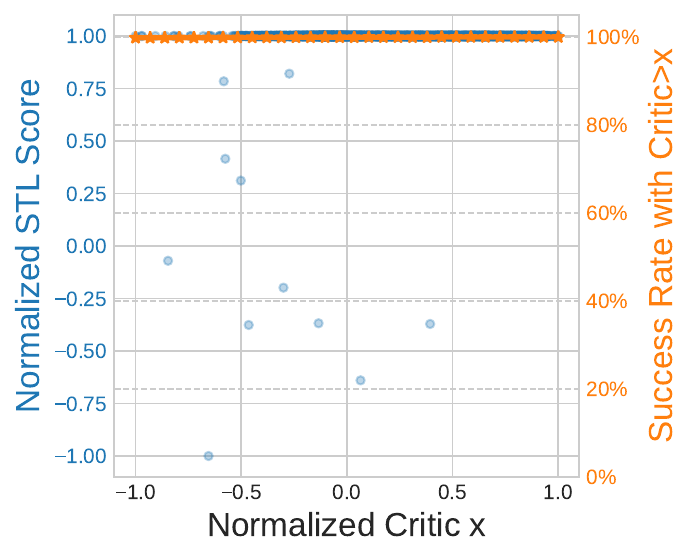}
      \caption{STL-01}
      \label{fig:scene-0lin-01}
  \end{subfigure}
  \begin{subfigure}[b]{0.19\textwidth}
      \centering \includegraphics[width=\textwidth]{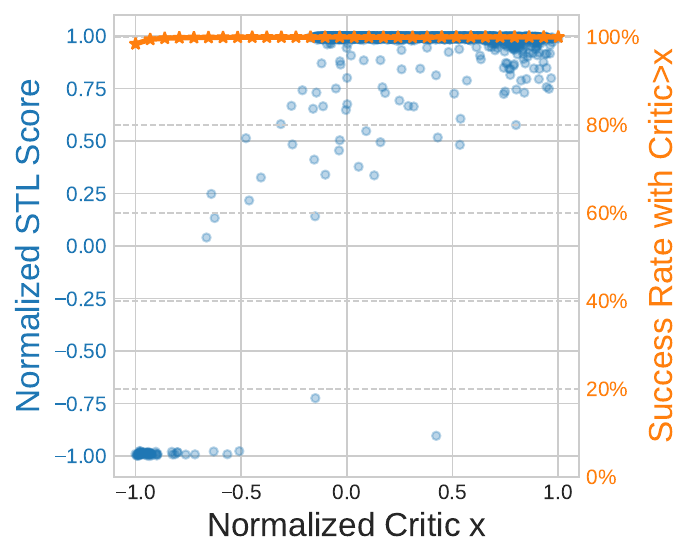}
      \caption{STL-02}
      \label{fig:scene-0lin-02}
  \end{subfigure}
  \begin{subfigure}[b]{0.19\textwidth}
      \centering \includegraphics[width=\textwidth]{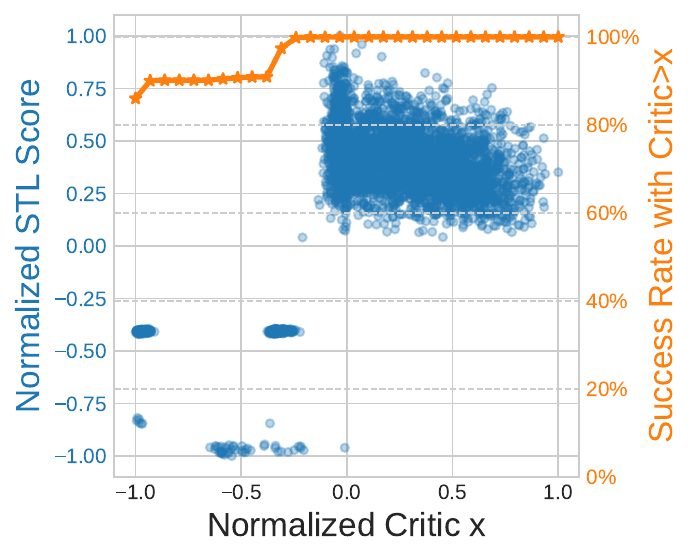}
      \caption{STL-03}
     \label{fig:scene-0lin-03}
  \end{subfigure}
  \begin{subfigure}[b]{0.19\textwidth}
      \centering \includegraphics[width=\textwidth]{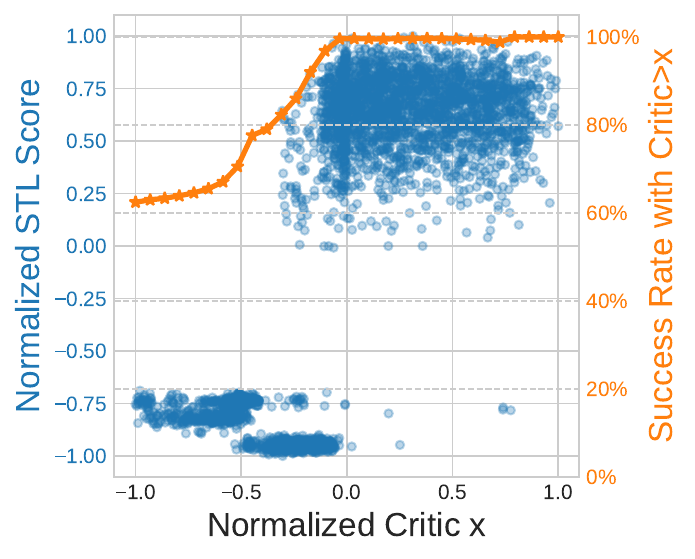}
      \caption{STL-04}
      \label{fig:scene-0lin-04}
  \end{subfigure}
  \begin{subfigure}[b]{0.19\textwidth}
      \centering \includegraphics[width=\textwidth]{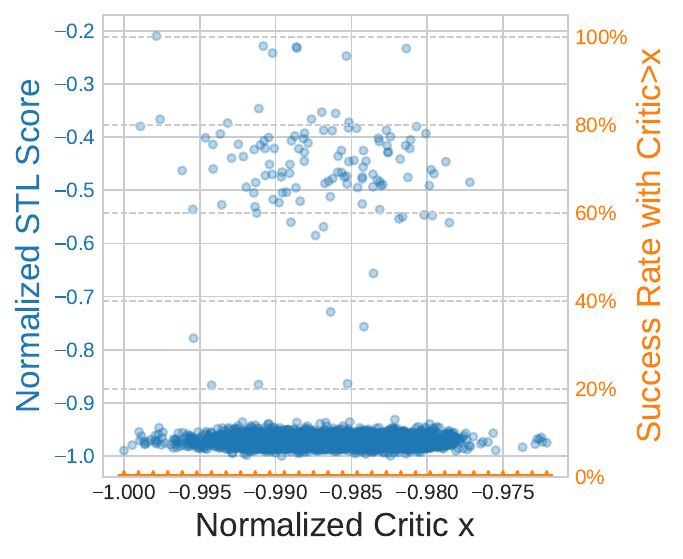}
      \caption{STL-05}
      \label{fig:scene-0lin-05}
  \end{subfigure}
  \begin{subfigure}[b]{0.19\textwidth}
      \centering \includegraphics[width=\textwidth]{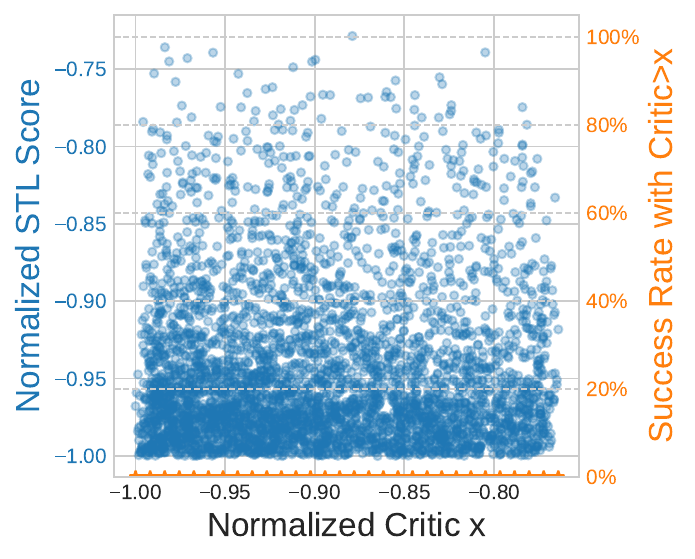}
      \caption{STL-06}
      \label{fig:scene-0lin-06}
  \end{subfigure}
  \begin{subfigure}[b]{0.19\textwidth}
      \centering \includegraphics[width=\textwidth]{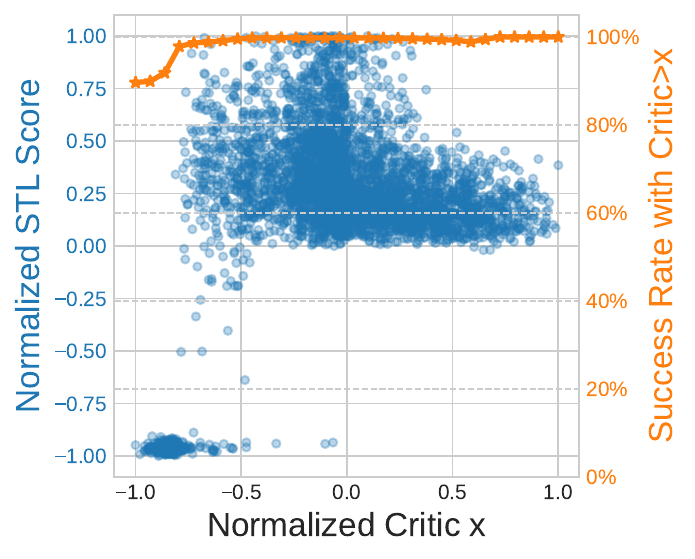}
      \caption{STL-07}
      \label{fig:scene-0lin-07}
  \end{subfigure}
  \begin{subfigure}[b]{0.19\textwidth}
      \centering \includegraphics[width=\textwidth]{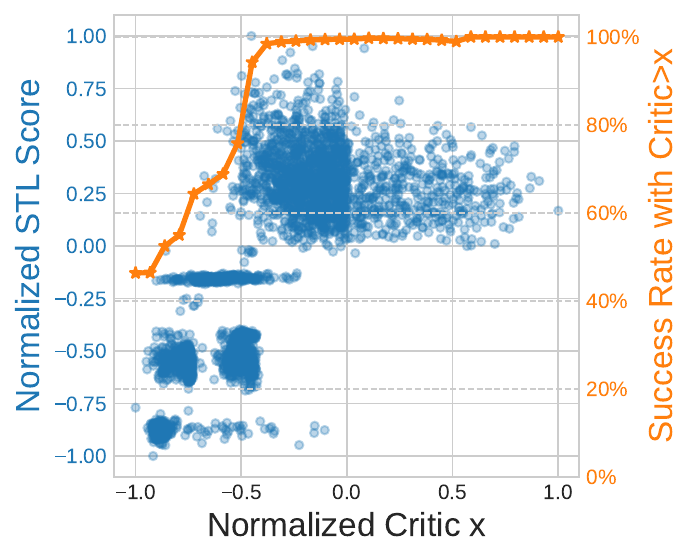}
      \caption{STL-08}
      \label{fig:scene-0lin-08}
  \end{subfigure}
  \begin{subfigure}[b]{0.19\textwidth}
      \centering \includegraphics[width=\textwidth]{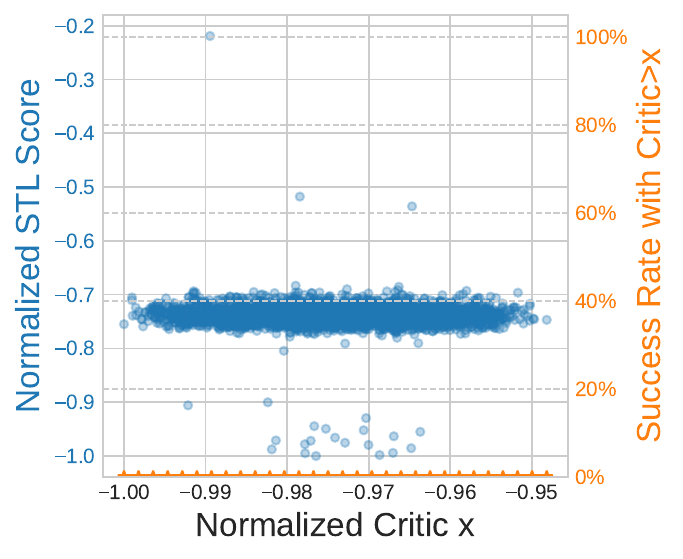}
      \caption{STL-09}
      \label{fig:scene-0lin-09}
  \end{subfigure}
  \begin{subfigure}[b]{0.19\textwidth}
      \centering \includegraphics[width=\textwidth]{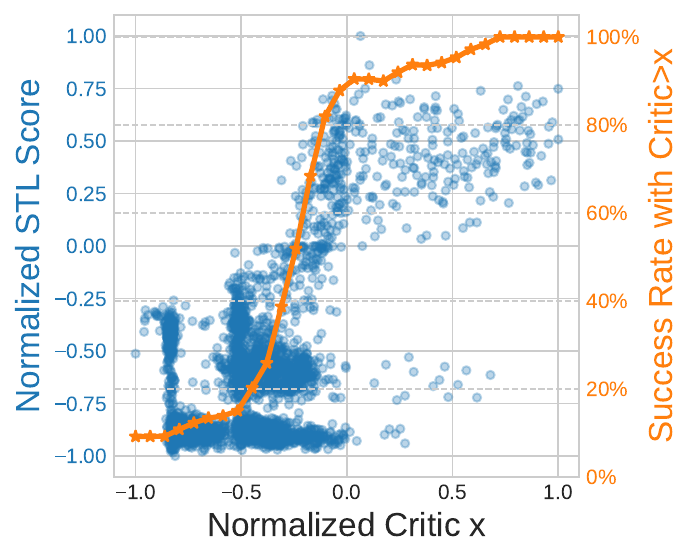}
      \caption{STL-10}
      \label{fig:scene-0lin-10}
  \end{subfigure}
  \caption{Correlation analysis between the critic and the STL score (seed=1007).}
  \label{fig:corr-plot-1007}
\end{figure}

\begin{figure}[!htbp]
  \centering
  \begin{subfigure}[b]{0.19\textwidth}
      \centering \includegraphics[width=\textwidth]{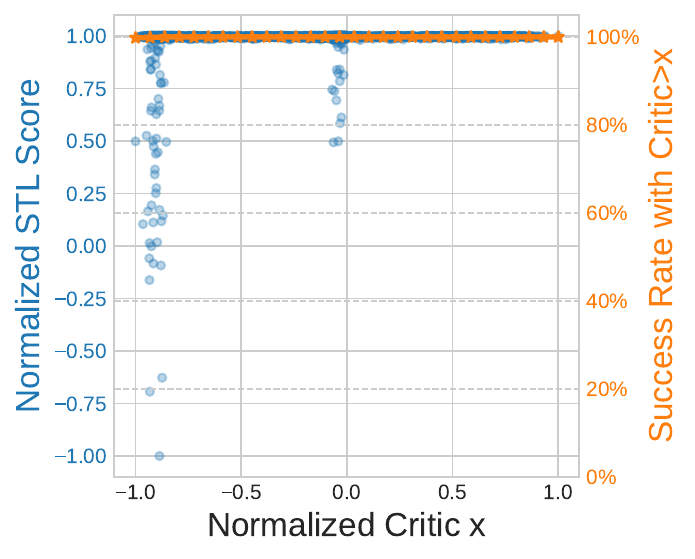}
      \caption{STL-01}
      \label{fig:scene-0lin-01-1}
  \end{subfigure}
  \begin{subfigure}[b]{0.19\textwidth}
      \centering \includegraphics[width=\textwidth]{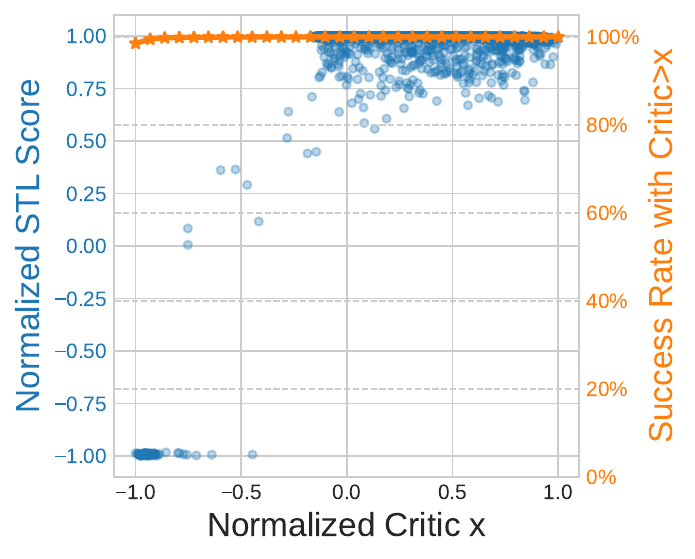}
      \caption{STL-02}
      \label{fig:scene-0lin-02-1}
  \end{subfigure}
  \begin{subfigure}[b]{0.19\textwidth}
      \centering \includegraphics[width=\textwidth]{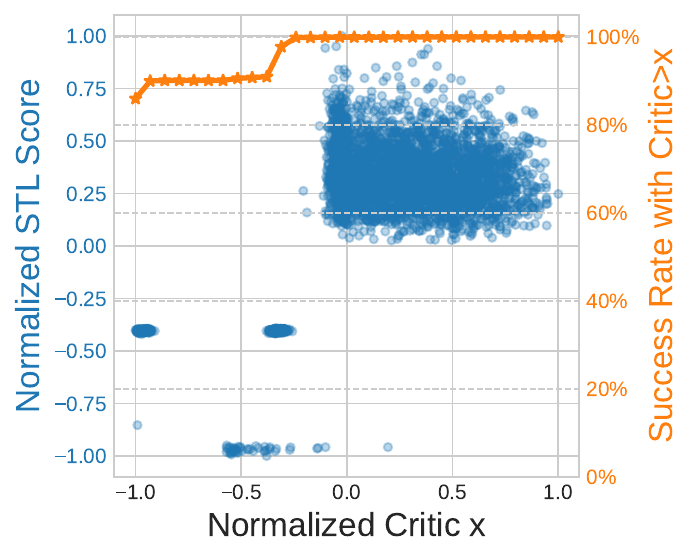}
      \caption{STL-03}
     \label{fig:scene-0lin-03-1}
  \end{subfigure}
  \begin{subfigure}[b]{0.19\textwidth}
      \centering \includegraphics[width=\textwidth]{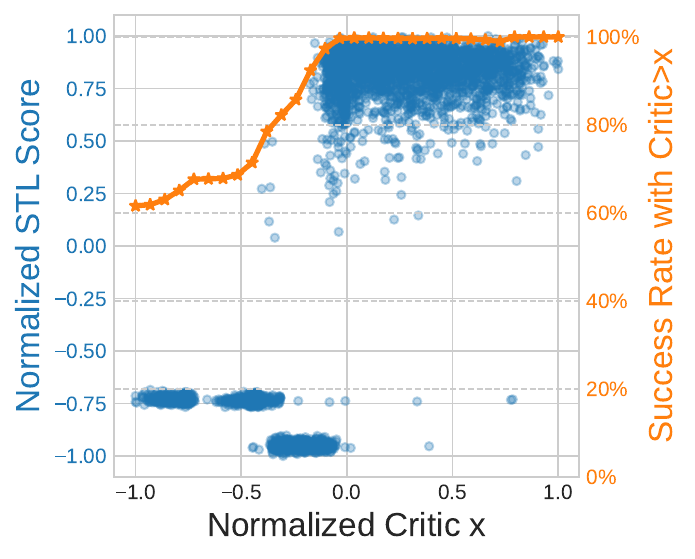}
      \caption{STL-04}
      \label{fig:scene-0lin-04-1}
  \end{subfigure}
  \begin{subfigure}[b]{0.19\textwidth}
      \centering \includegraphics[width=\textwidth]{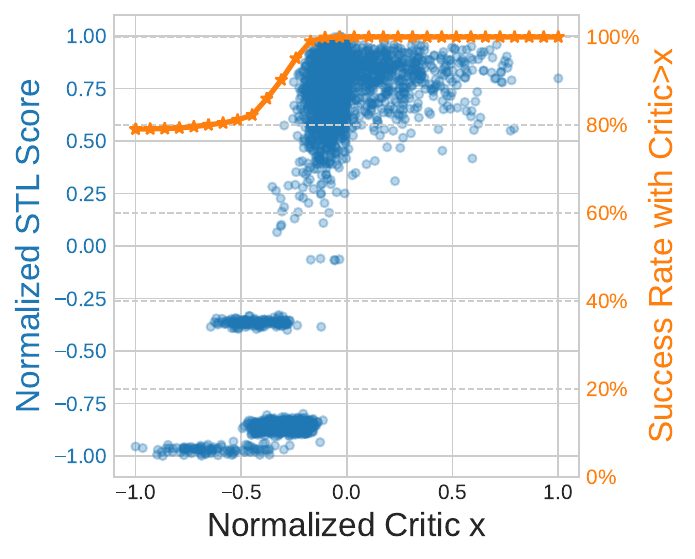}
      \caption{STL-05}
      \label{fig:scene-0lin-05-1}
  \end{subfigure}
  \begin{subfigure}[b]{0.19\textwidth}
      \centering \includegraphics[width=\textwidth]{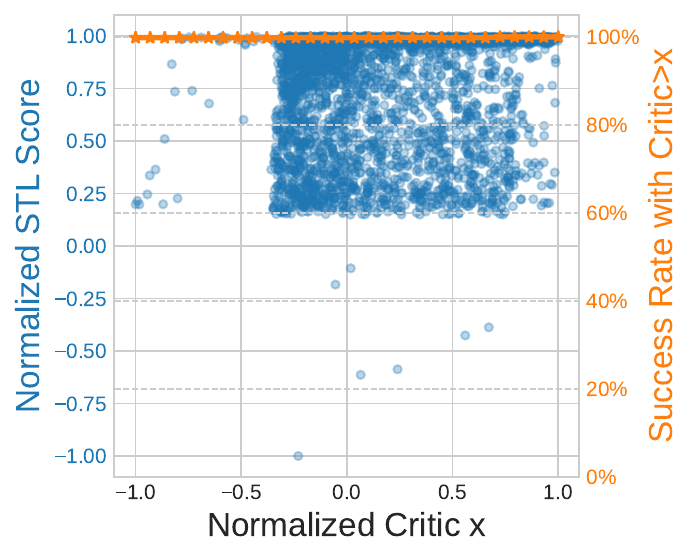}
      \caption{STL-06}
      \label{fig:scene-0lin-06-1}
  \end{subfigure}
  \begin{subfigure}[b]{0.19\textwidth}
      \centering \includegraphics[width=\textwidth]{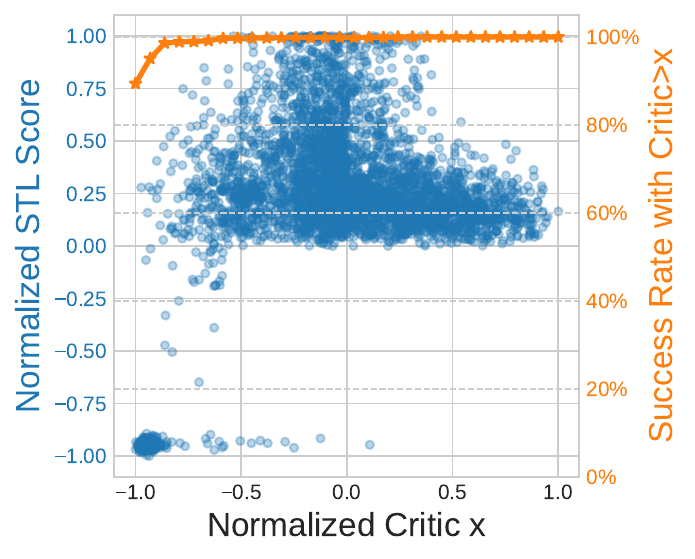}
      \caption{STL-07}
      \label{fig:scene-0lin-07-1}
  \end{subfigure}
  \begin{subfigure}[b]{0.19\textwidth}
      \centering \includegraphics[width=\textwidth]{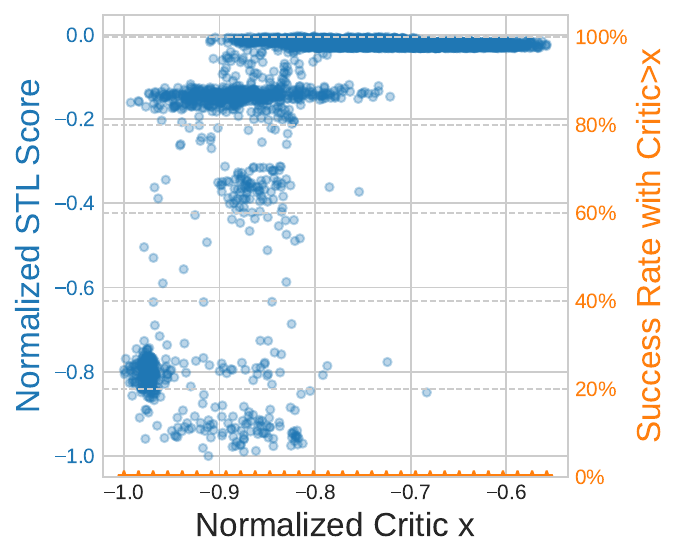}
      \caption{STL-08}
      \label{fig:scene-0lin-08-1}
  \end{subfigure}
  \begin{subfigure}[b]{0.19\textwidth}
      \centering \includegraphics[width=\textwidth]{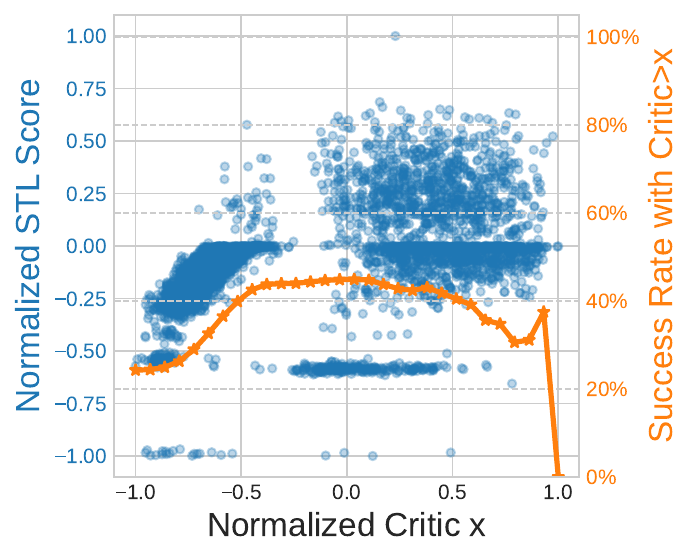}
      \caption{STL-09}
      \label{fig:scene-0lin-09-1}
  \end{subfigure}
  \begin{subfigure}[b]{0.19\textwidth}
      \centering \includegraphics[width=\textwidth]{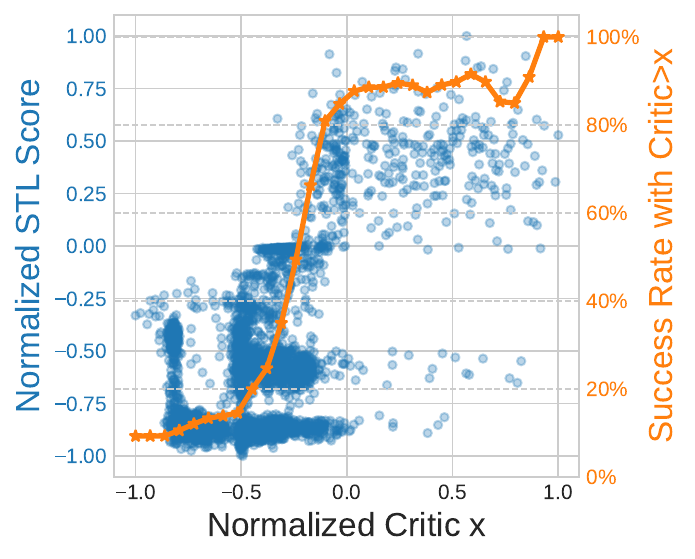}
      \caption{STL-10}
      \label{fig:scene-0lin-10-1}
  \end{subfigure}
  \caption{Correlation analysis between the critic and the STL score (seed=1008).}
  \label{fig:corr-plot-1008}
\end{figure}

To validate that our learned critic in TGPO can really reflect the ``promising" time variables that lead to STL satisfaction, in the ``Linear" environment, for the TGPO algorithm, we randomly sample 4096 points from the pretrained critic and rollout the corresponding trajectories to generate the STL robustness score. We plot the (critic value, STL score) scatter plot, together with the cumulative STL success rate curve for samples with a critic value greater than x. As shown in Fig.~\ref{fig:corr-plot-1007} (for seed=1007) and Fig.~\ref{fig:corr-plot-1008} (for seed=1008) from the blue scatter plots, whenever the critic value (left) is higher, the STL score is more likely to be higher, and hence more likely to satisfy STL. If we look at the orange curve, as the Critic value x increases, in most cases the probability for the corresponding traces satisfying the STL score is monotonously increasing or plateau at 100\%, which indicates that our critic is learned correctly (note that if the critic is not learned well, it could learn for some time variables that bring in high critic value but result in low STL scores, like STL-09 in Fig.~\ref{fig:corr-plot-1008}) and can be used to find ``promising" time variable assignments.

\subsection{STL Task details}
\label{appendix:stls}
Under each simulation environment, we make 10 STL formulas in two different categories (``two-level" and ``multi-level"). Here we only consider predicates related with ``Reach", ``Stay", ``Avoid" certain objects in the scene. They are listed as follows.
\subsubsection{STLs in ``Linear" environment}

\begin{figure}[!htbp]
  \centering
  \begin{subfigure}[b]{0.19\textwidth}
      \centering \includegraphics[width=\textwidth]{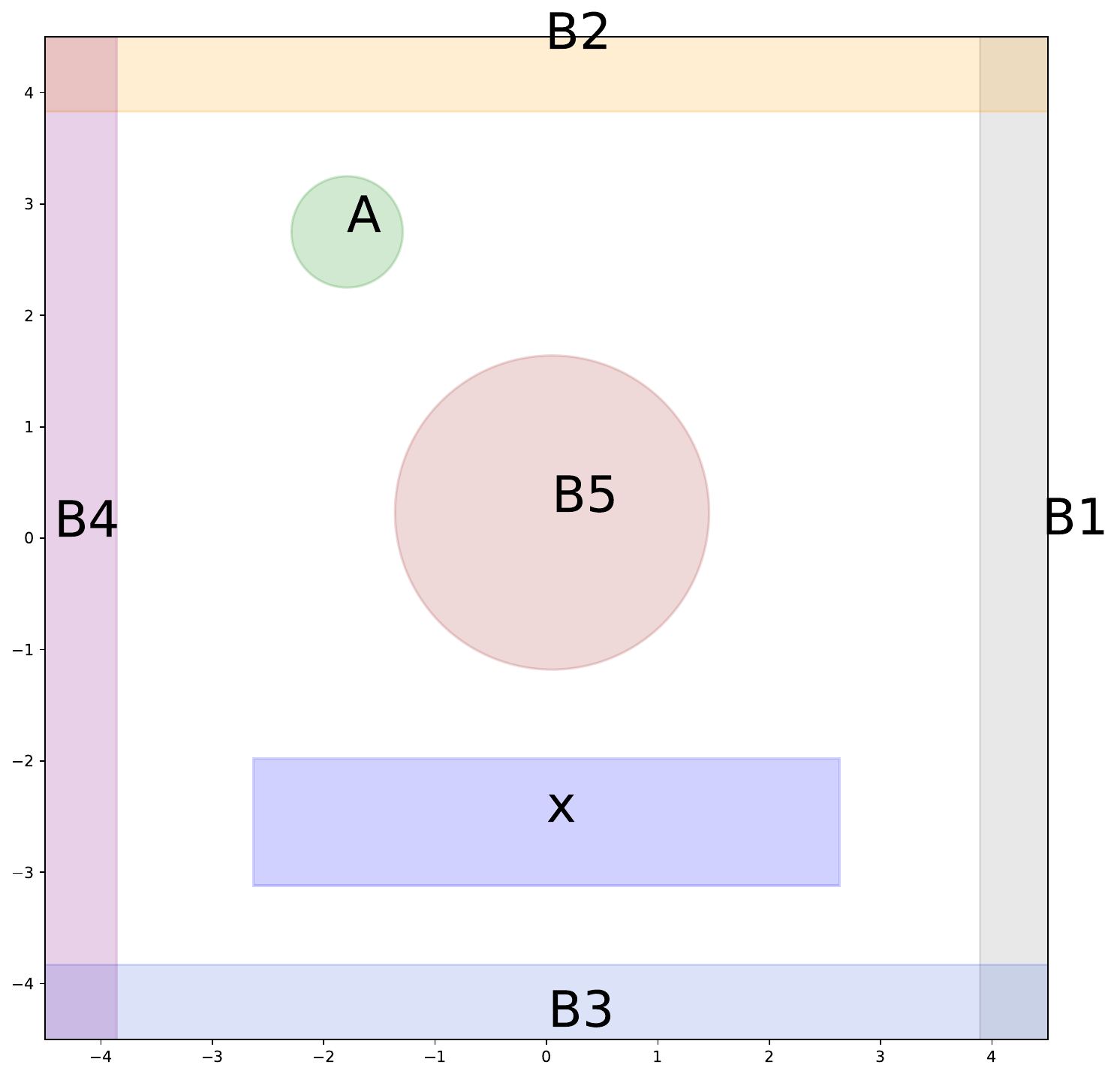}
      \caption{STL-01}
      \label{fig:scene-0lin-task-01}
  \end{subfigure}
  \begin{subfigure}[b]{0.19\textwidth}
      \centering \includegraphics[width=\textwidth]{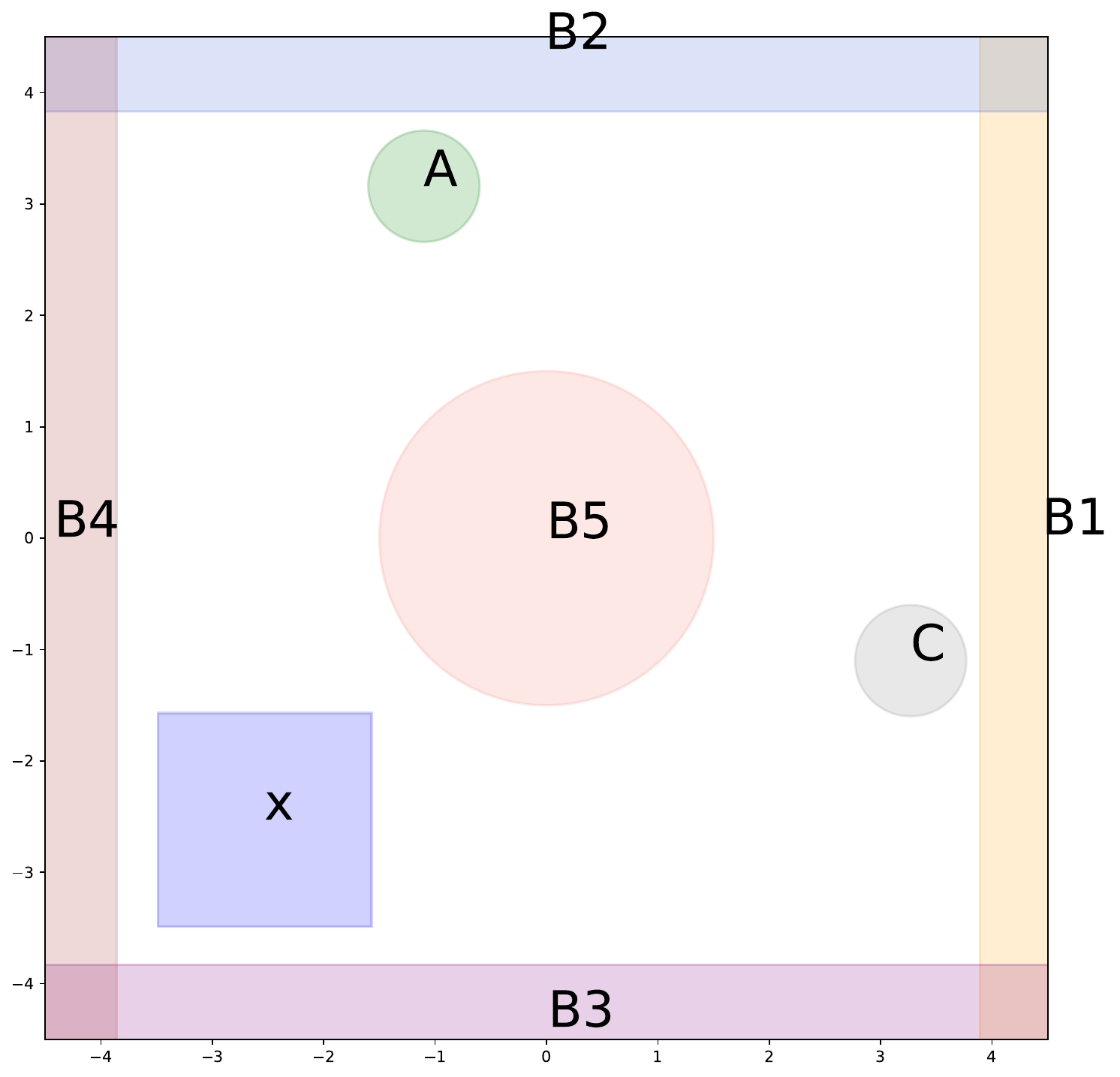}
      \caption{STL-02}
      \label{fig:scene-0lin-task-02}
  \end{subfigure}
  \begin{subfigure}[b]{0.19\textwidth}
      \centering \includegraphics[width=\textwidth]{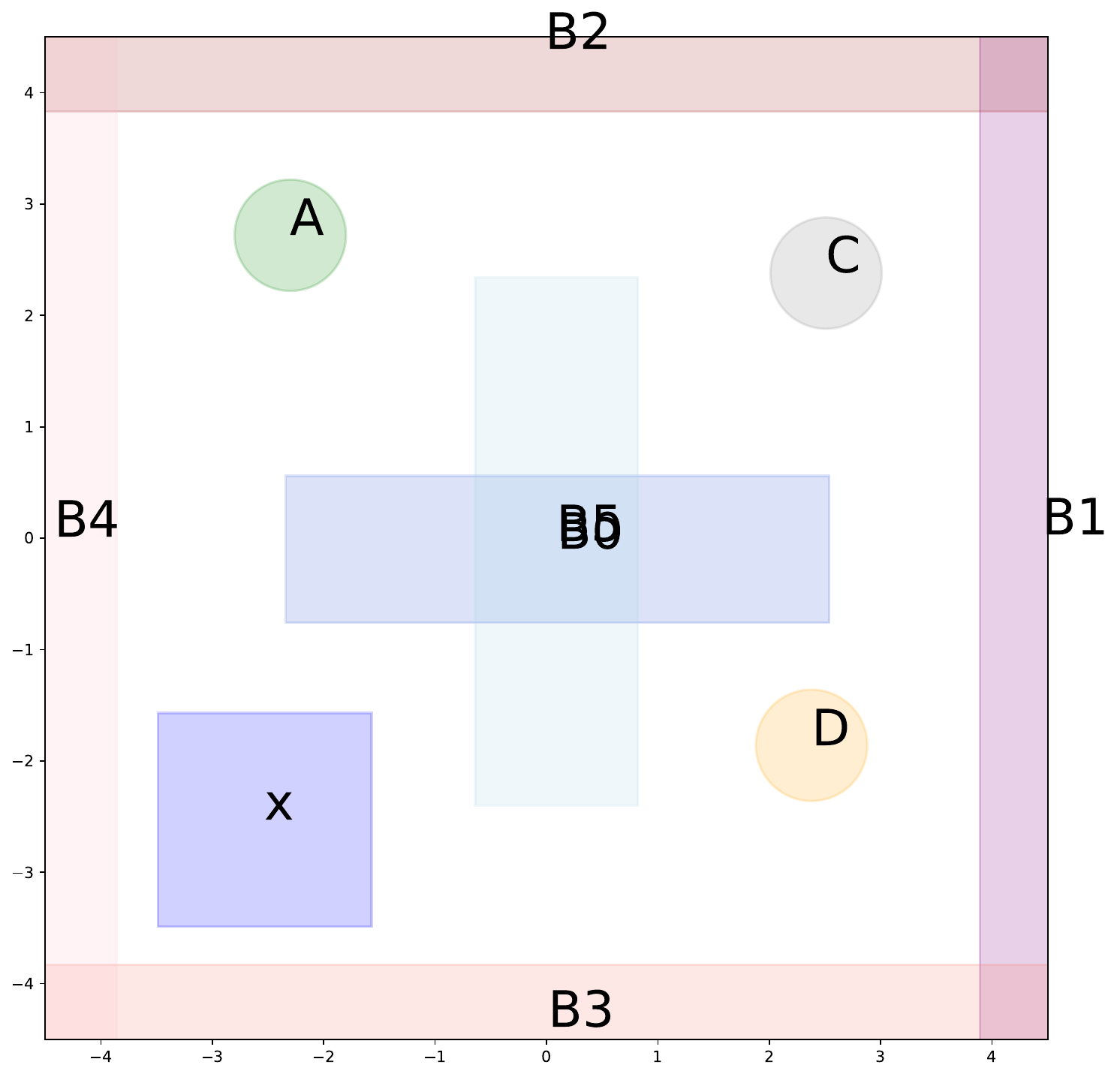}
      \caption{STL-03}
      \label{fig:scene-0lin-task-03}
  \end{subfigure}
  \begin{subfigure}[b]{0.19\textwidth}
      \centering \includegraphics[width=\textwidth]{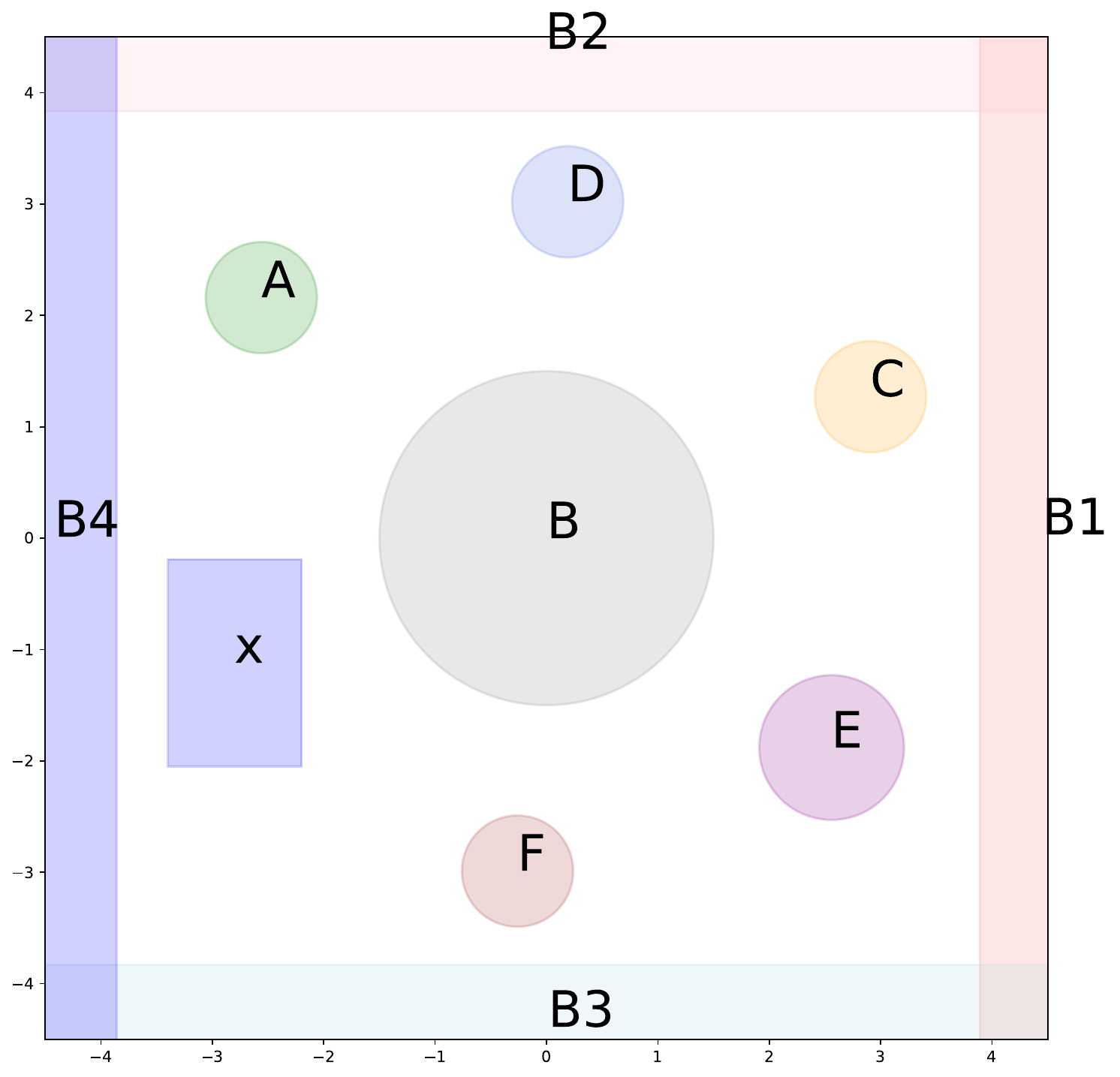}
      \caption{STL-04}
      \label{fig:scene-0lin-task-04}
  \end{subfigure}
  \begin{subfigure}[b]{0.19\textwidth}
      \centering \includegraphics[width=\textwidth]{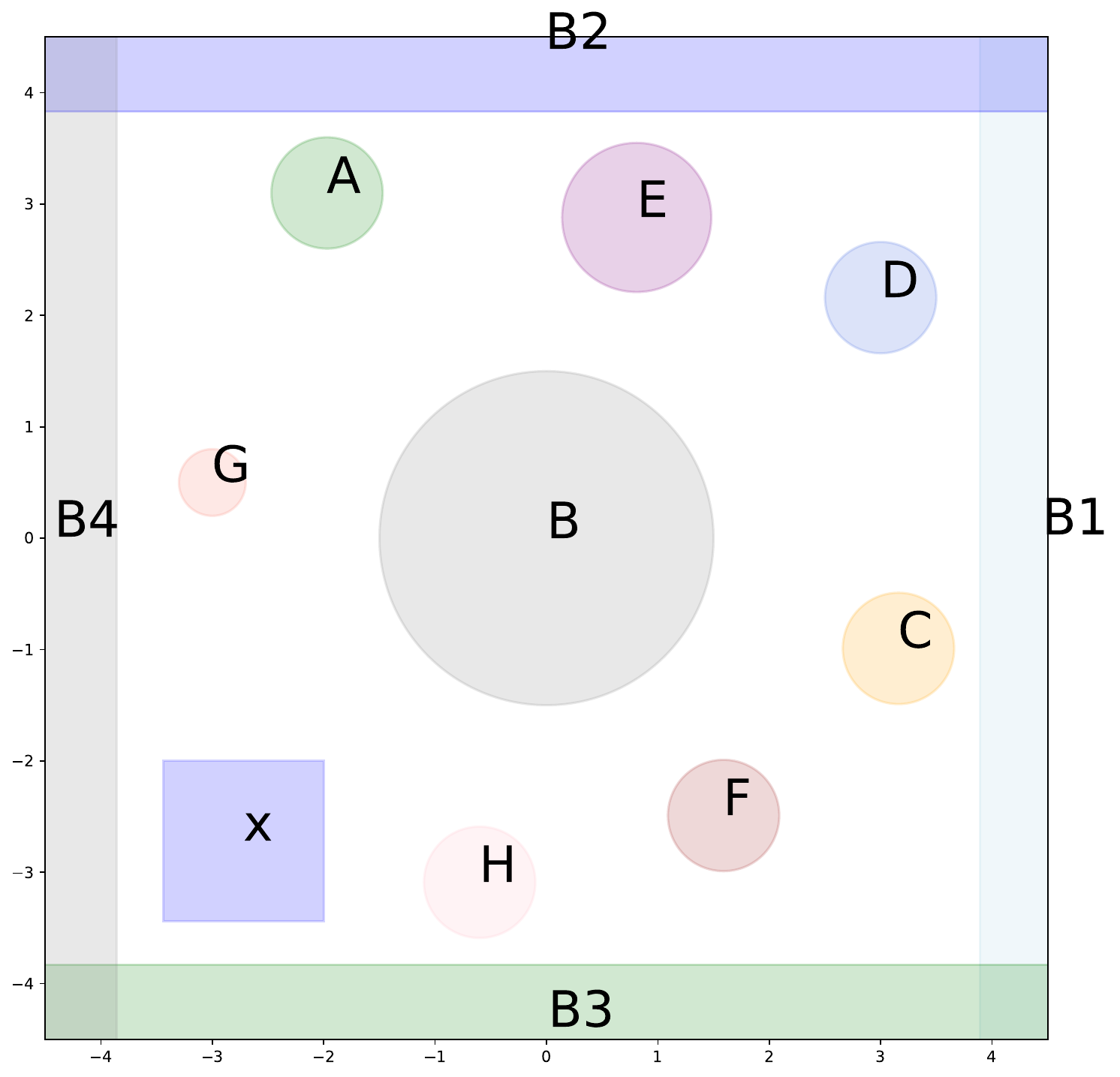}
      \caption{STL-05}
      \label{fig:scene-0lin-task-05}
  \end{subfigure}
  \caption{Scene for Linear: STL tasks 01 to 05}
  \label{fig:scene-cat-0lin-tasks-01-05}
\end{figure}
\noindent \textbf{STL-01 (Two-layer):} \quad $ F_{[5:7]} ( F_{[50:85]} (A) \land G_{[0:90]} (\neg B_5) \land G_{[0:90]} (\neg B_1) \land G_{[0:90]} (\neg B_2) \land G_{[0:90]} (\neg B_3) \land G_{[0:90]} (\neg B_4) ) $

\noindent \textbf{STL-02 (Two-layer):} \quad $ F_{[5:10]} ( F_{[0:50]} (A) \land G_{[60:80]} (C) \land G_{[0:90]} (\neg B_5) \land G_{[0:90]} (\neg B_1) \land G_{[0:90]} (\neg B_2) \land G_{[0:90]} (\neg B_3) \land G_{[0:90]} (\neg B_4) ) $

\noindent \textbf{STL-03 (Two-layer):} \quad $ F_{[5:10]} ( F_{[0:50]} (A) \land F_{[40:60]} (C) \land G_{[70:80]} (D) \land G_{[0:90]} (\neg B_5) \land G_{[0:90]} (\neg B_0) \land G_{[0:90]} (\neg B_1) \land G_{[0:90]} (\neg B_2) \land G_{[0:90]} (\neg B_3) \land G_{[0:90]} (\neg B_4) ) $

\noindent \textbf{STL-04 (Two-layer):} \quad $ F_{[5:10]} ( F_{[0:50]} (A) \land F_{[40:50]} (C) \land F_{[70:80]} (F) \land G_{[50:60]} (D) \land G_{[0:90]} (\neg B) \land G_{[0:90]} (\neg E) \land G_{[0:90]} (\neg B_1) \land G_{[0:90]} (\neg B_2) \land G_{[0:90]} (\neg B_3) \land G_{[0:90]} (\neg B_4) ) $

\noindent \textbf{STL-05 (Two-layer):} \quad $ F_{[5:10]} ( F_{[0:30]} (A) \land F_{[30:50]} (C) \land F_{[70:80]} (F) \land F_{[75:88]} (H) \land G_{[50:60]} (D) \land G_{[0:90]} (\neg B) \land G_{[0:90]} (\neg E) \land G_{[0:90]} (\neg G) \land G_{[0:90]} (\neg B_1) \land G_{[0:90]} (\neg B_2) \land G_{[0:90]} (\neg B_3) \land G_{[0:90]} (\neg B_4) ) $

\begin{figure}[!htbp]
  \centering
  \begin{subfigure}[b]{0.19\textwidth}
      \centering \includegraphics[width=\textwidth]{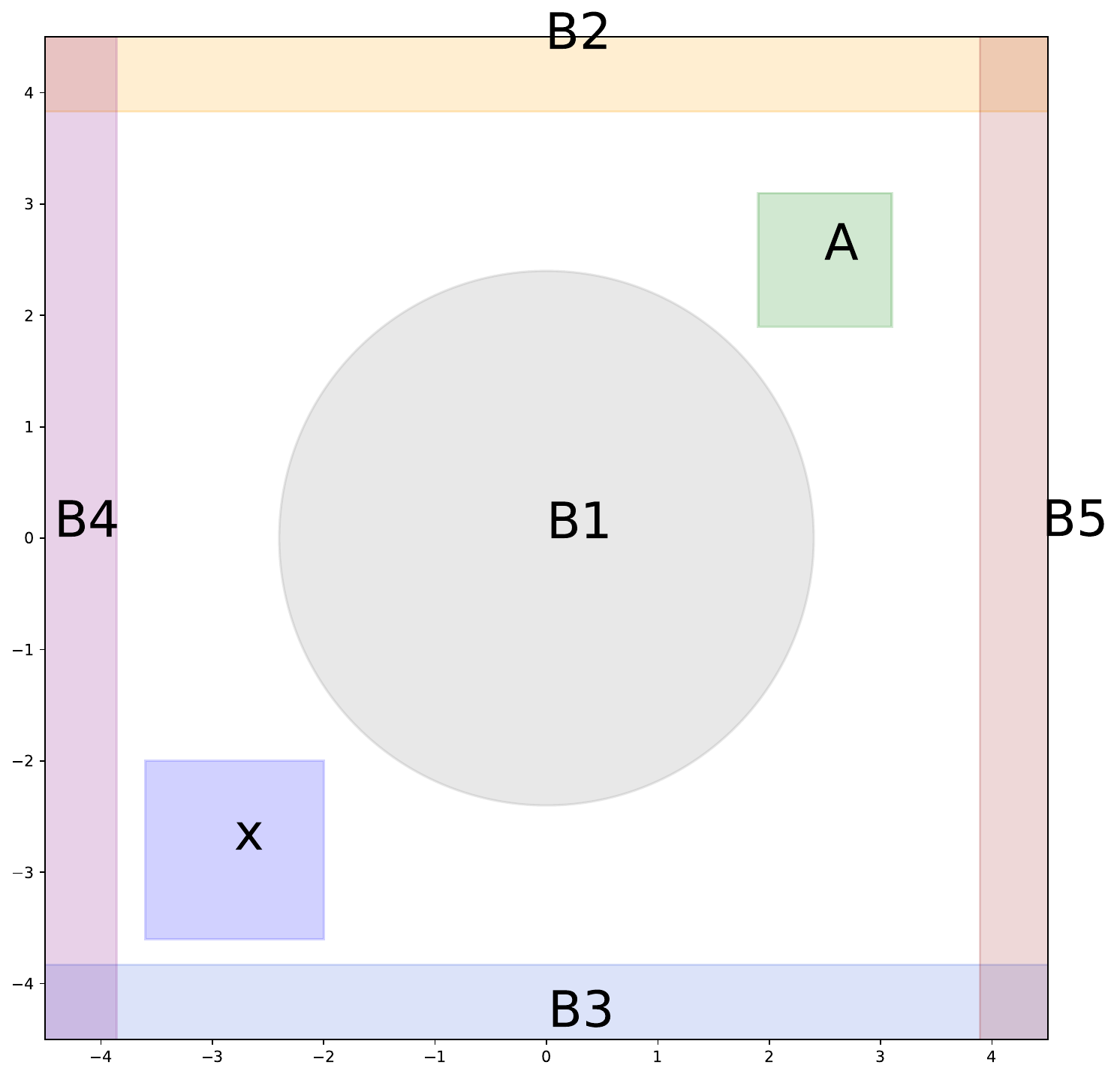}
      \caption{STL-06}
      \label{fig:scene-0lin-task-06}
  \end{subfigure}
  \begin{subfigure}[b]{0.19\textwidth}
      \centering \includegraphics[width=\textwidth]{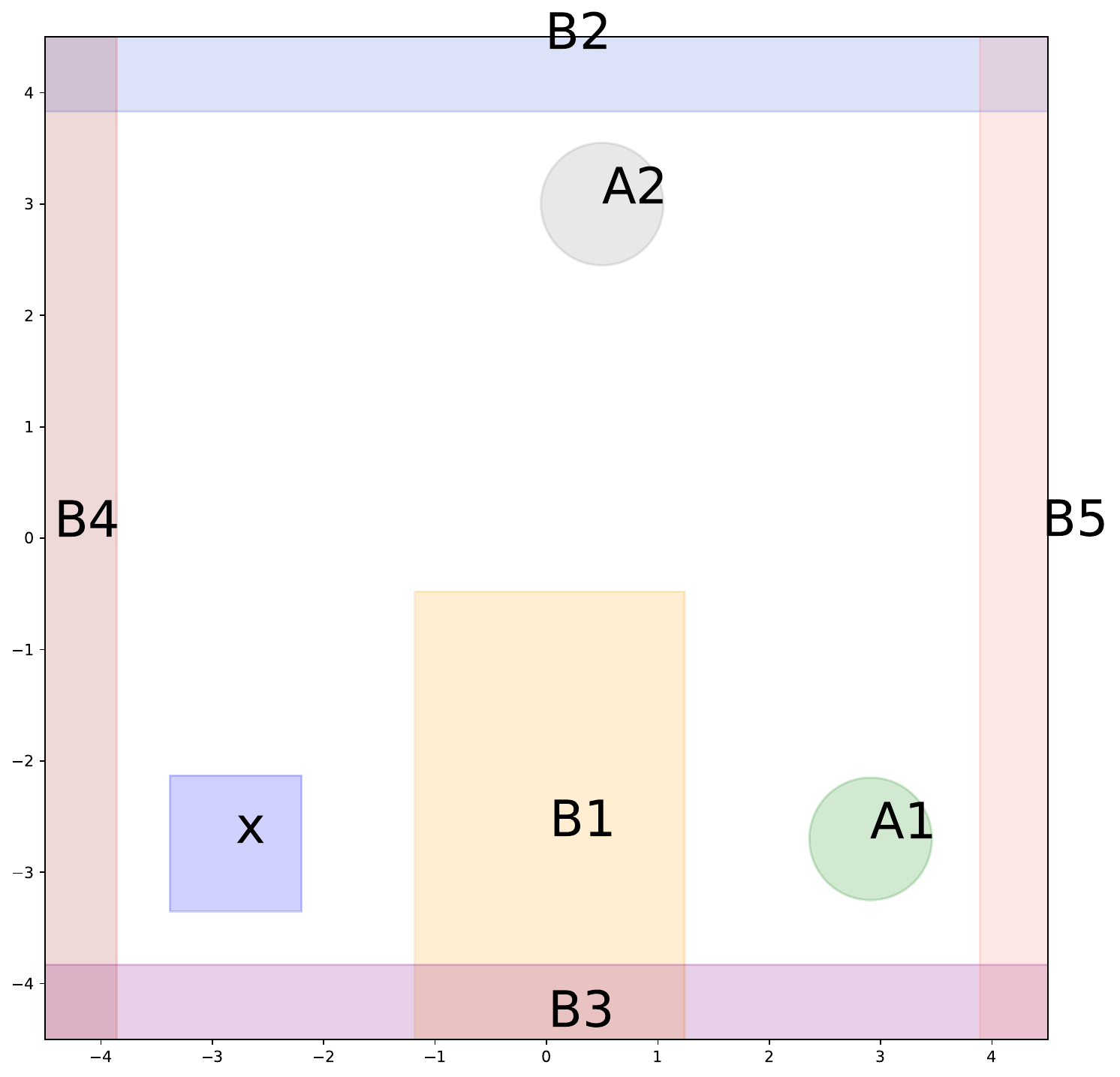}
      \caption{STL-07}
      \label{fig:scene-0lin-task-07}
  \end{subfigure}
  \begin{subfigure}[b]{0.19\textwidth}
      \centering \includegraphics[width=\textwidth]{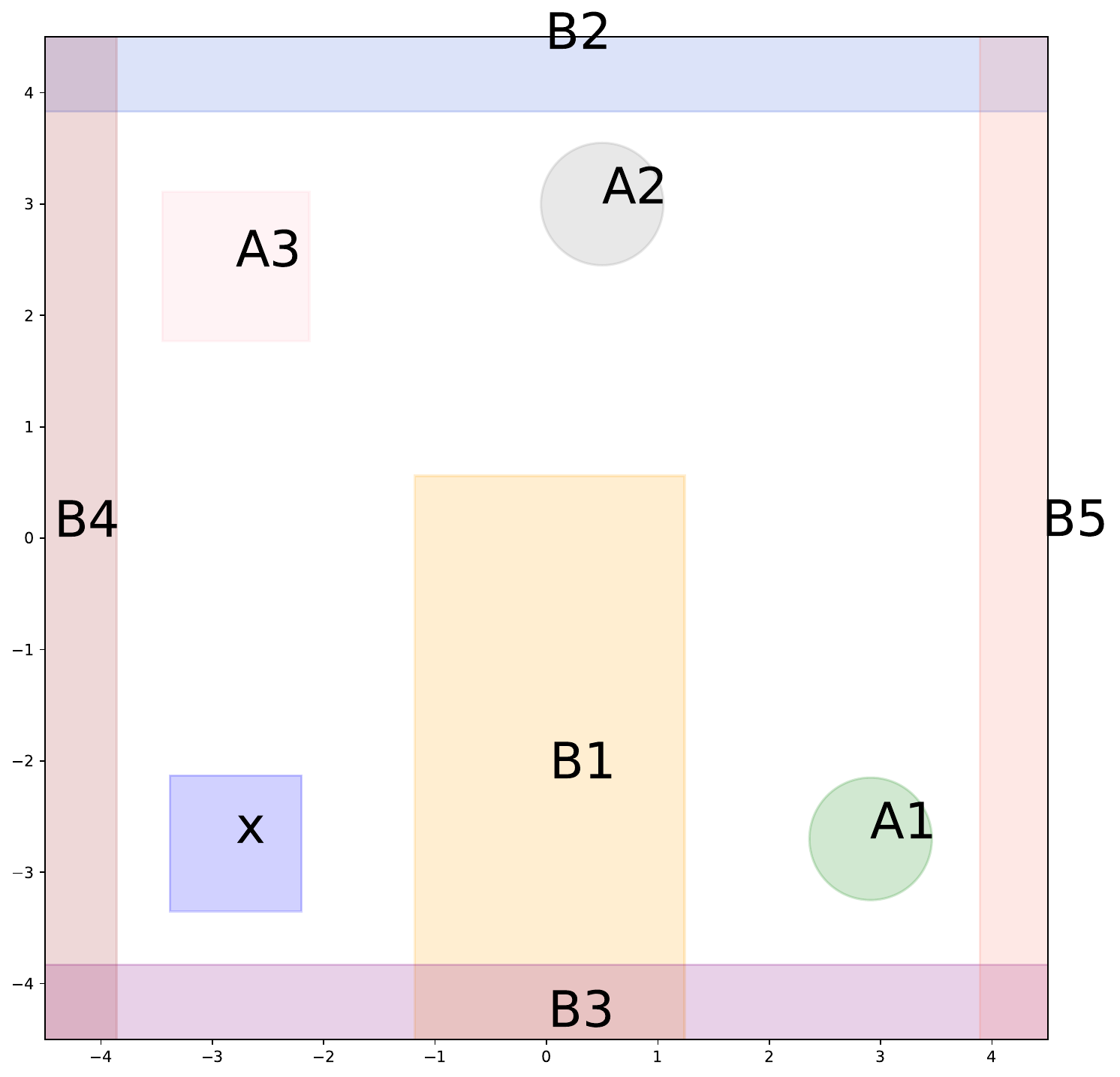}
      \caption{STL-08}
      \label{fig:scene-0lin-task-08}
  \end{subfigure}
  \begin{subfigure}[b]{0.19\textwidth}
      \centering \includegraphics[width=\textwidth]{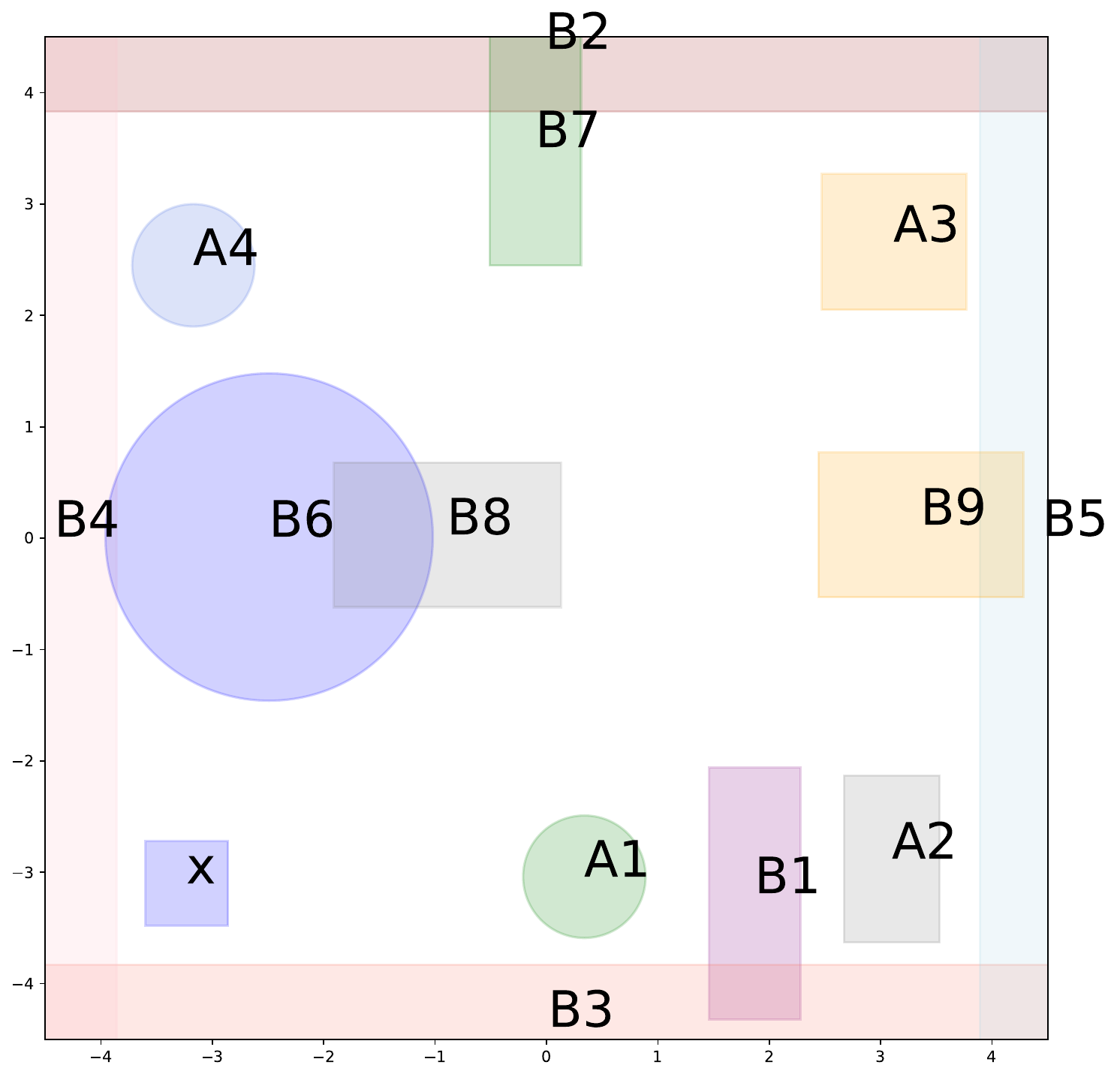}
      \caption{STL-09}
      \label{fig:scene-0lin-task-09}
  \end{subfigure}
  \begin{subfigure}[b]{0.19\textwidth}
      \centering \includegraphics[width=\textwidth]{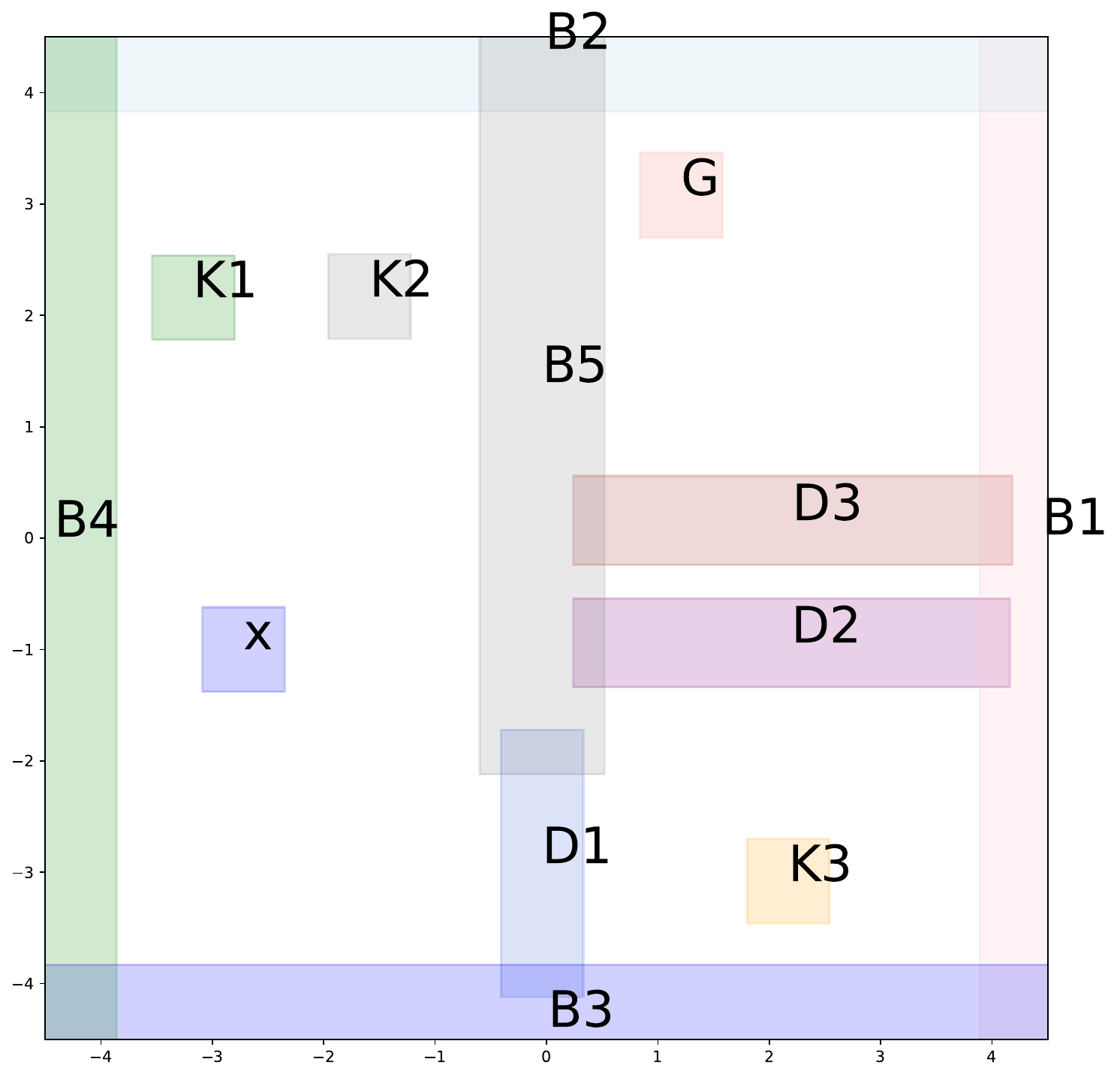}
      \caption{STL-10}
      \label{fig:scene-0lin-task-10}
  \end{subfigure}
  \caption{Scene for Linear: STL tasks 06 to 10}
  \label{fig:scene-cat-0lin-tasks-06-10}
\end{figure}
\noindent \textbf{STL-06 (Multi-layer):} \quad $  F_{[10:90]} (A) \land G_{[0:100]} (\neg B_1) \land G_{[0:100]} (\neg B_2) \land G_{[0:100]} (\neg B_3) \land G_{[0:100]} (\neg B_4) \land G_{[0:100]} (\neg B_5)  $

\noindent \textbf{STL-07 (Multi-layer):} \quad $  F_{[0:90]} (A_1) \land F_{[40:80]} (A_2) \land G_{[0:100]} (\neg B_1) \land G_{[0:100]} (\neg B_2) \land G_{[0:100]} (\neg B_3) \land G_{[0:100]} (\neg B_4) \land G_{[0:100]} (\neg B_5)  $

\noindent \textbf{STL-08 (Multi-layer):} \quad $  F_{[0:90]} (A_1) \land F_{[40:80]} ( A_2 \land F_{[10:20]} (G_{[0:10]} (A_3)) ) \land G_{[0:100]} (\neg B_1) \land G_{[0:100]} (\neg B_2) \land G_{[0:100]} (\neg B_3) \land G_{[0:100]} (\neg B_4) \land G_{[0:100]} (\neg B_5)  $

\noindent \textbf{STL-09 (Multi-layer):} \quad $  F_{[5:20]} ( A_1 \land F_{[10:20]} ( G_{[0:5]} (A_2) \land  F_{[10:30]} (G_{[0:5]} (A_3)) \land F_{[10:30]} (G_{[0:10]} (A_4))  ) ) \land G_{[0:100]} (\neg B_1) \land G_{[0:100]} (\neg B_2) \land G_{[0:100]} (\neg B_3) \land G_{[0:100]} (\neg B_4) \land G_{[0:100]} (\neg B_5) \land G_{[0:100]} (\neg B_6) \land G_{[0:100]} (\neg B_7) \land G_{[0:100]} (\neg B_8) \land G_{[0:100]} (\neg B_9)  $

\noindent \textbf{STL-10 (Multi-layer):} \quad $  (\neg D_1)U_{[0:100]}(K_1) \land (\neg D_2)U_{[0:100]}(K_2) \land (\neg D_3)U_{[0:100]}(K_3) \land F_{[80:90]} (G_{[0:5]} (G)) \land G_{[0:100]} (\neg B_1) \land G_{[0:100]} (\neg B_2) \land G_{[0:100]} (\neg B_3) \land G_{[0:100]} (\neg B_4) \land G_{[0:100]} (\neg B_5)  $

\subsubsection{STLs in ``Unicycle" environment}

\begin{figure}[!htbp]
  \centering
  \begin{subfigure}[b]{0.19\textwidth}
      \centering \includegraphics[width=\textwidth]{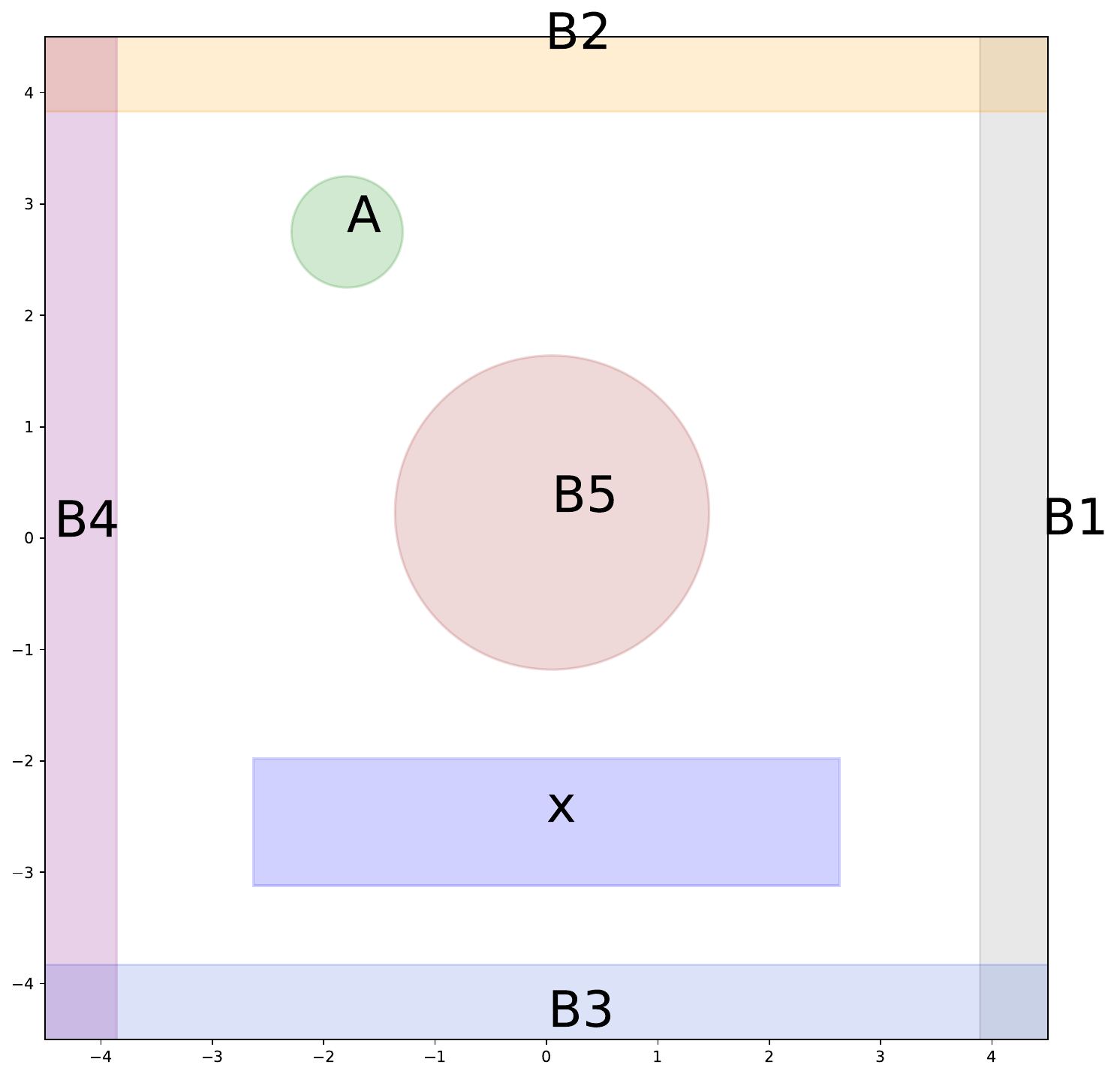}
      \caption{STL-01}
      \label{fig:scene-1car-task-01}
  \end{subfigure}
  \begin{subfigure}[b]{0.19\textwidth}
      \centering \includegraphics[width=\textwidth]{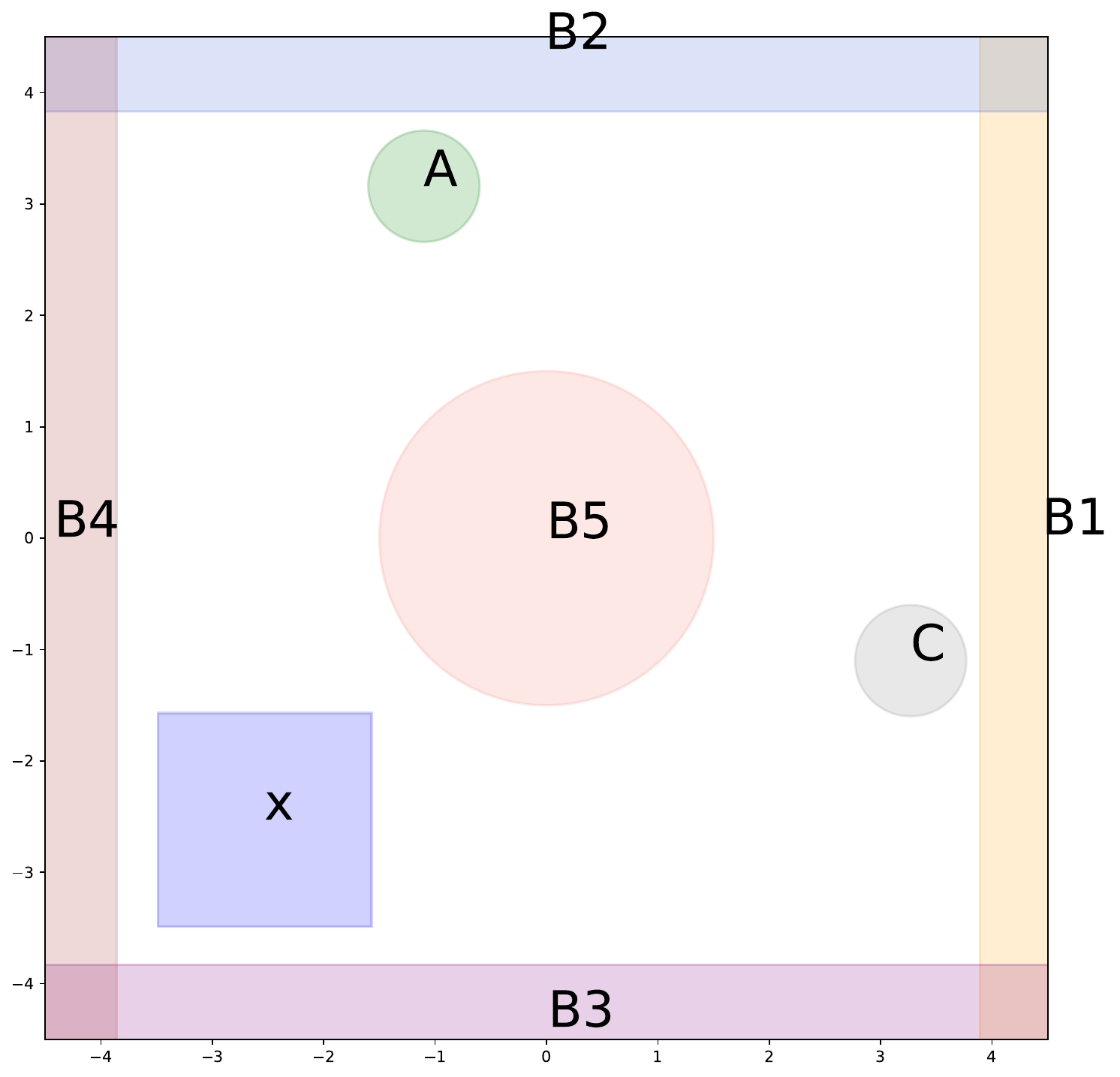}
      \caption{STL-02}
      \label{fig:scene-1car-task-02}
  \end{subfigure}
  \begin{subfigure}[b]{0.19\textwidth}
      \centering \includegraphics[width=\textwidth]{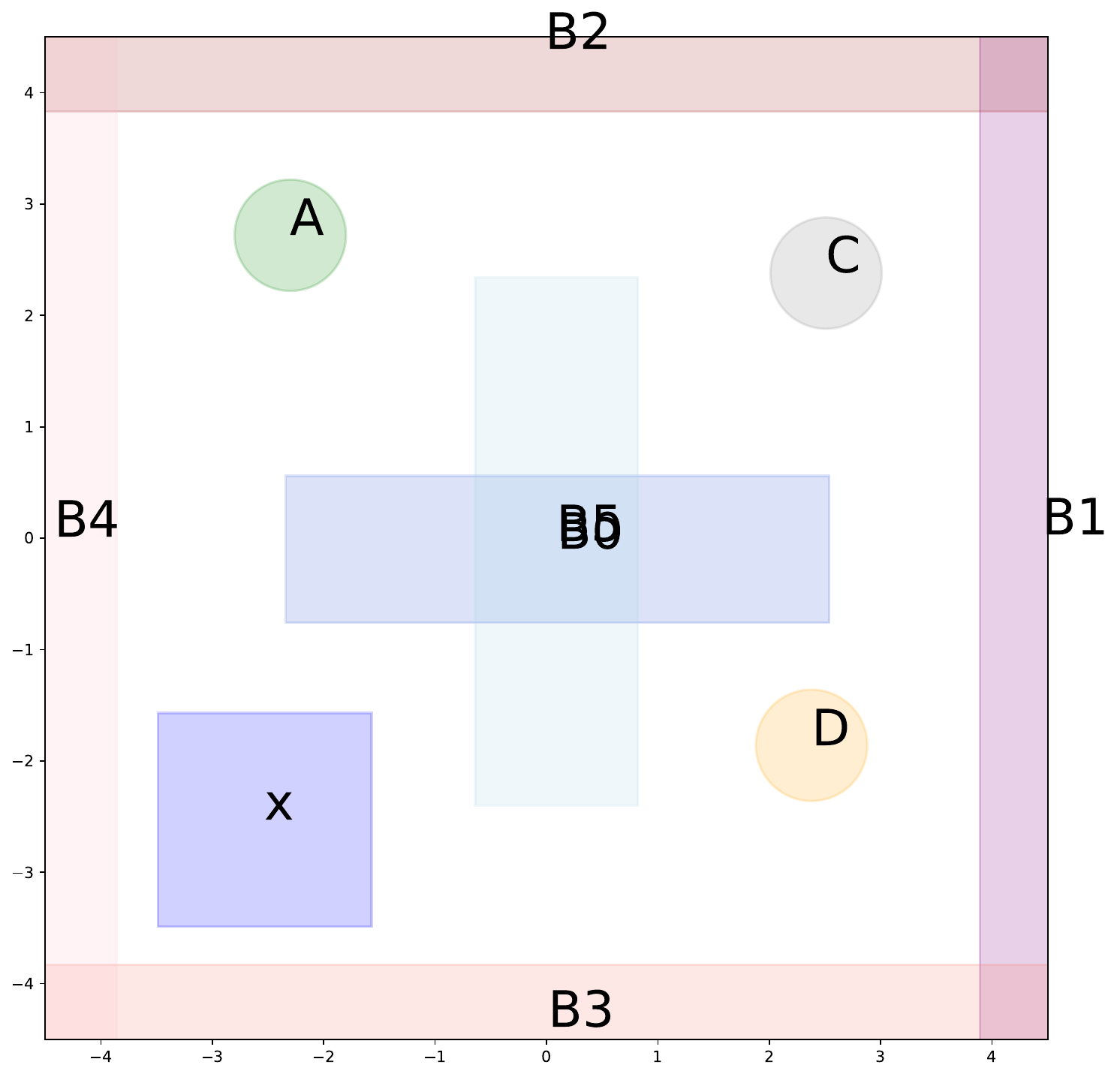}
      \caption{STL-03}
      \label{fig:scene-1car-task-03}
  \end{subfigure}
  \begin{subfigure}[b]{0.19\textwidth}
      \centering \includegraphics[width=\textwidth]{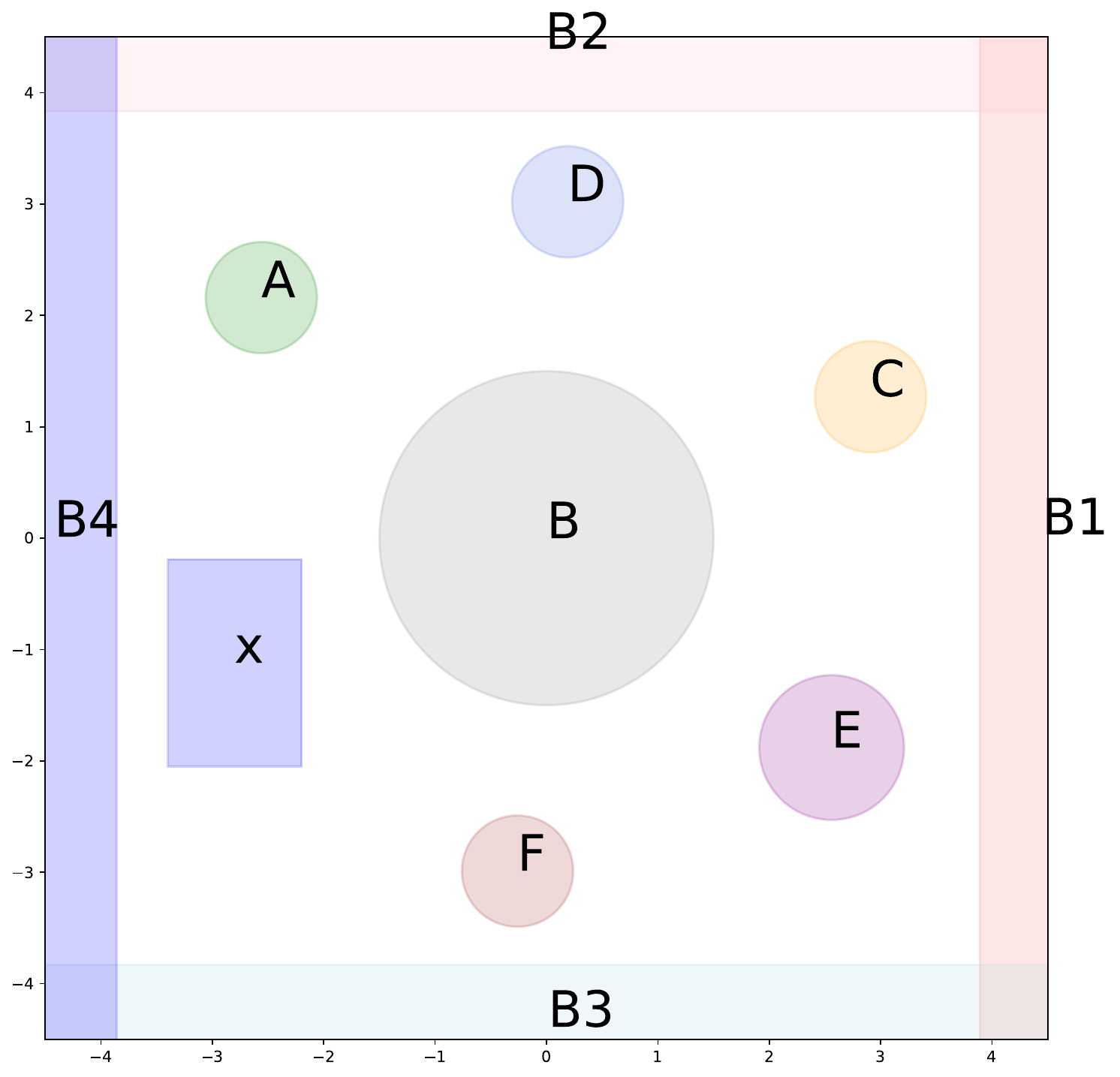}
      \caption{STL-04}
      \label{fig:scene-1car-task-04}
  \end{subfigure}
  \begin{subfigure}[b]{0.19\textwidth}
      \centering \includegraphics[width=\textwidth]{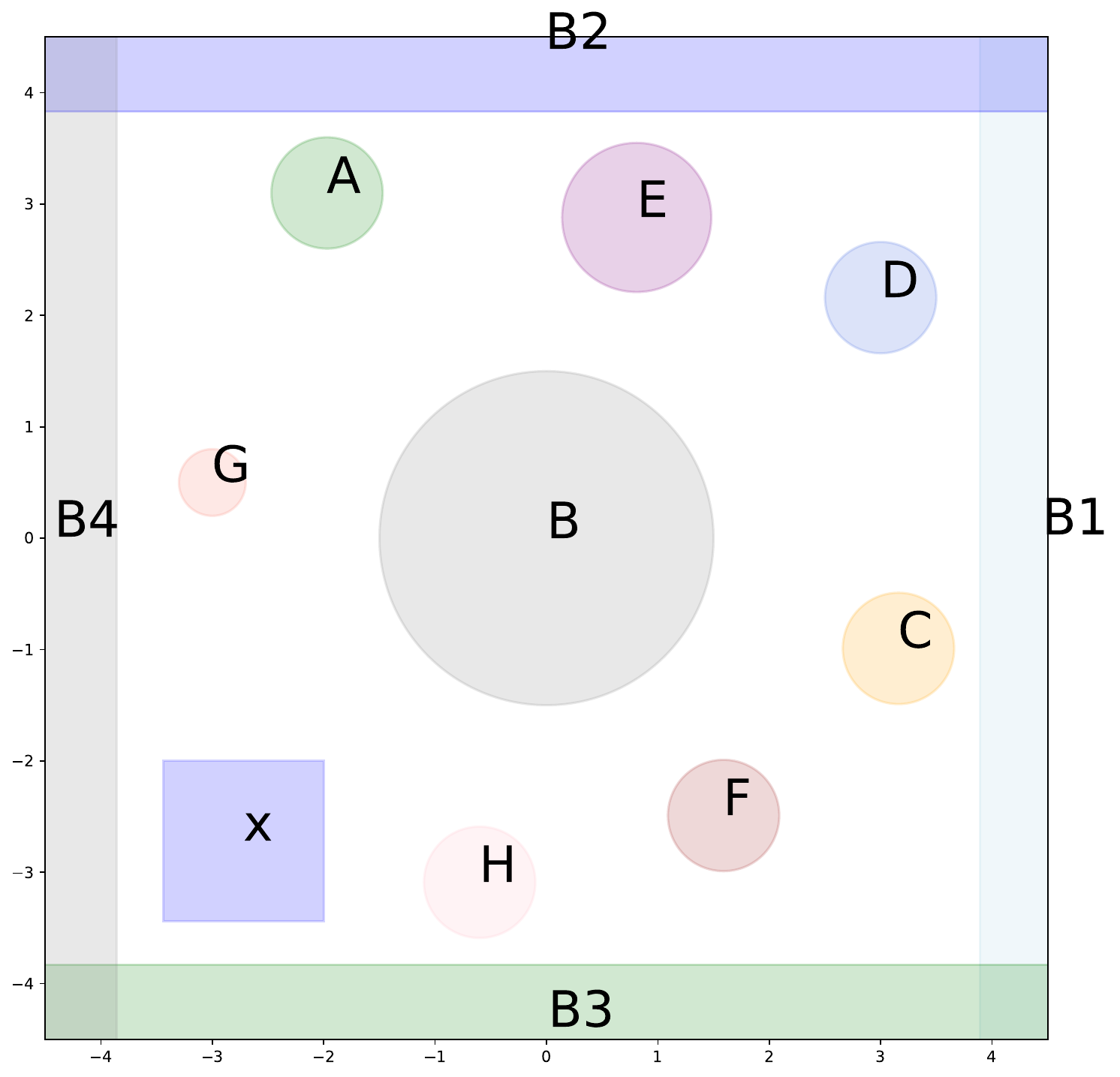}
      \caption{STL-05}
      \label{fig:scene-1car-task-05}
  \end{subfigure}
  \caption{Scene for Unicycle: STL tasks 01 to 05}
  \label{fig:scene-cat-1car-tasks-01-05}
\end{figure}
\noindent \textbf{STL-01 (Two-layer):} \quad $ F_{[5:7]} ( F_{[50:85]} (A) \land G_{[0:90]} (\neg B_5) \land G_{[0:90]} (\neg B_1) \land G_{[0:90]} (\neg B_2) \land G_{[0:90]} (\neg B_3) \land G_{[0:90]} (\neg B_4) ) $

\noindent \textbf{STL-02 (Two-layer):} \quad $ F_{[5:10]} ( F_{[0:50]} (A) \land G_{[60:80]} (C) \land G_{[0:90]} (\neg B_5) \land G_{[0:90]} (\neg B_1) \land G_{[0:90]} (\neg B_2) \land G_{[0:90]} (\neg B_3) \land G_{[0:90]} (\neg B_4) ) $

\noindent \textbf{STL-03 (Two-layer):} \quad $ F_{[5:10]} ( F_{[0:50]} (A) \land F_{[40:60]} (C) \land G_{[70:80]} (D) \land G_{[0:90]} (\neg B_5) \land G_{[0:90]} (\neg B_0) \land G_{[0:90]} (\neg B_1) \land G_{[0:90]} (\neg B_2) \land G_{[0:90]} (\neg B_3) \land G_{[0:90]} (\neg B_4) ) $

\noindent \textbf{STL-04 (Two-layer):} \quad $ F_{[5:10]} ( F_{[0:50]} (A) \land F_{[40:50]} (C) \land F_{[70:80]} (F) \land G_{[50:60]} (D) \land G_{[0:90]} (\neg B) \land G_{[0:90]} (\neg E) \land G_{[0:90]} (\neg B_1) \land G_{[0:90]} (\neg B_2) \land G_{[0:90]} (\neg B_3) \land G_{[0:90]} (\neg B_4) ) $

\noindent \textbf{STL-05 (Two-layer):} \quad $ F_{[5:10]} ( F_{[0:30]} (A) \land F_{[30:50]} (C) \land F_{[70:80]} (F) \land F_{[75:88]} (H) \land G_{[50:60]} (D) \land G_{[0:90]} (\neg B) \land G_{[0:90]} (\neg E) \land G_{[0:90]} (\neg G) \land G_{[0:90]} (\neg B_1) \land G_{[0:90]} (\neg B_2) \land G_{[0:90]} (\neg B_3) \land G_{[0:90]} (\neg B_4) ) $

\begin{figure}[!htbp]
  \centering
  \begin{subfigure}[b]{0.19\textwidth}
      \centering \includegraphics[width=\textwidth]{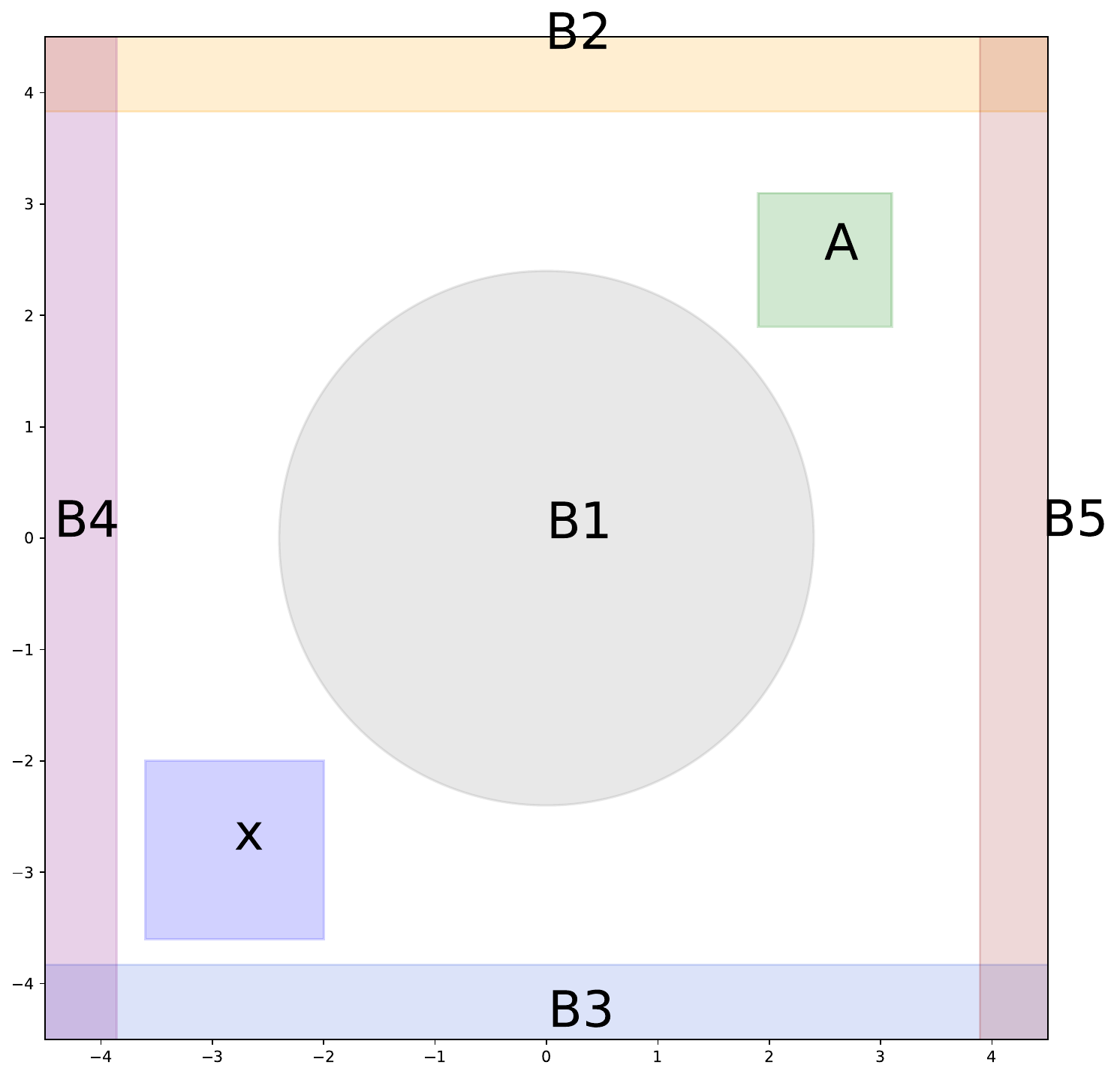}
      \caption{STL-06}
      \label{fig:scene-1car-task-06}
  \end{subfigure}
  \begin{subfigure}[b]{0.19\textwidth}
      \centering \includegraphics[width=\textwidth]{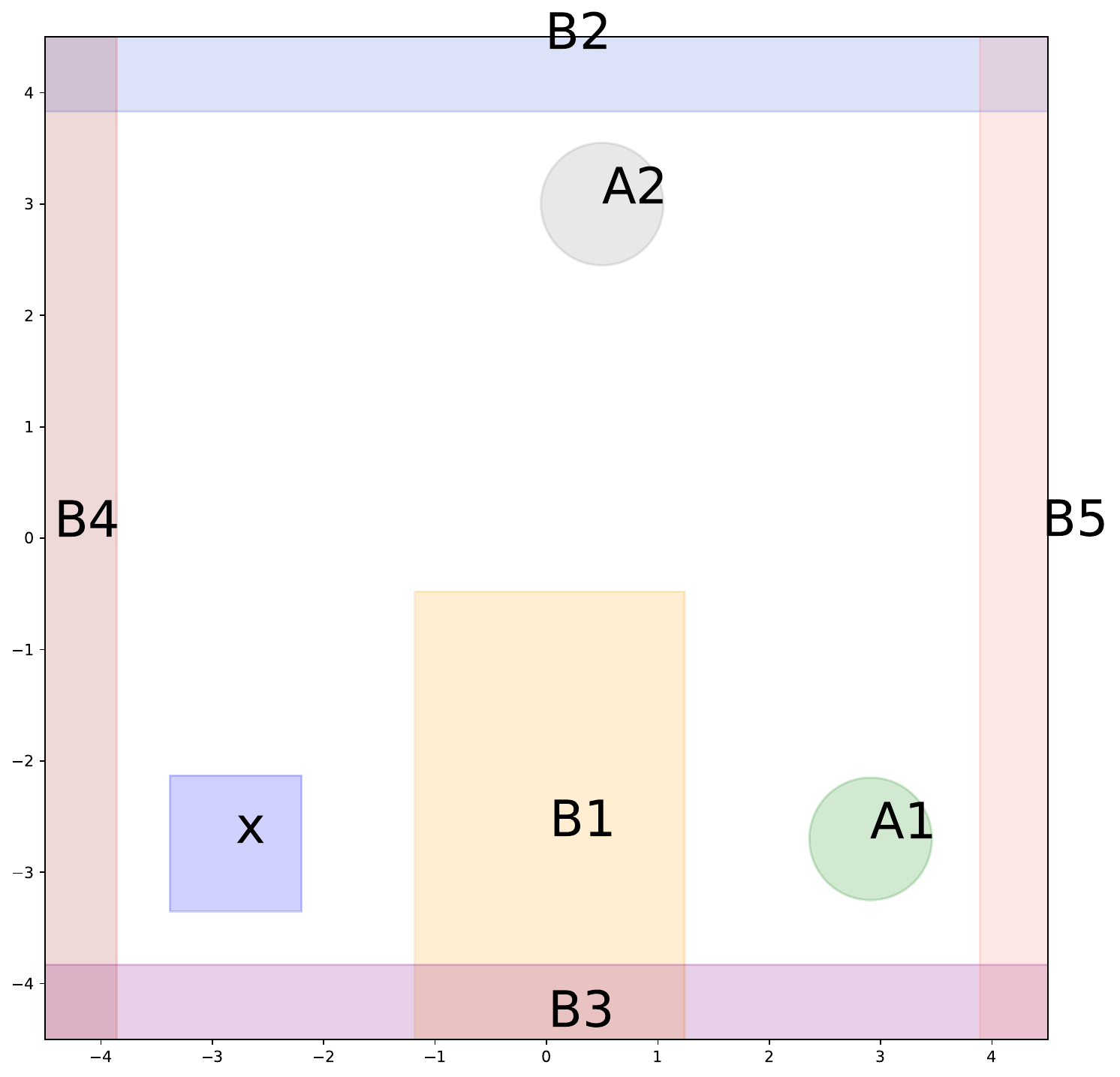}
      \caption{STL-07}
      \label{fig:scene-1car-task-07}
  \end{subfigure}
  \begin{subfigure}[b]{0.19\textwidth}
      \centering \includegraphics[width=\textwidth]{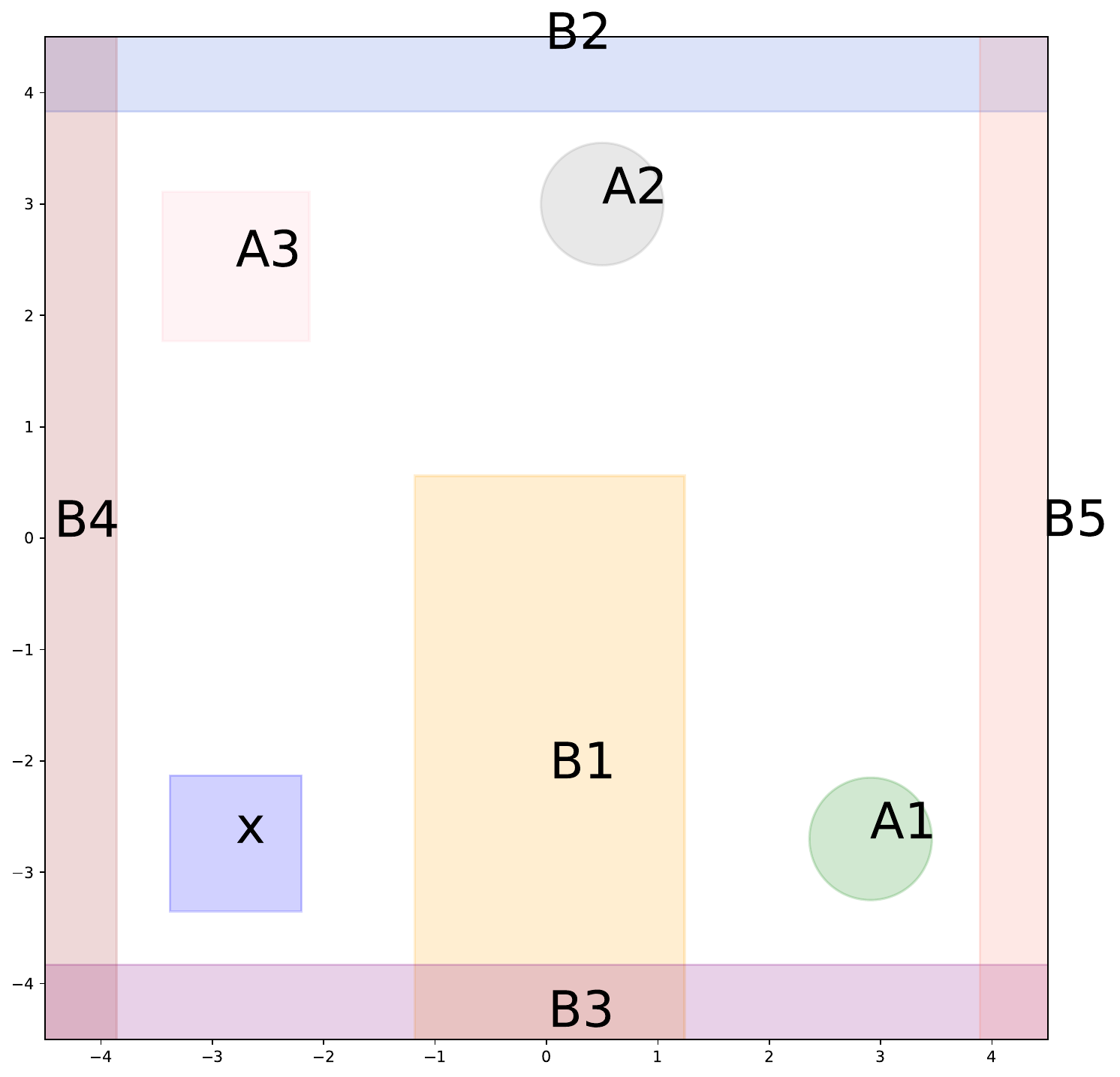}
      \caption{STL-08}
      \label{fig:scene-1car-task-08}
  \end{subfigure}
  \begin{subfigure}[b]{0.19\textwidth}
      \centering \includegraphics[width=\textwidth]{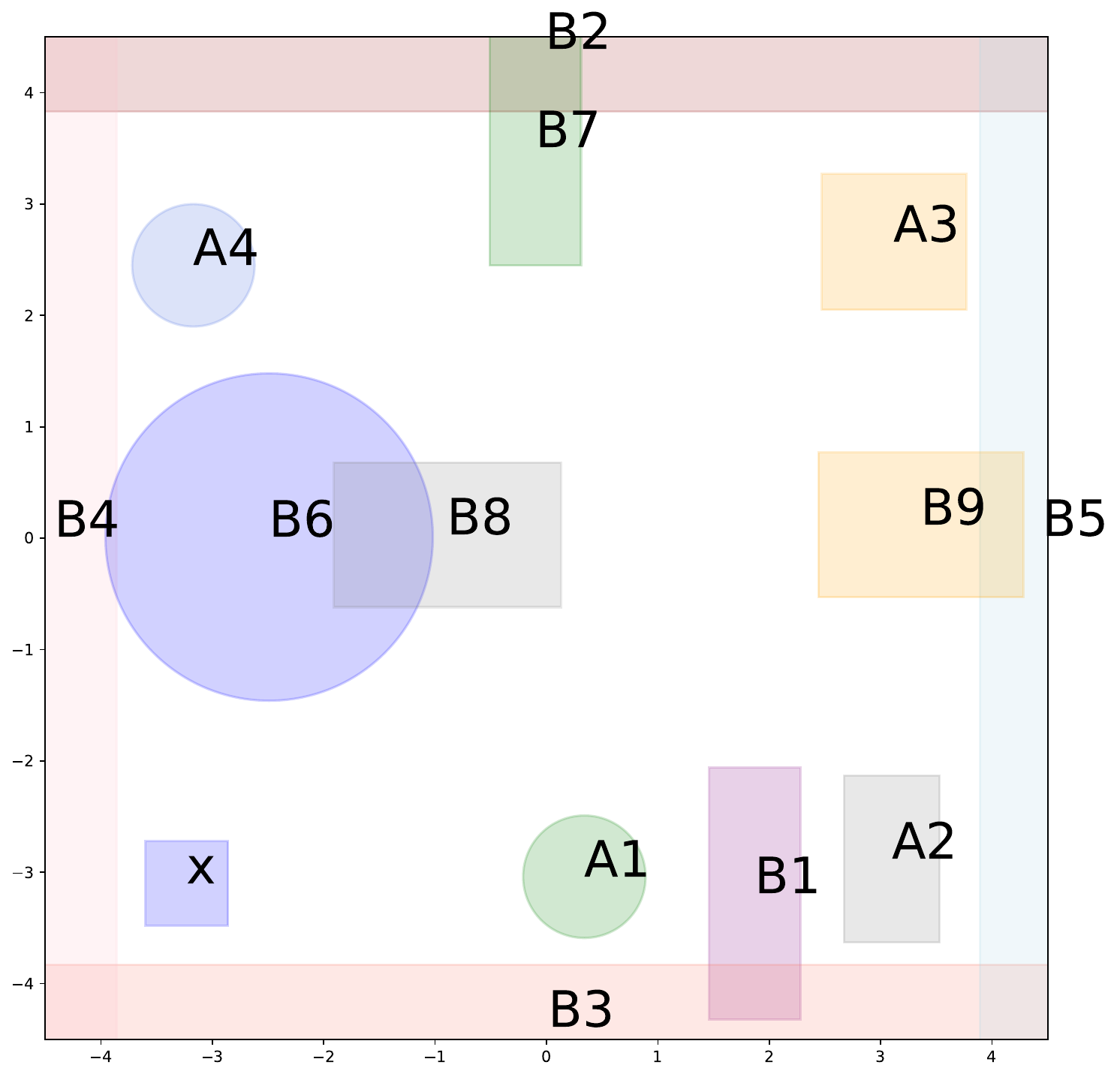}
      \caption{STL-09}
      \label{fig:scene-1car-task-09}
  \end{subfigure}
  \begin{subfigure}[b]{0.19\textwidth}
      \centering \includegraphics[width=\textwidth]{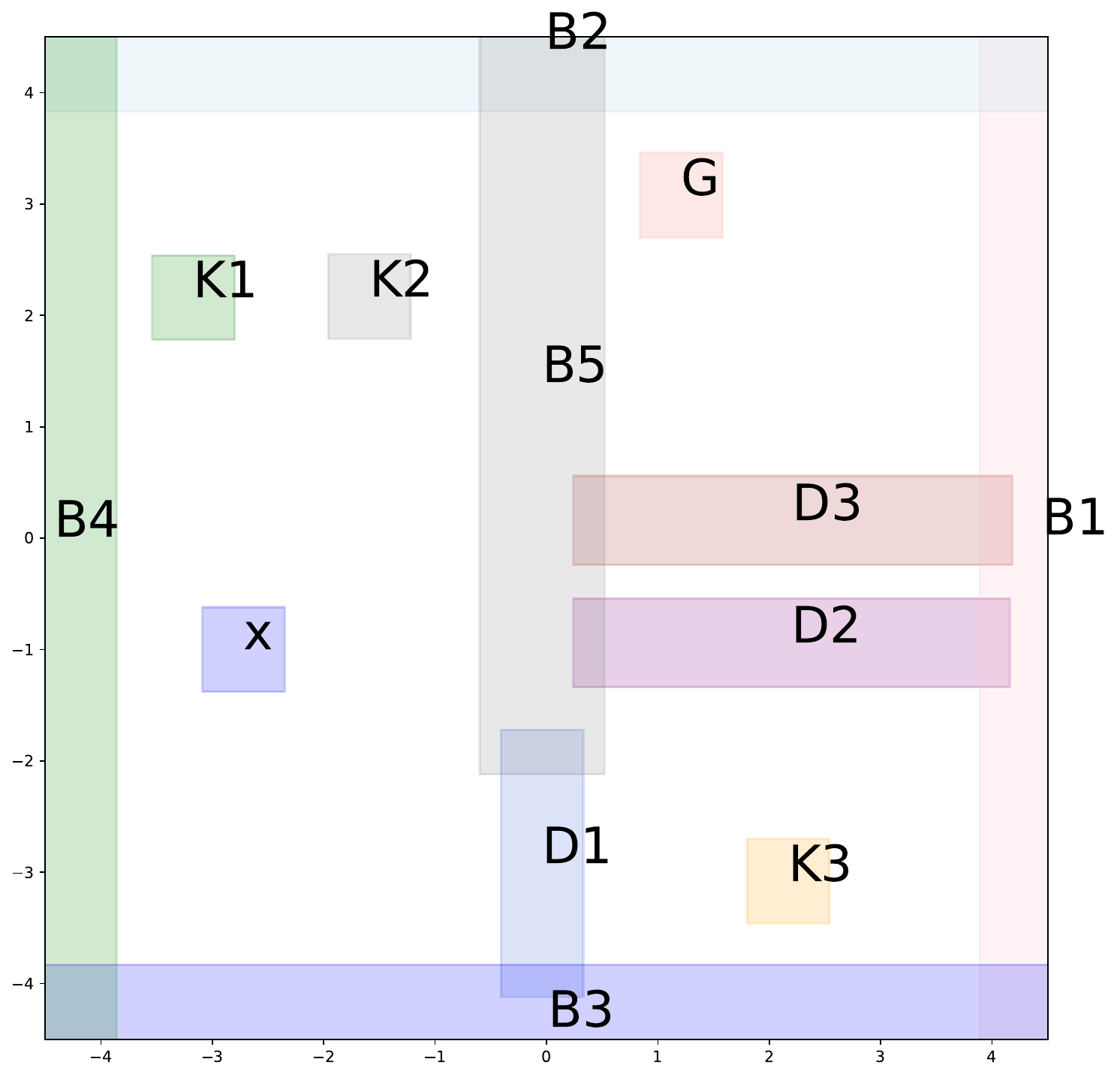}
      \caption{STL-10}
      \label{fig:scene-1car-task-10}
  \end{subfigure}
  \caption{Scene for Unicycle: STL tasks 06 to 10}
  \label{fig:scene-cat-1car-tasks-06-10}
\end{figure}
\noindent \textbf{STL-06 (Multi-layer):} \quad $  F_{[10:90]} (A) \land G_{[0:100]} (\neg B_1) \land G_{[0:100]} (\neg B_2) \land G_{[0:100]} (\neg B_3) \land G_{[0:100]} (\neg B_4) \land G_{[0:100]} (\neg B_5)  $

\noindent \textbf{STL-07 (Multi-layer):} \quad $  F_{[0:90]} (A_1) \land F_{[40:80]} (A_2) \land G_{[0:100]} (\neg B_1) \land G_{[0:100]} (\neg B_2) \land G_{[0:100]} (\neg B_3) \land G_{[0:100]} (\neg B_4) \land G_{[0:100]} (\neg B_5)  $

\noindent \textbf{STL-08 (Multi-layer):} \quad $  F_{[0:90]} (A_1) \land F_{[40:80]} ( A_2 \land F_{[10:20]} (G_{[0:10]} (A_3)) ) \land G_{[0:100]} (\neg B_1) \land G_{[0:100]} (\neg B_2) \land G_{[0:100]} (\neg B_3) \land G_{[0:100]} (\neg B_4) \land G_{[0:100]} (\neg B_5)  $

\noindent \textbf{STL-09 (Multi-layer):} \quad $  F_{[5:20]} ( A_1 \land F_{[10:20]} ( G_{[0:5]} (A_2) \land  F_{[10:30]} (G_{[0:5]} (A_3)) \land F_{[10:30]} (G_{[0:10]} (A_4))  ) ) \land G_{[0:100]} (\neg B_1) \land G_{[0:100]} (\neg B_2) \land G_{[0:100]} (\neg B_3) \land G_{[0:100]} (\neg B_4) \land G_{[0:100]} (\neg B_5) \land G_{[0:100]} (\neg B_6) \land G_{[0:100]} (\neg B_7) \land G_{[0:100]} (\neg B_8) \land G_{[0:100]} (\neg B_9)  $

\noindent \textbf{STL-10 (Multi-layer):} \quad $  (\neg D_1)U_{[0:100]}(K_1) \land (\neg D_2)U_{[0:100]}(K_2) \land (\neg D_3)U_{[0:100]}(K_3) \land F_{[80:90]} (G_{[0:5]} (G)) \land G_{[0:100]} (\neg B_1) \land G_{[0:100]} (\neg B_2) \land G_{[0:100]} (\neg B_3) \land G_{[0:100]} (\neg B_4) \land G_{[0:100]} (\neg B_5)  $

\subsubsection{STLs in ``Franka Panda" environment}

\begin{figure}[!htbp]
  \centering \includegraphics[width=0.6\textwidth]{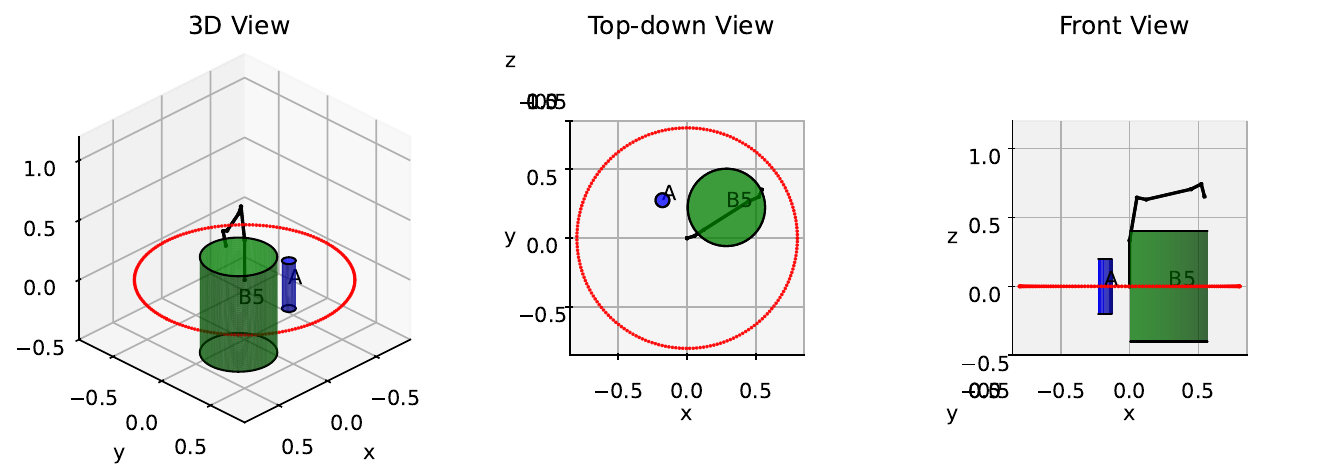}
  \caption{Scene for Franka Panda: STL task 01}
  \label{fig:scene-2arm-task-01}
\end{figure}
\noindent \textbf{STL-01 (Two-layer):} \quad $ F_{[5:7]} ( F_{[50:85]} (A) \land G_{[0:90]} (\neg B_5) \land G_{[0:100]} (\neg W_1) \land G_{[0:100]} (\neg W_2) \land G_{[0:100]} (\neg W_3) \land G_{[0:100]} (\neg W_4) \land G_{[0:100]} (\neg W_5) \land G_{[0:100]} (\neg W_6) ) $

\begin{figure}[!htbp]
  \centering \includegraphics[width=0.6\textwidth]{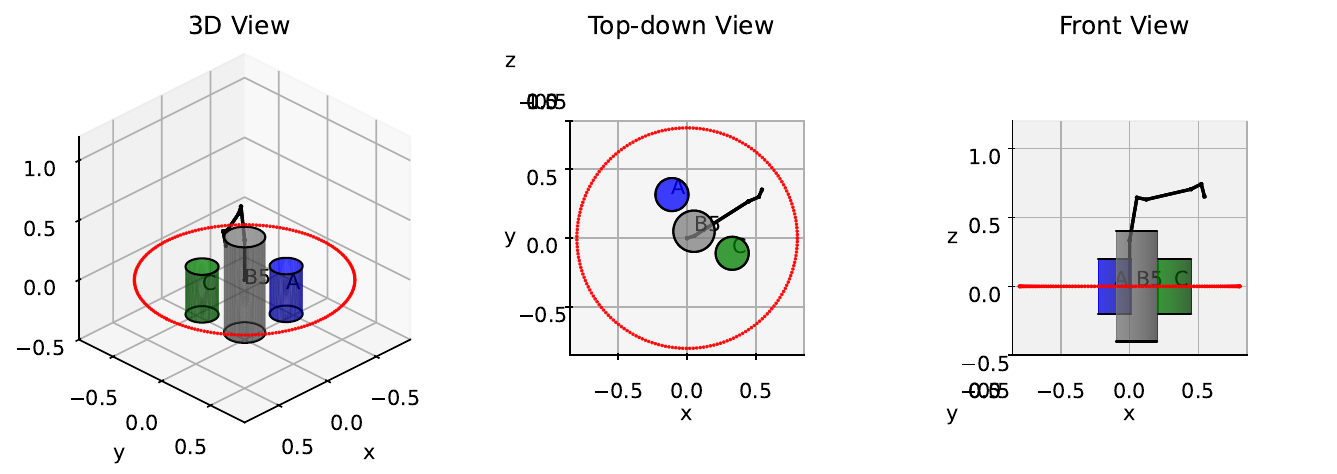}
  \caption{Scene for Franka Panda: STL task 02}
  \label{fig:scene-2arm-task-02}
\end{figure}
\noindent \textbf{STL-02 (Two-layer):} \quad $ F_{[5:10]} ( F_{[0:50]} (A) \land G_{[60:80]} (C) \land G_{[0:90]} (\neg B_5) \land G_{[0:100]} (\neg W_1) \land G_{[0:100]} (\neg W_2) \land G_{[0:100]} (\neg W_3) \land G_{[0:100]} (\neg W_4) \land G_{[0:100]} (\neg W_5) \land G_{[0:100]} (\neg W_6) ) $

\begin{figure}[!htbp]
  \centering \includegraphics[width=0.6\textwidth]{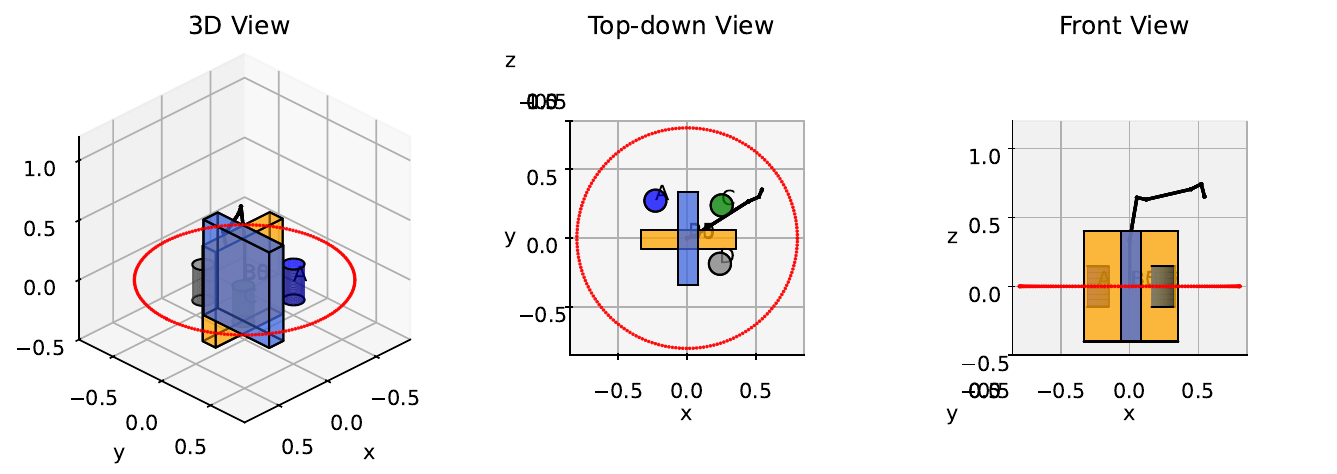}
  \caption{Scene for Franka Panda: STL task 03}
  \label{fig:scene-2arm-task-03}
\end{figure}
\noindent \textbf{STL-03 (Two-layer):} \quad $ F_{[5:10]} ( F_{[0:50]} (A) \land F_{[40:60]} (C) \land G_{[70:80]} (D) \land G_{[0:90]} (\neg B_5) \land G_{[0:90]} (\neg B_0) \land G_{[0:100]} (\neg W_1) \land G_{[0:100]} (\neg W_2) \land G_{[0:100]} (\neg W_3) \land G_{[0:100]} (\neg W_4) \land G_{[0:100]} (\neg W_5) \land G_{[0:100]} (\neg W_6) ) $

\begin{figure}[!htbp]
  \centering \includegraphics[width=0.6\textwidth]{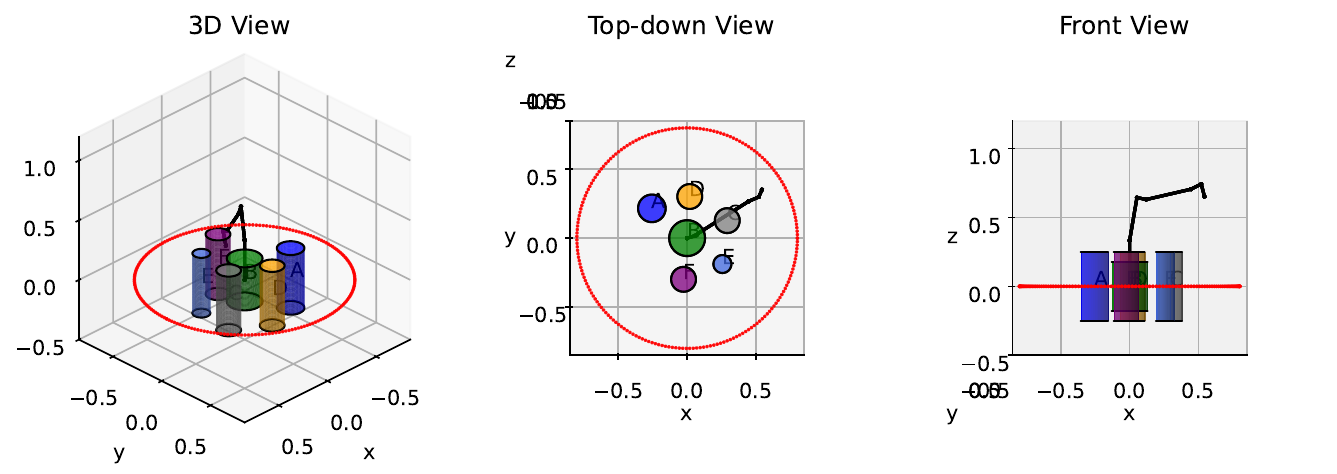}
  \caption{Scene for Franka Panda: STL task 04}
  \label{fig:scene-2arm-task-04}
\end{figure}
\noindent \textbf{STL-04 (Two-layer):} \quad $ F_{[5:10]} ( F_{[0:50]} (A) \land F_{[40:50]} (C) \land F_{[70:80]} (F) \land G_{[50:60]} (D) \land G_{[0:90]} (\neg B) \land G_{[0:90]} (\neg E) \land G_{[0:100]} (\neg W_1) \land G_{[0:100]} (\neg W_2) \land G_{[0:100]} (\neg W_3) \land G_{[0:100]} (\neg W_4) \land G_{[0:100]} (\neg W_5) \land G_{[0:100]} (\neg W_6) ) $

\begin{figure}[!htbp]
  \centering \includegraphics[width=0.6\textwidth]{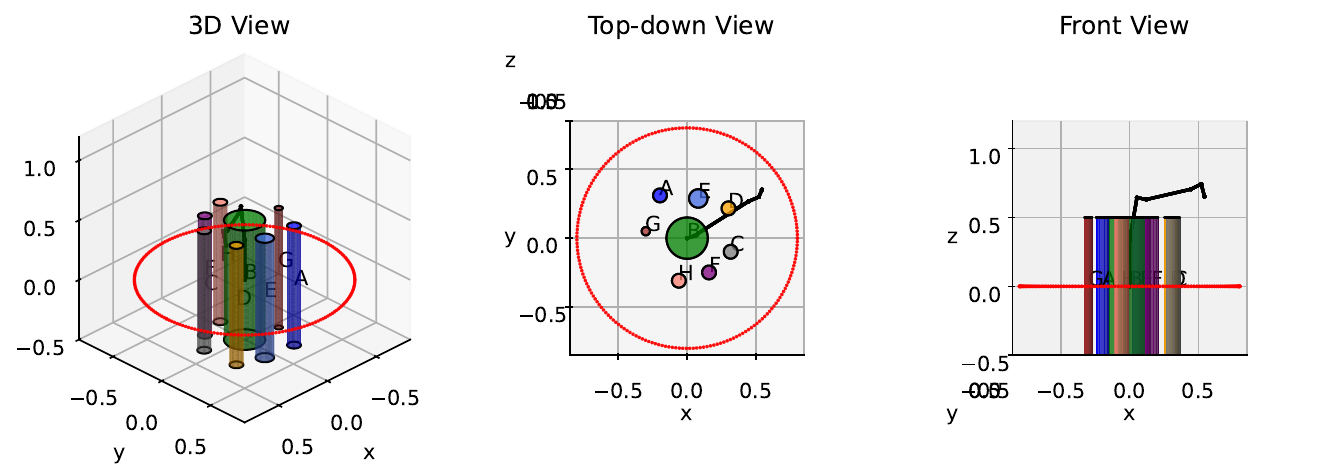}
  \caption{Scene for Franka Panda: STL task 05}
  \label{fig:scene-2arm-task-05}
\end{figure}
\noindent \textbf{STL-05 (Two-layer):} \quad $ F_{[5:10]} ( F_{[0:30]} (A) \land F_{[30:50]} (C) \land F_{[70:80]} (F) \land F_{[75:88]} (H) \land G_{[50:60]} (D) \land G_{[0:90]} (\neg B) \land G_{[0:90]} (\neg E) \land G_{[0:90]} (\neg G) \land G_{[0:100]} (\neg W_1) \land G_{[0:100]} (\neg W_2) \land G_{[0:100]} (\neg W_3) \land G_{[0:100]} (\neg W_4) \land G_{[0:100]} (\neg W_5) \land G_{[0:100]} (\neg W_6) ) $

\begin{figure}[!htbp]
  \centering \includegraphics[width=0.6\textwidth]{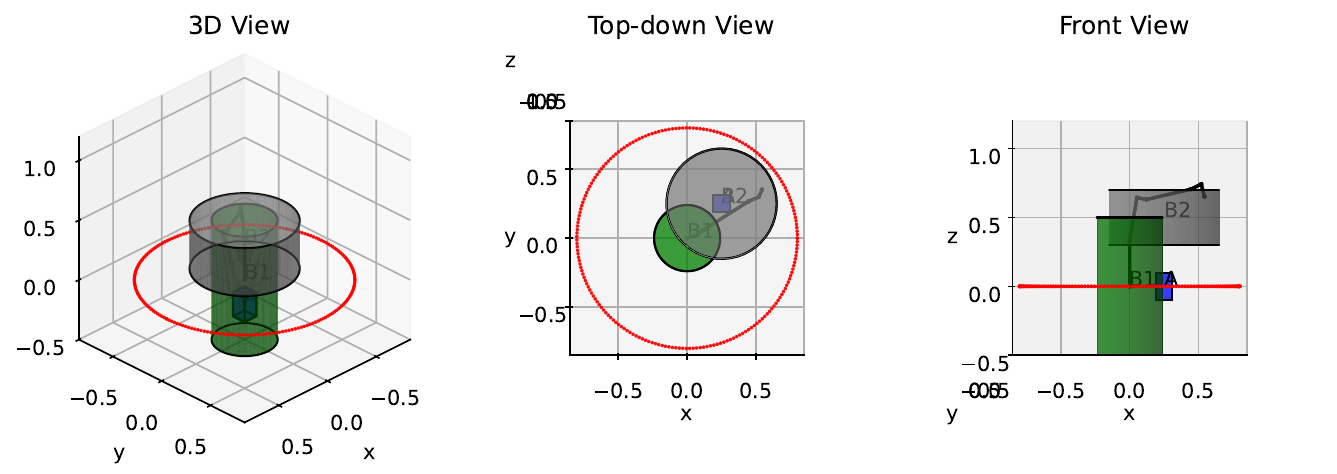}
  \caption{Scene for Franka Panda: STL task 06}
  \label{fig:scene-2arm-task-06}
\end{figure}
\noindent \textbf{STL-06 (Multi-layer):} \quad $  F_{[10:90]} (A) \land G_{[0:100]} (\neg B_1) \land G_{[0:100]} (\neg B_2) \land G_{[0:100]} (\neg W_1) \land G_{[0:100]} (\neg W_2) \land G_{[0:100]} (\neg W_3) \land G_{[0:100]} (\neg W_4) \land G_{[0:100]} (\neg W_5) \land G_{[0:100]} (\neg W_6)  $

\begin{figure}[!htbp]
  \centering \includegraphics[width=0.6\textwidth]{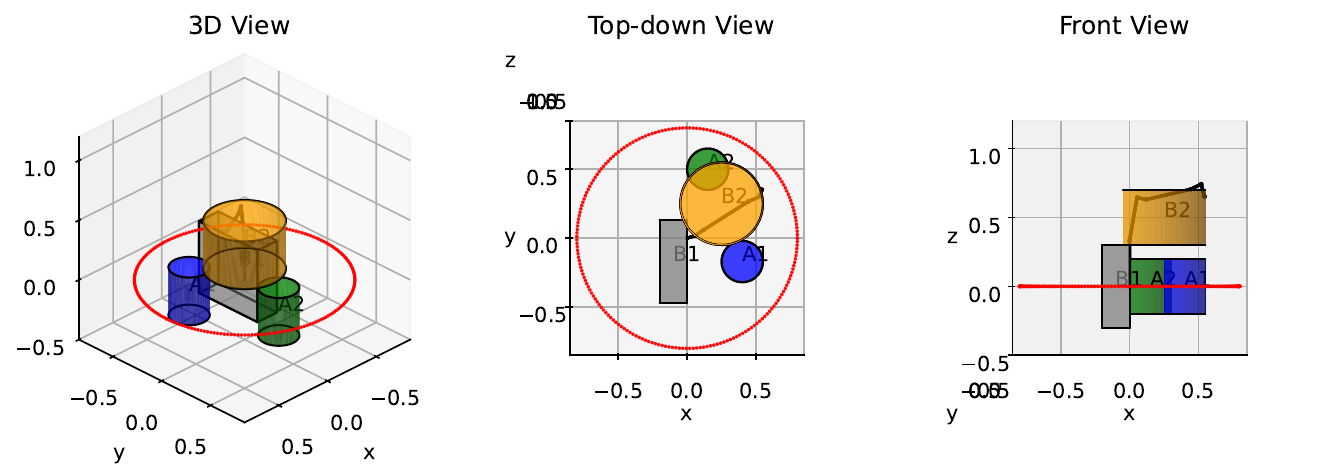}
  \caption{Scene for Franka Panda: STL task 07}
  \label{fig:scene-2arm-task-07}
\end{figure}
\noindent \textbf{STL-07 (Multi-layer):} \quad $  F_{[0:90]} (A_1) \land F_{[40:80]} (A_2) \land G_{[0:100]} (\neg B_1) \land G_{[0:100]} (\neg B_2) \land G_{[0:100]} (\neg W_1) \land G_{[0:100]} (\neg W_2) \land G_{[0:100]} (\neg W_3) \land G_{[0:100]} (\neg W_4) \land G_{[0:100]} (\neg W_5) \land G_{[0:100]} (\neg W_6)  $

\begin{figure}[!htbp]
  \centering \includegraphics[width=0.6\textwidth]{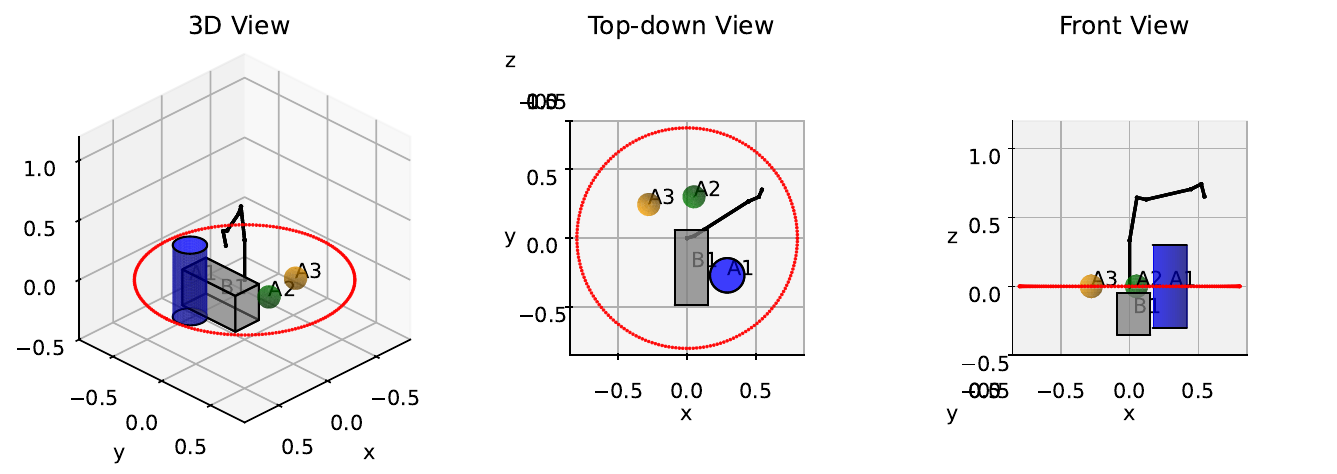}
  \caption{Scene for Franka Panda: STL task 08}
  \label{fig:scene-2arm-task-08}
\end{figure}
\noindent \textbf{STL-08 (Multi-layer):} \quad $  F_{[0:90]} (A_1) \land F_{[40:60]} ( A_2 \land F_{[15:30]} (G_{[0:5]} (A_3)) ) \land G_{[0:100]} (\neg B_1) \land G_{[0:100]} (\neg W_1) \land G_{[0:100]} (\neg W_2) \land G_{[0:100]} (\neg W_3) \land G_{[0:100]} (\neg W_4) \land G_{[0:100]} (\neg W_5) \land G_{[0:100]} (\neg W_6)  $

\begin{figure}[!htbp]
  \centering \includegraphics[width=0.6\textwidth]{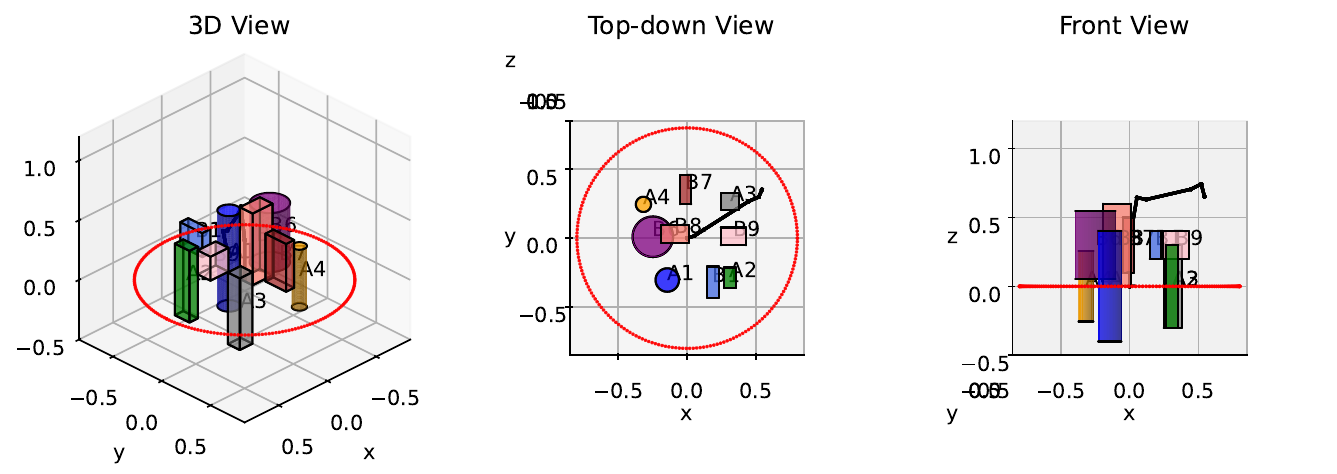}
  \caption{Scene for Franka Panda: STL task 09}
  \label{fig:scene-2arm-task-09}
\end{figure}
\noindent \textbf{STL-09 (Multi-layer):} \quad $  F_{[25:30]} ( A_1 \land F_{[20:28]} ( G_{[0:5]} (A_2) \land  F_{[10:30]} (G_{[0:5]} (A_3)) \land F_{[10:30]} (G_{[0:10]} (A_4))  ) ) \land G_{[0:100]} (\neg B_1) \land G_{[0:100]} (\neg B_6) \land G_{[0:100]} (\neg B_7) \land G_{[0:100]} (\neg B_8) \land G_{[0:100]} (\neg B_9) \land G_{[0:100]} (\neg W_1) \land G_{[0:100]} (\neg W_2) \land G_{[0:100]} (\neg W_3) \land G_{[0:100]} (\neg W_4) \land G_{[0:100]} (\neg W_5) \land G_{[0:100]} (\neg W_6)  $

\begin{figure}[!htbp]
  \centering \includegraphics[width=0.6\textwidth]{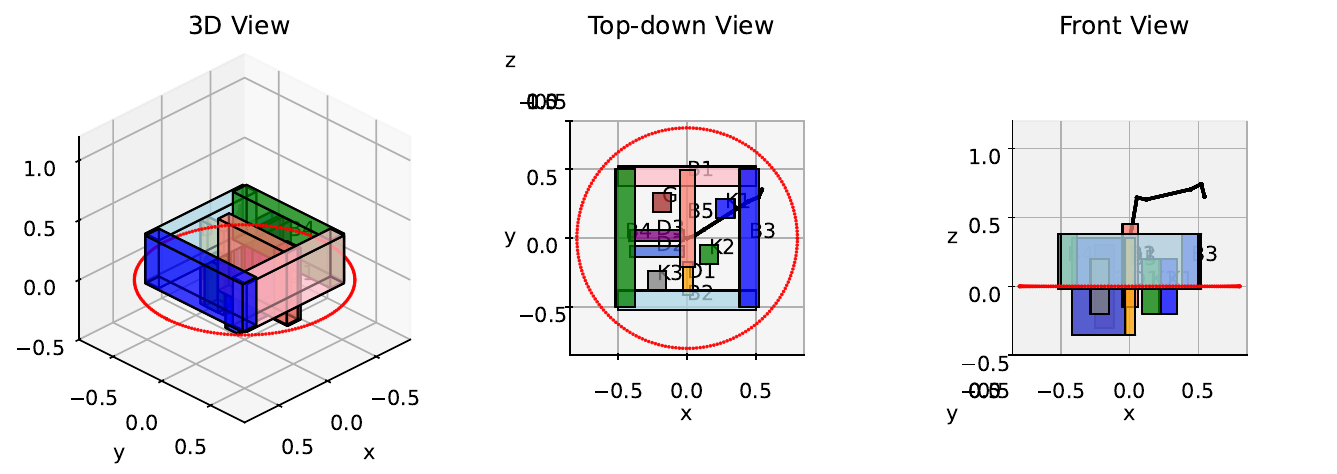}
  \caption{Scene for Franka Panda: STL task 10}
  \label{fig:scene-2arm-task-10}
\end{figure}
\noindent \textbf{STL-10 (Multi-layer):} \quad $  (\neg D_1)U_{[0:100]}(K_1) \land (\neg D_2)U_{[0:100]}(K_2) \land (\neg D_3)U_{[0:100]}(K_3) \land F_{[80:90]} (G_{[0:5]} (G)) \land G_{[0:100]} (\neg B_5) \land G_{[0:100]} (\neg B_1) \land G_{[0:100]} (\neg B_2) \land G_{[0:100]} (\neg B_3) \land G_{[0:100]} (\neg B_4) \land G_{[0:100]} (\neg W_1) \land G_{[0:100]} (\neg W_2) \land G_{[0:100]} (\neg W_3) \land G_{[0:100]} (\neg W_4) \land G_{[0:100]} (\neg W_5) \land G_{[0:100]} (\neg W_6)  $

\subsubsection{STLs in ``Quadrotor" environment}

\begin{figure}[!htbp]
  \centering \includegraphics[width=0.6\textwidth]{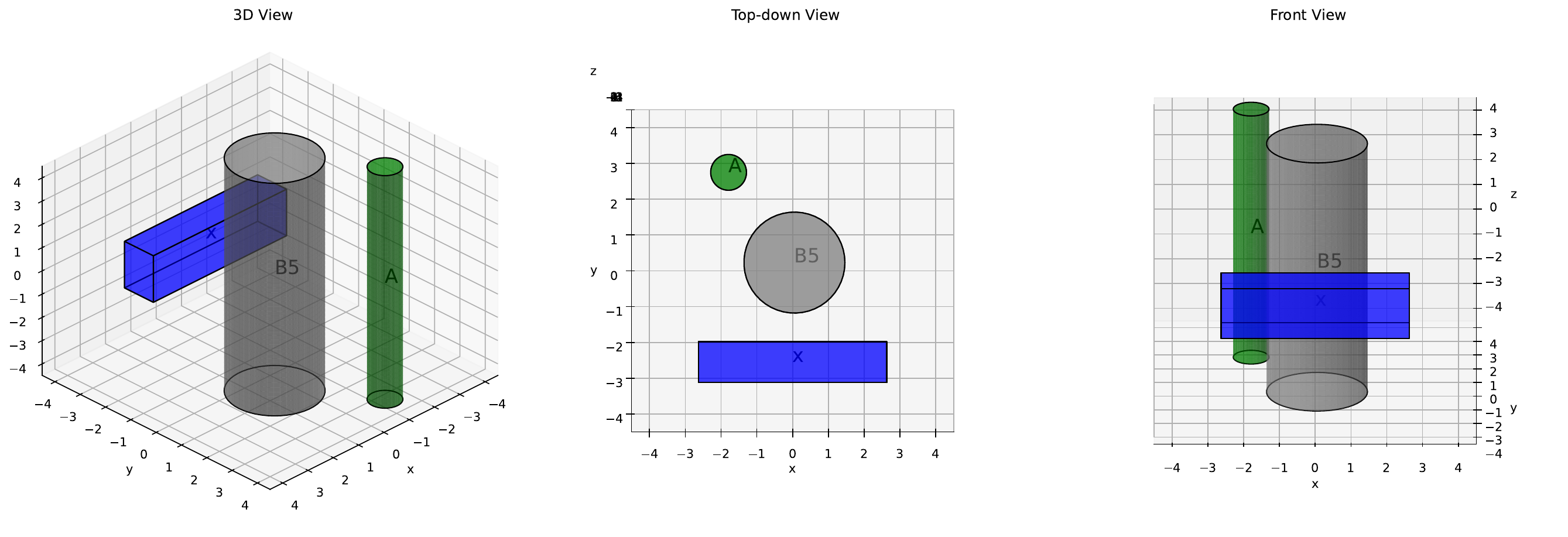}
  \caption{Scene for Quadrotor: STL task 01}
  \label{fig:scene-3drone-task-01}
\end{figure}
\noindent \textbf{STL-01 (Two-layer):} \quad $ F_{[5:7]} ( F_{[50:85]} (A) \land G_{[0:90]} (\neg B_5) \land G_{[0:100]} (\neg W_1) \land G_{[0:100]} (\neg W_2) \land G_{[0:100]} (\neg W_3) \land G_{[0:100]} (\neg W_4) \land G_{[0:100]} (\neg W_5) \land G_{[0:100]} (\neg W_6) ) $

\begin{figure}[!htbp]
  \centering \includegraphics[width=0.6\textwidth]{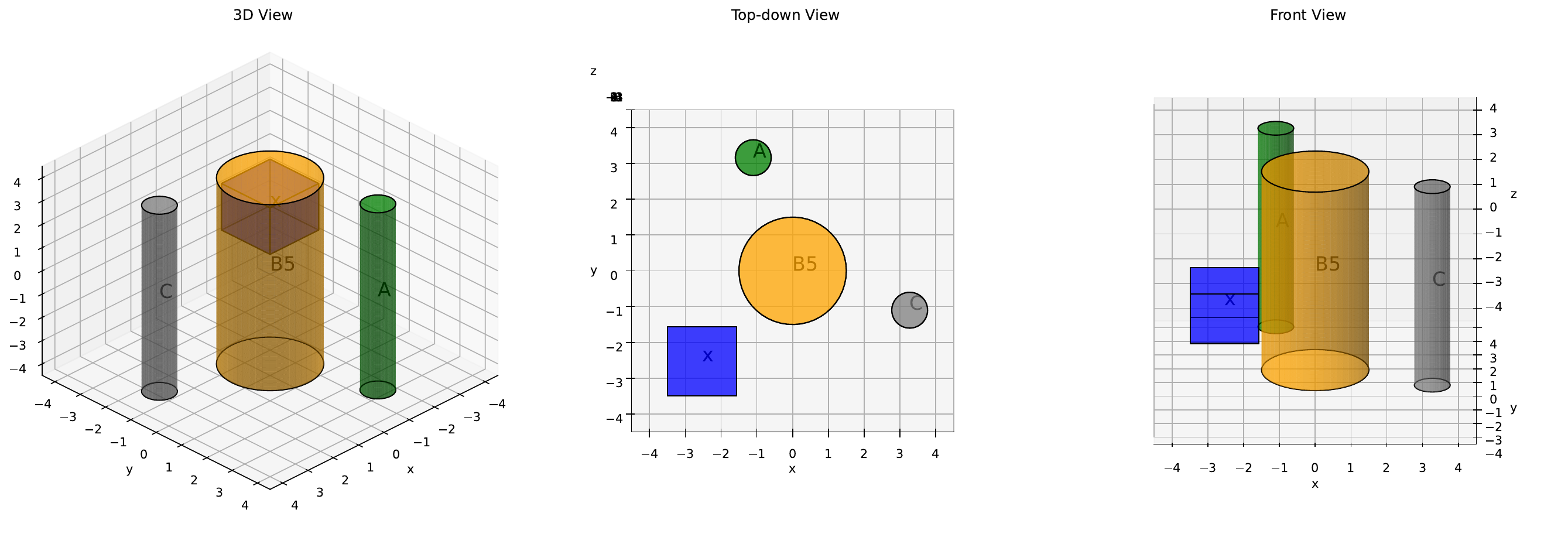}
  \caption{Scene for Quadrotor: STL task 02}
  \label{fig:scene-3drone-task-02}
\end{figure}
\noindent \textbf{STL-02 (Two-layer):} \quad $ F_{[5:10]} ( F_{[0:50]} (A) \land G_{[60:80]} (C) \land G_{[0:90]} (\neg B_5) \land G_{[0:100]} (\neg W_1) \land G_{[0:100]} (\neg W_2) \land G_{[0:100]} (\neg W_3) \land G_{[0:100]} (\neg W_4) \land G_{[0:100]} (\neg W_5) \land G_{[0:100]} (\neg W_6) ) $

\begin{figure}[!htbp]
  \centering \includegraphics[width=0.6\textwidth]{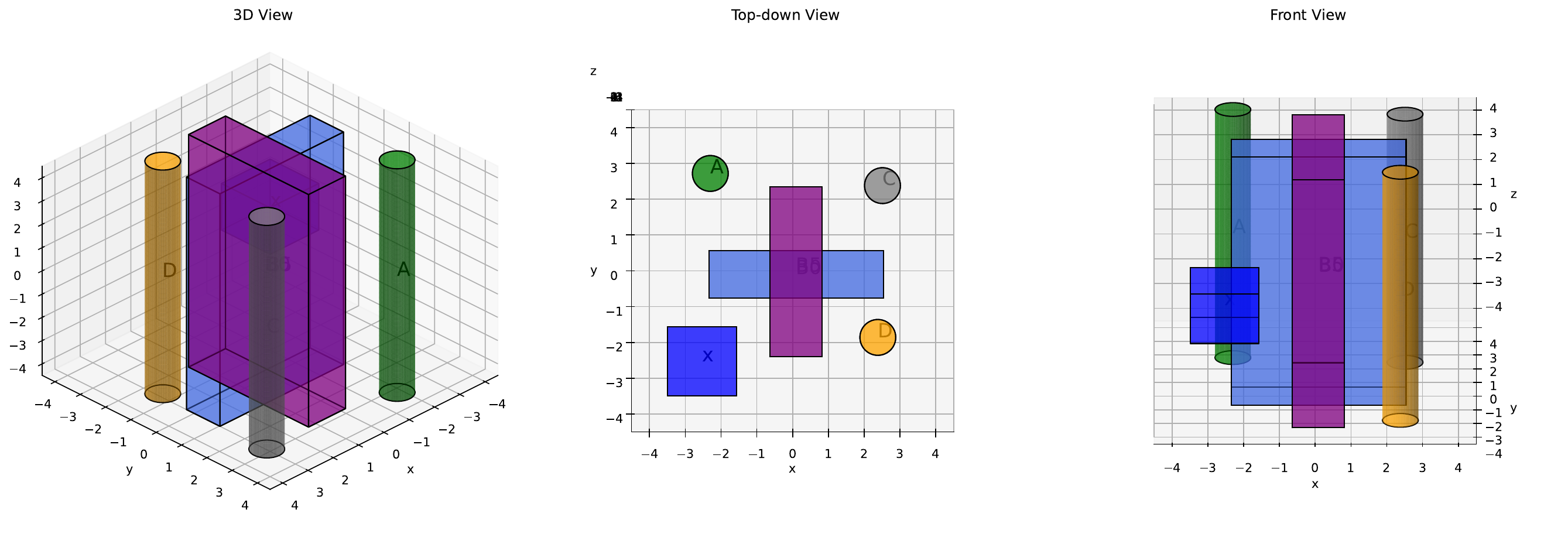}
  \caption{Scene for Quadrotor: STL task 03}
  \label{fig:scene-3drone-task-03}
\end{figure}
\noindent \textbf{STL-03 (Two-layer):} \quad $ F_{[5:10]} ( F_{[0:50]} (A) \land F_{[40:60]} (C) \land G_{[70:80]} (D) \land G_{[0:90]} (\neg B_5) \land G_{[0:90]} (\neg B_0) \land G_{[0:100]} (\neg W_1) \land G_{[0:100]} (\neg W_2) \land G_{[0:100]} (\neg W_3) \land G_{[0:100]} (\neg W_4) \land G_{[0:100]} (\neg W_5) \land G_{[0:100]} (\neg W_6) ) $

\begin{figure}[!htbp]
  \centering \includegraphics[width=0.6\textwidth]{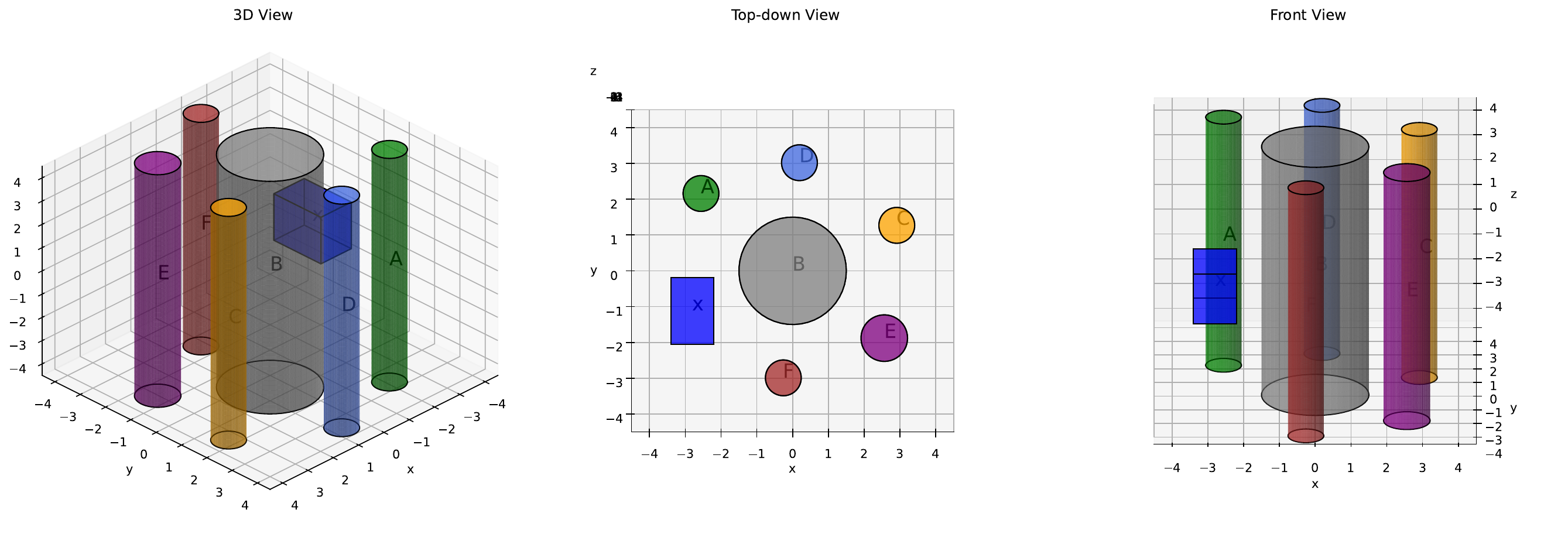}
  \caption{Scene for Quadrotor: STL task 04}
  \label{fig:scene-3drone-task-04}
\end{figure}
\noindent \textbf{STL-04 (Two-layer):} \quad $ F_{[5:10]} ( F_{[0:50]} (A) \land F_{[40:50]} (C) \land F_{[70:80]} (F) \land G_{[50:60]} (D) \land G_{[0:90]} (\neg B) \land G_{[0:90]} (\neg E) \land G_{[0:100]} (\neg W_1) \land G_{[0:100]} (\neg W_2) \land G_{[0:100]} (\neg W_3) \land G_{[0:100]} (\neg W_4) \land G_{[0:100]} (\neg W_5) \land G_{[0:100]} (\neg W_6) ) $

\begin{figure}[!htbp]
  \centering \includegraphics[width=0.6\textwidth]{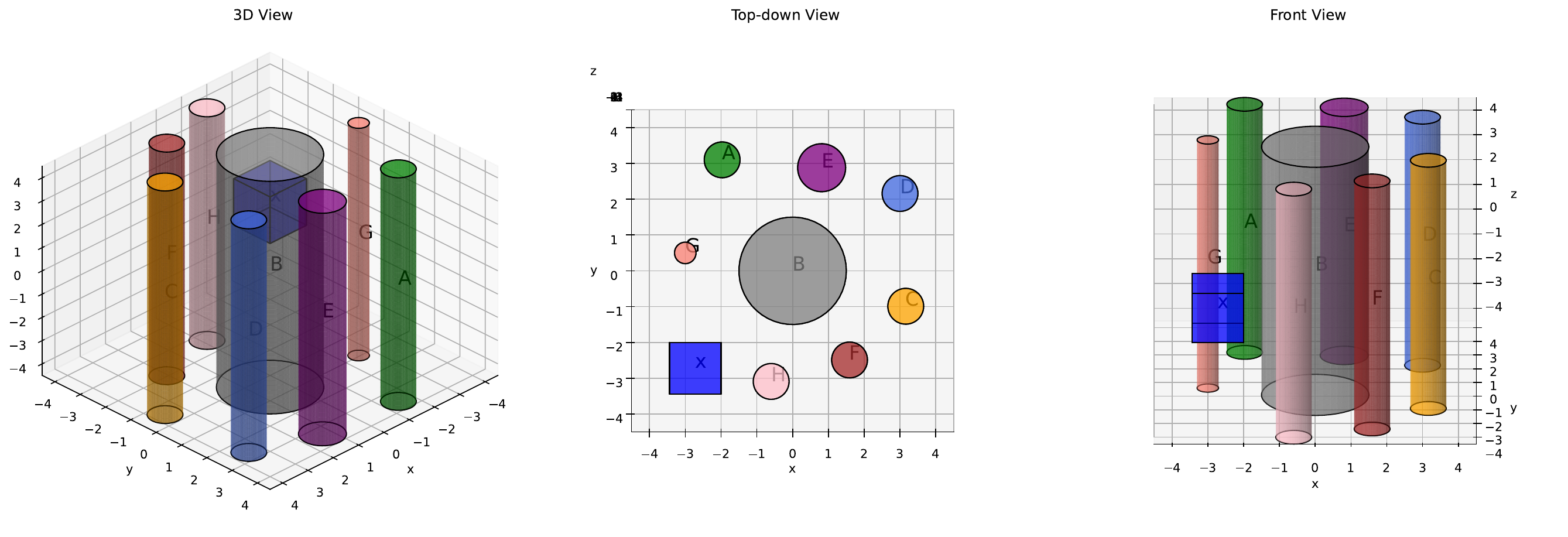}
  \caption{Scene for Quadrotor: STL task 05}
  \label{fig:scene-3drone-task-05}
\end{figure}
\noindent \textbf{STL-05 (Two-layer):} \quad $ F_{[5:10]} ( F_{[0:30]} (A) \land F_{[30:50]} (C) \land F_{[70:80]} (F) \land F_{[75:88]} (H) \land G_{[50:60]} (D) \land G_{[0:90]} (\neg B) \land G_{[0:90]} (\neg E) \land G_{[0:90]} (\neg G) \land G_{[0:100]} (\neg W_1) \land G_{[0:100]} (\neg W_2) \land G_{[0:100]} (\neg W_3) \land G_{[0:100]} (\neg W_4) \land G_{[0:100]} (\neg W_5) \land G_{[0:100]} (\neg W_6) ) $

\begin{figure}[!htbp]
  \centering \includegraphics[width=0.6\textwidth]{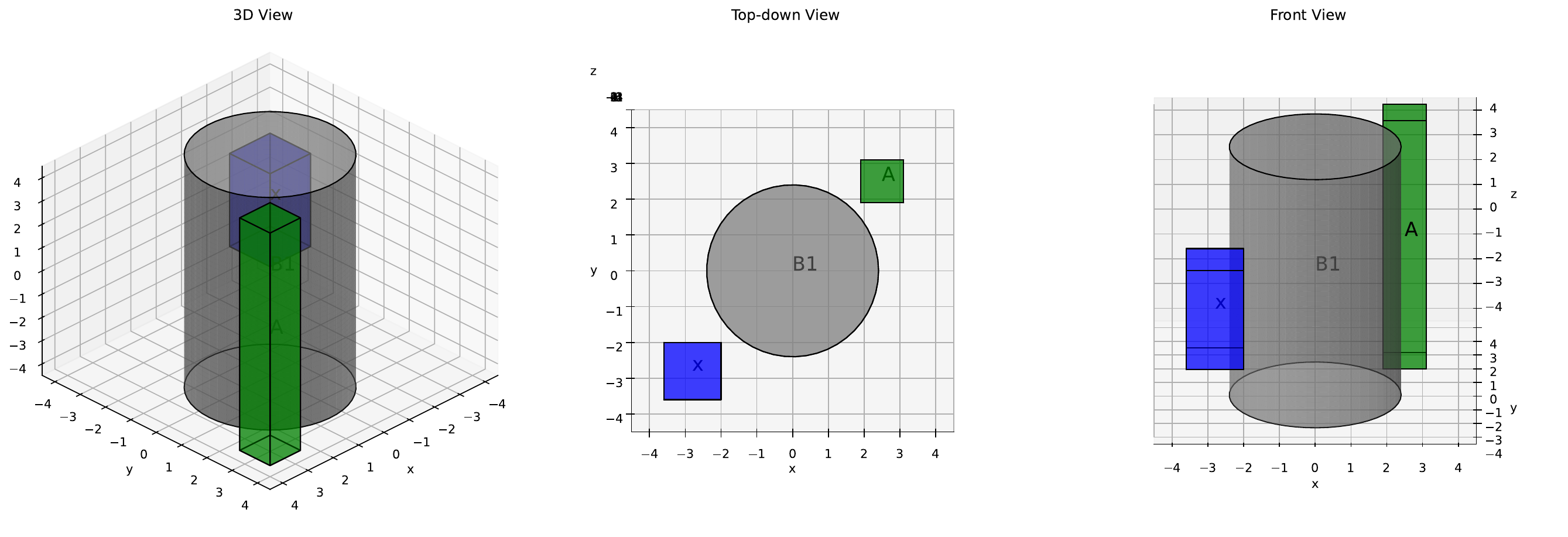}
  \caption{Scene for Quadrotor: STL task 06}
  \label{fig:scene-3drone-task-06}
\end{figure}
\noindent \textbf{STL-06 (Multi-layer):} \quad $  F_{[10:90]} (A) \land G_{[0:100]} (\neg B_1) \land G_{[0:100]} (\neg W_1) \land G_{[0:100]} (\neg W_2) \land G_{[0:100]} (\neg W_3) \land G_{[0:100]} (\neg W_4) \land G_{[0:100]} (\neg W_5) \land G_{[0:100]} (\neg W_6)  $

\begin{figure}[!htbp]
  \centering \includegraphics[width=0.6\textwidth]{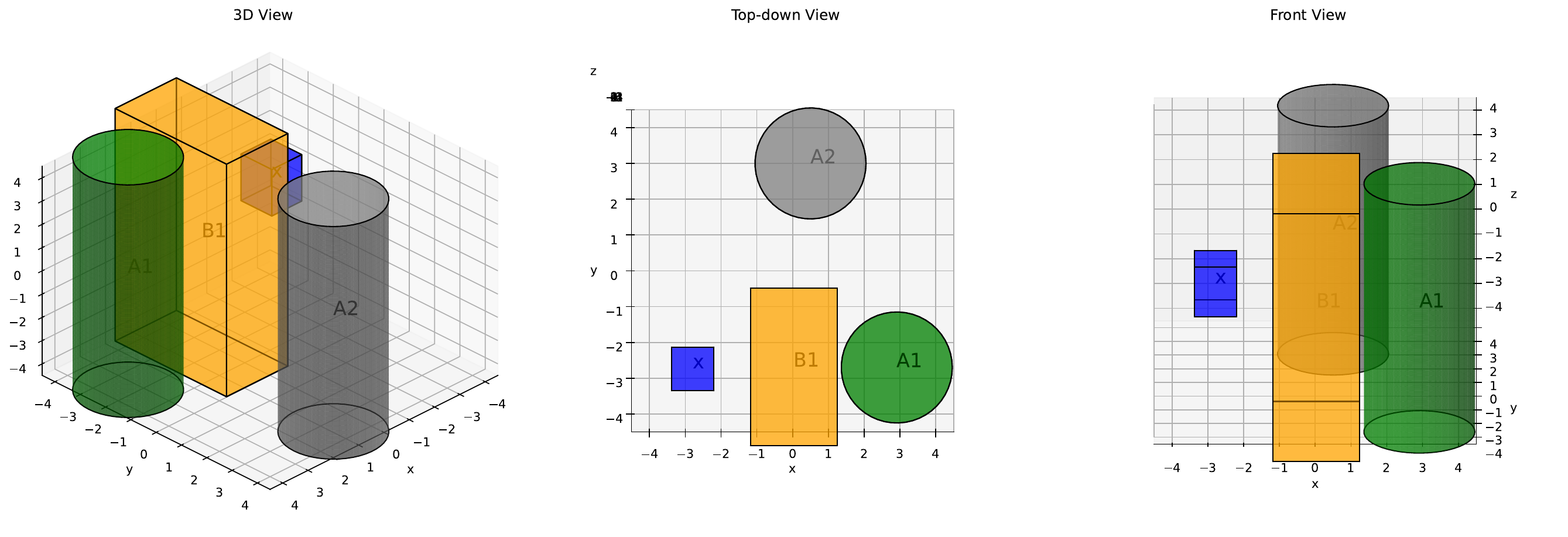}
  \caption{Scene for Quadrotor: STL task 07}
  \label{fig:scene-3drone-task-07}
\end{figure}
\noindent \textbf{STL-07 (Multi-layer):} \quad $  F_{[0:90]} (A_1) \land F_{[40:80]} (A_2) \land G_{[0:100]} (\neg B_1) \land G_{[0:100]} (\neg W_1) \land G_{[0:100]} (\neg W_2) \land G_{[0:100]} (\neg W_3) \land G_{[0:100]} (\neg W_4) \land G_{[0:100]} (\neg W_5) \land G_{[0:100]} (\neg W_6)  $

\begin{figure}[!htbp]
  \centering \includegraphics[width=0.6\textwidth]{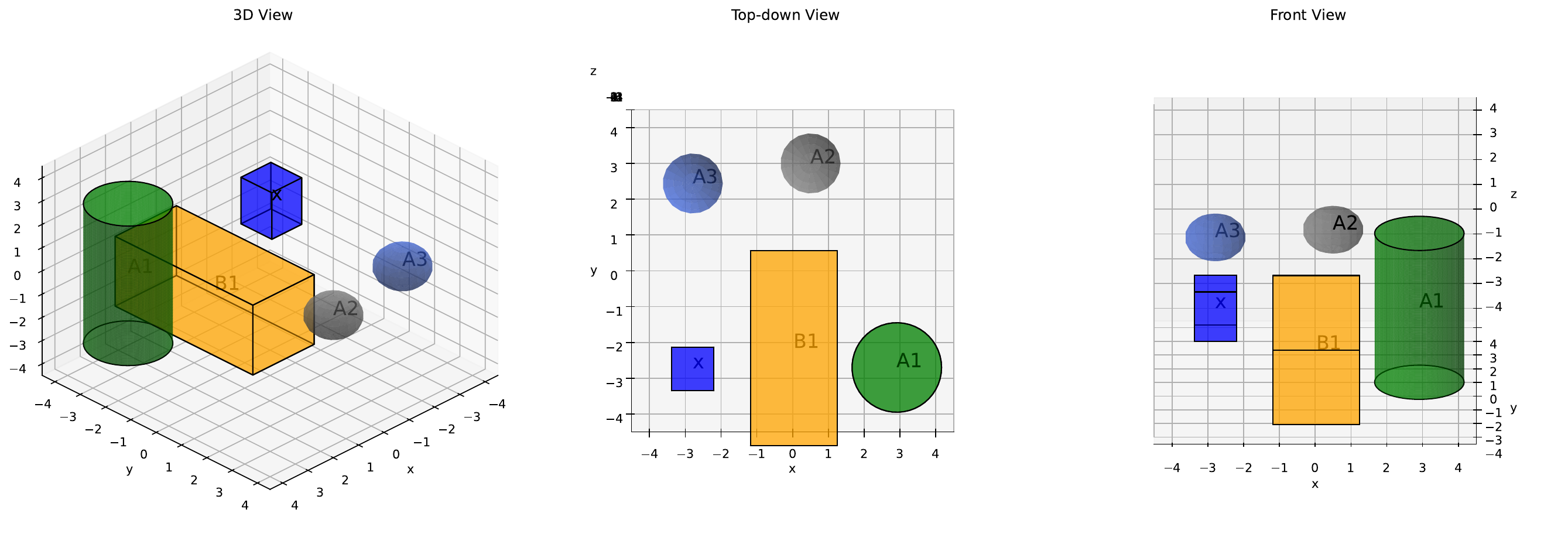}
  \caption{Scene for Quadrotor: STL task 08}
  \label{fig:scene-3drone-task-08}
\end{figure}
\noindent \textbf{STL-08 (Multi-layer):} \quad $  F_{[0:90]} (A_1) \land F_{[40:60]} ( A_2 \land F_{[15:30]} (G_{[0:5]} (A_3)) ) \land G_{[0:100]} (\neg B_1) \land G_{[0:100]} (\neg W_1) \land G_{[0:100]} (\neg W_2) \land G_{[0:100]} (\neg W_3) \land G_{[0:100]} (\neg W_4) \land G_{[0:100]} (\neg W_5) \land G_{[0:100]} (\neg W_6)  $

\begin{figure}[!htbp]
  \centering \includegraphics[width=0.6\textwidth]{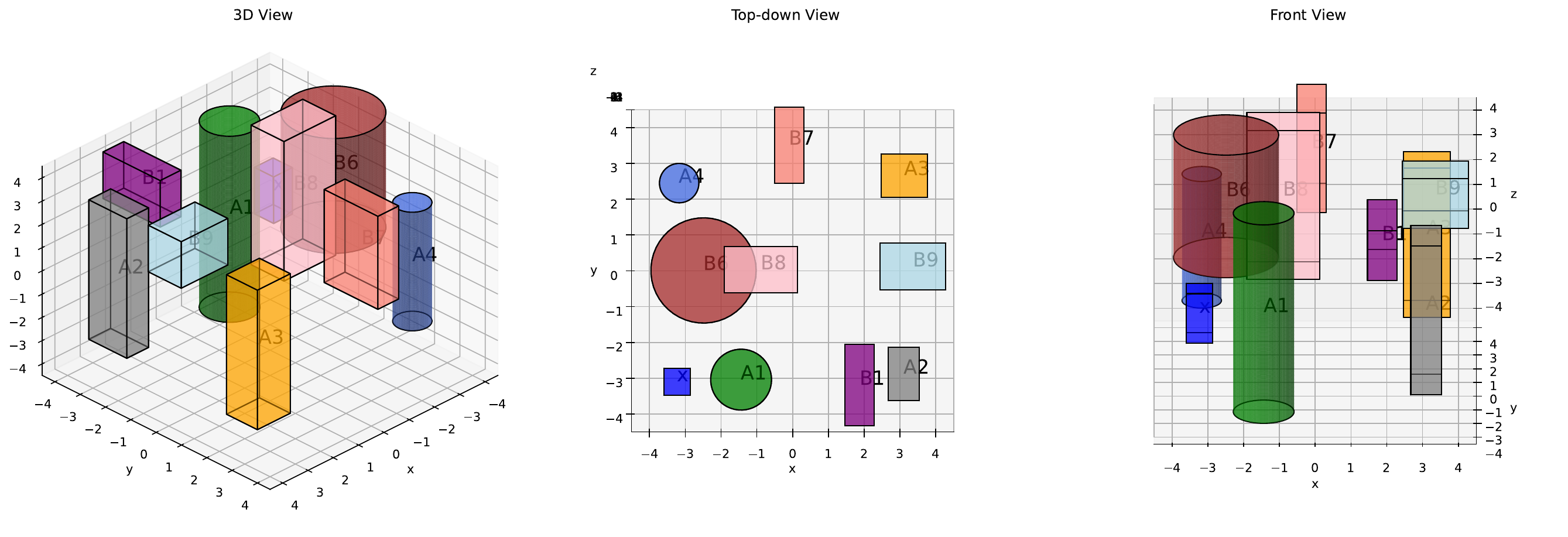}
  \caption{Scene for Quadrotor: STL task 09}
  \label{fig:scene-3drone-task-09}
\end{figure}
\noindent \textbf{STL-09 (Multi-layer):} \quad $  F_{[25:30]} ( A_1 \land F_{[20:28]} ( G_{[0:5]} (A_2) \land  F_{[10:30]} (G_{[0:5]} (A_3)) \land F_{[10:30]} (G_{[0:10]} (A_4))  ) ) \land G_{[0:100]} (\neg B_1) \land G_{[0:100]} (\neg B_6) \land G_{[0:100]} (\neg B_7) \land G_{[0:100]} (\neg B_8) \land G_{[0:100]} (\neg B_9) \land G_{[0:100]} (\neg W_1) \land G_{[0:100]} (\neg W_2) \land G_{[0:100]} (\neg W_3) \land G_{[0:100]} (\neg W_4) \land G_{[0:100]} (\neg W_5) \land G_{[0:100]} (\neg W_6)  $

\begin{figure}[!htbp]
  \centering \includegraphics[width=0.6\textwidth]{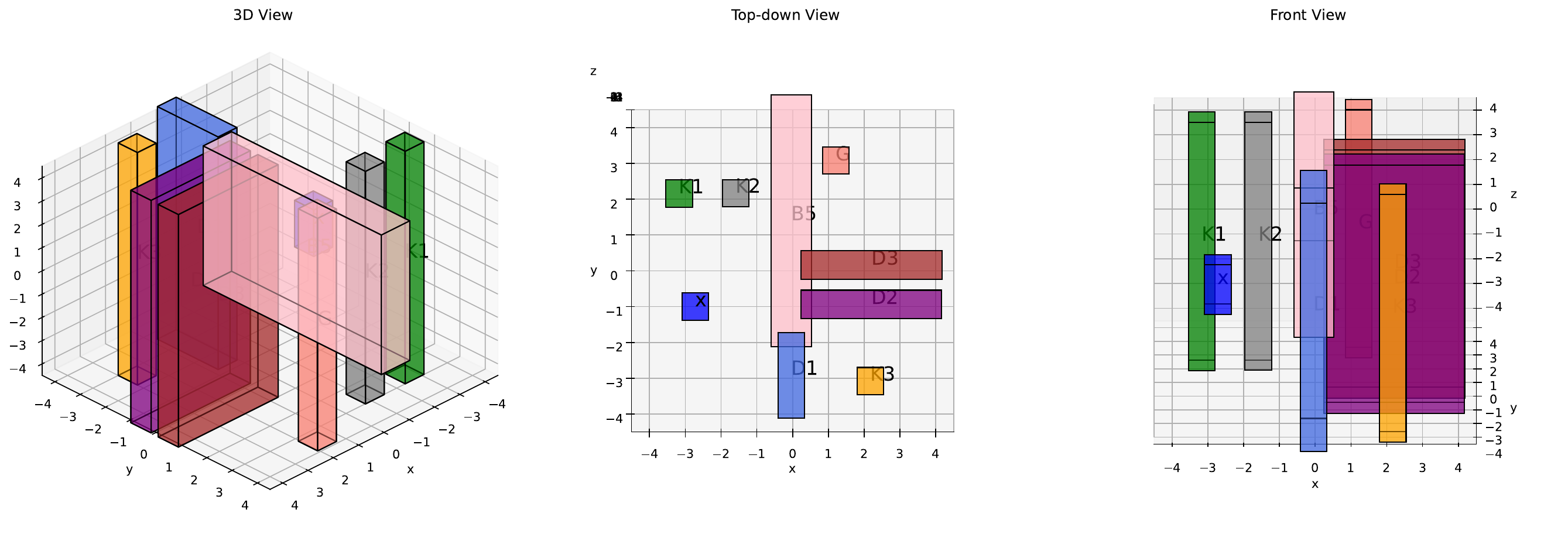}
  \caption{Scene for Quadrotor: STL task 10}
  \label{fig:scene-3drone-task-10}
\end{figure}
\noindent \textbf{STL-10 (Multi-layer):} \quad $  (\neg D_1)U_{[0:100]}(K_1) \land (\neg D_2)U_{[0:100]}(K_2) \land (\neg D_3)U_{[0:100]}(K_3) \land F_{[80:90]} (G_{[0:5]} (G)) \land G_{[0:100]} (\neg B_5) \land G_{[0:100]} (\neg W_1) \land G_{[0:100]} (\neg W_2) \land G_{[0:100]} (\neg W_3) \land G_{[0:100]} (\neg W_4) \land G_{[0:100]} (\neg W_5) \land G_{[0:100]} (\neg W_6)  $

\subsubsection{STLs in ``Ant" environment}

\begin{figure}[!htbp]
  \centering
  \begin{subfigure}[b]{0.19\textwidth}
      \centering \includegraphics[width=\textwidth]{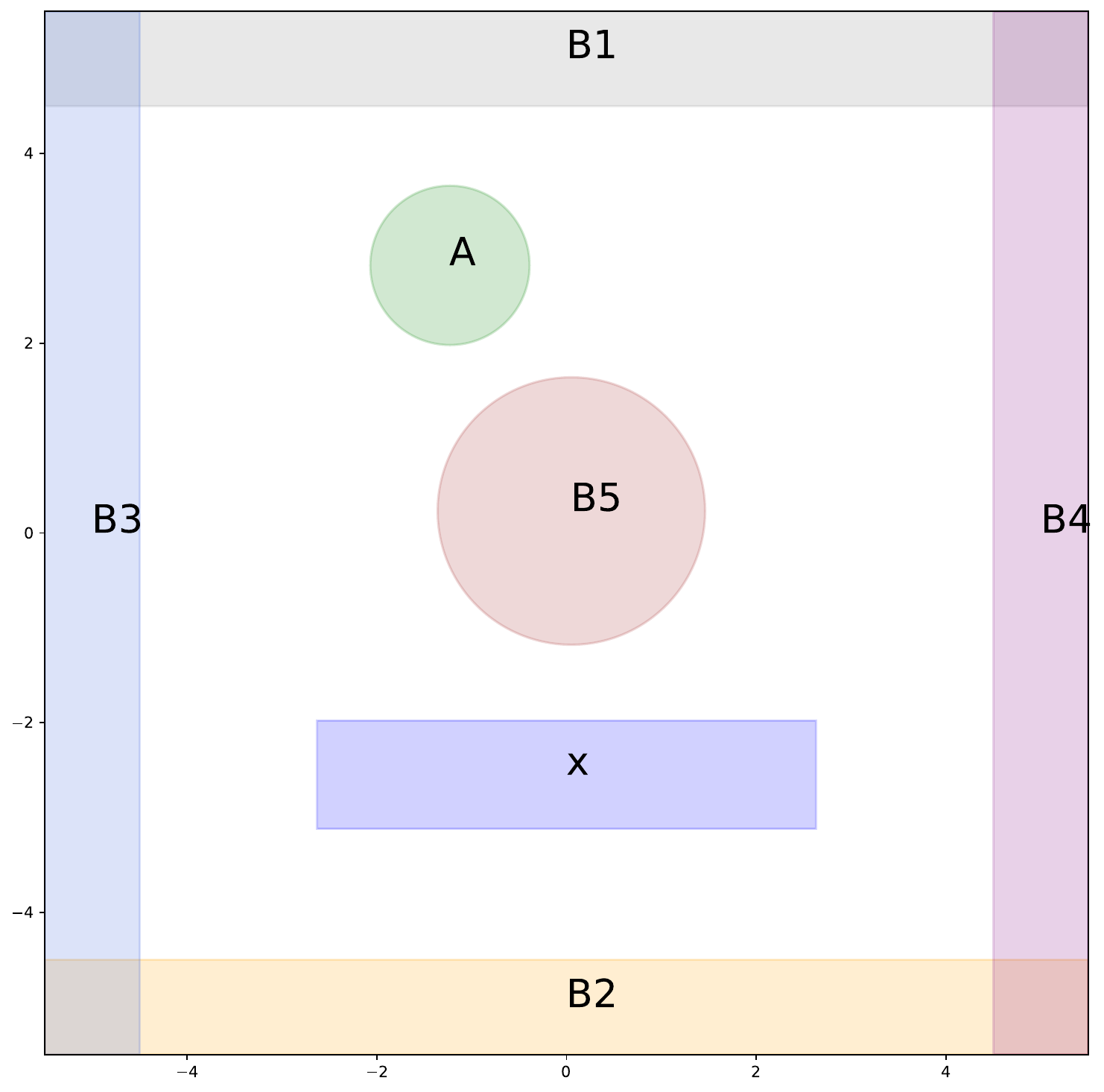}
      \caption{STL-01}
      \label{fig:scene-4ant-task-01}
  \end{subfigure}
  \begin{subfigure}[b]{0.19\textwidth}
      \centering \includegraphics[width=\textwidth]{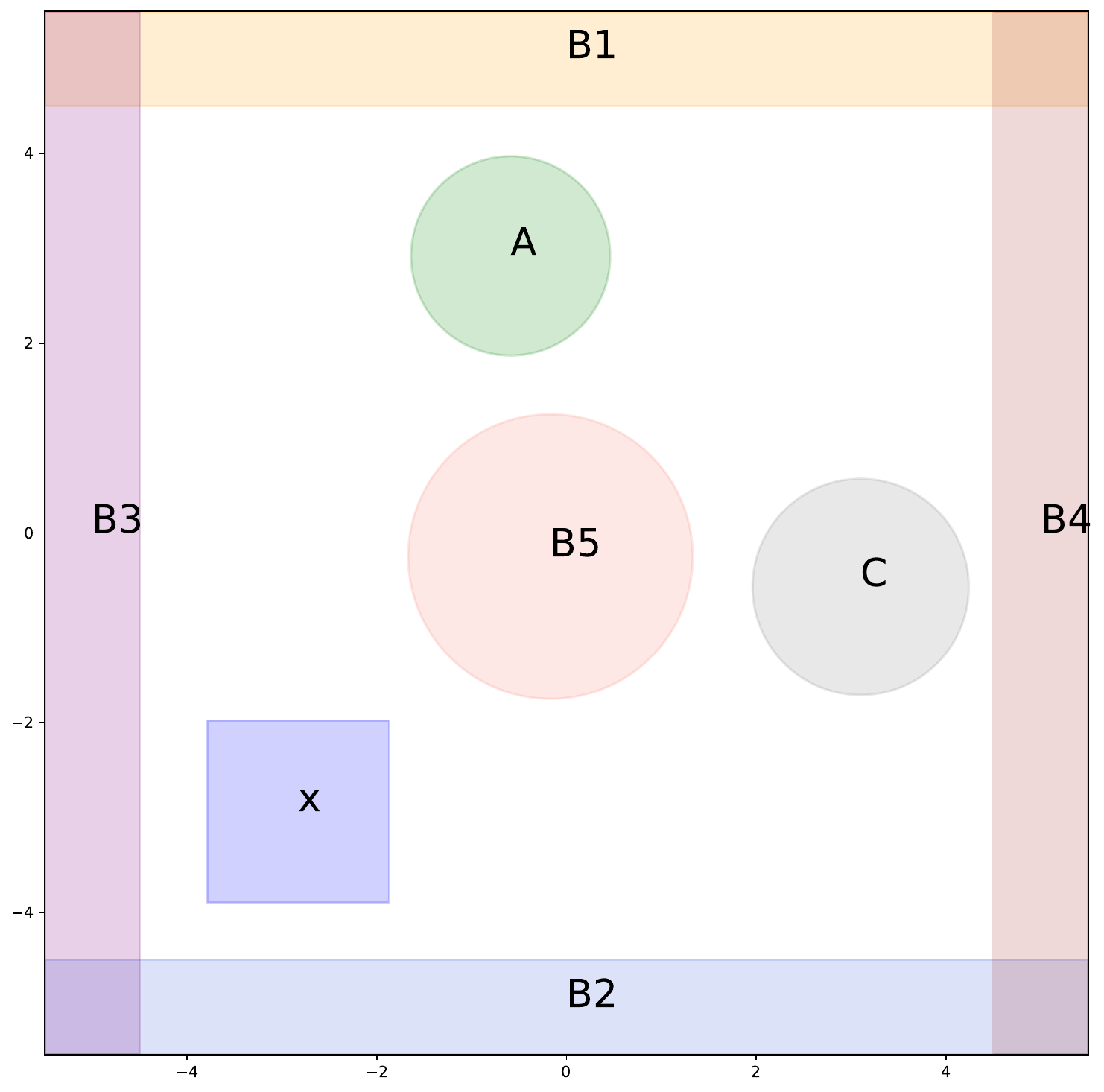}
      \caption{STL-02}
      \label{fig:scene-4ant-task-02}
  \end{subfigure}
  \begin{subfigure}[b]{0.19\textwidth}
      \centering \includegraphics[width=\textwidth]{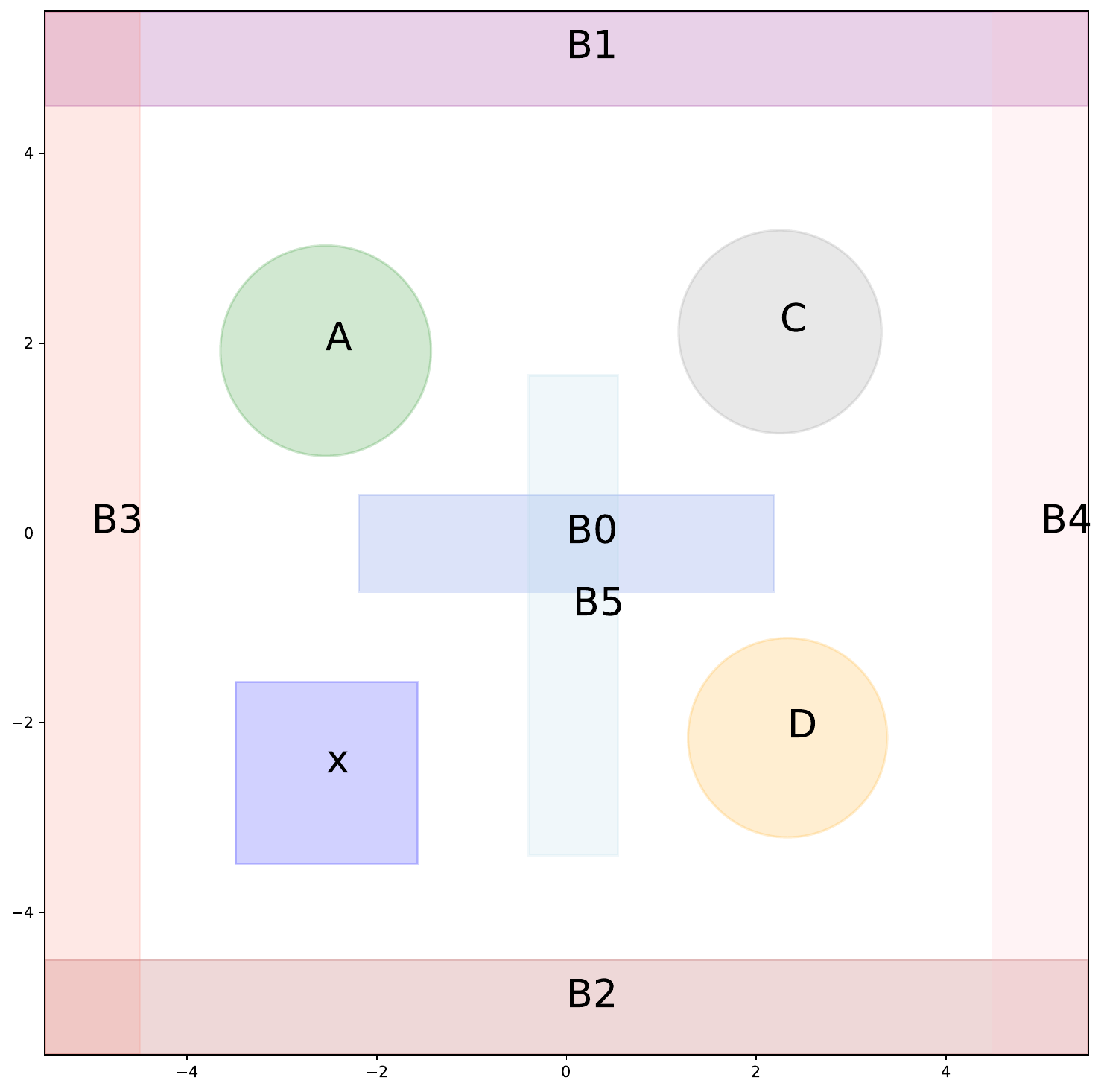}
      \caption{STL-03}
      \label{fig:scene-4ant-task-03}
  \end{subfigure}
  \begin{subfigure}[b]{0.19\textwidth}
      \centering \includegraphics[width=\textwidth]{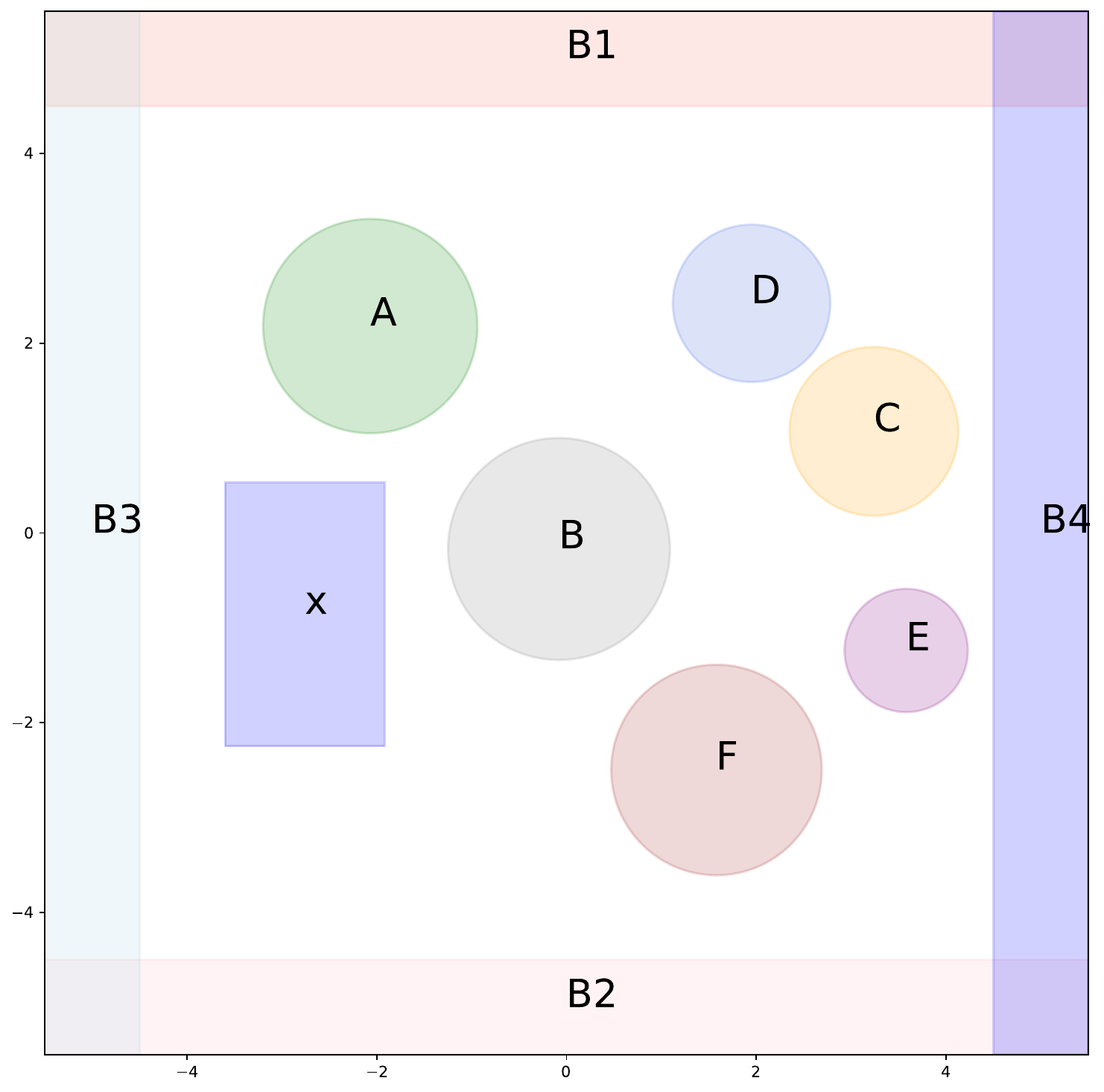}
      \caption{STL-04}
      \label{fig:scene-4ant-task-04}
  \end{subfigure}
  \begin{subfigure}[b]{0.19\textwidth}
      \centering \includegraphics[width=\textwidth]{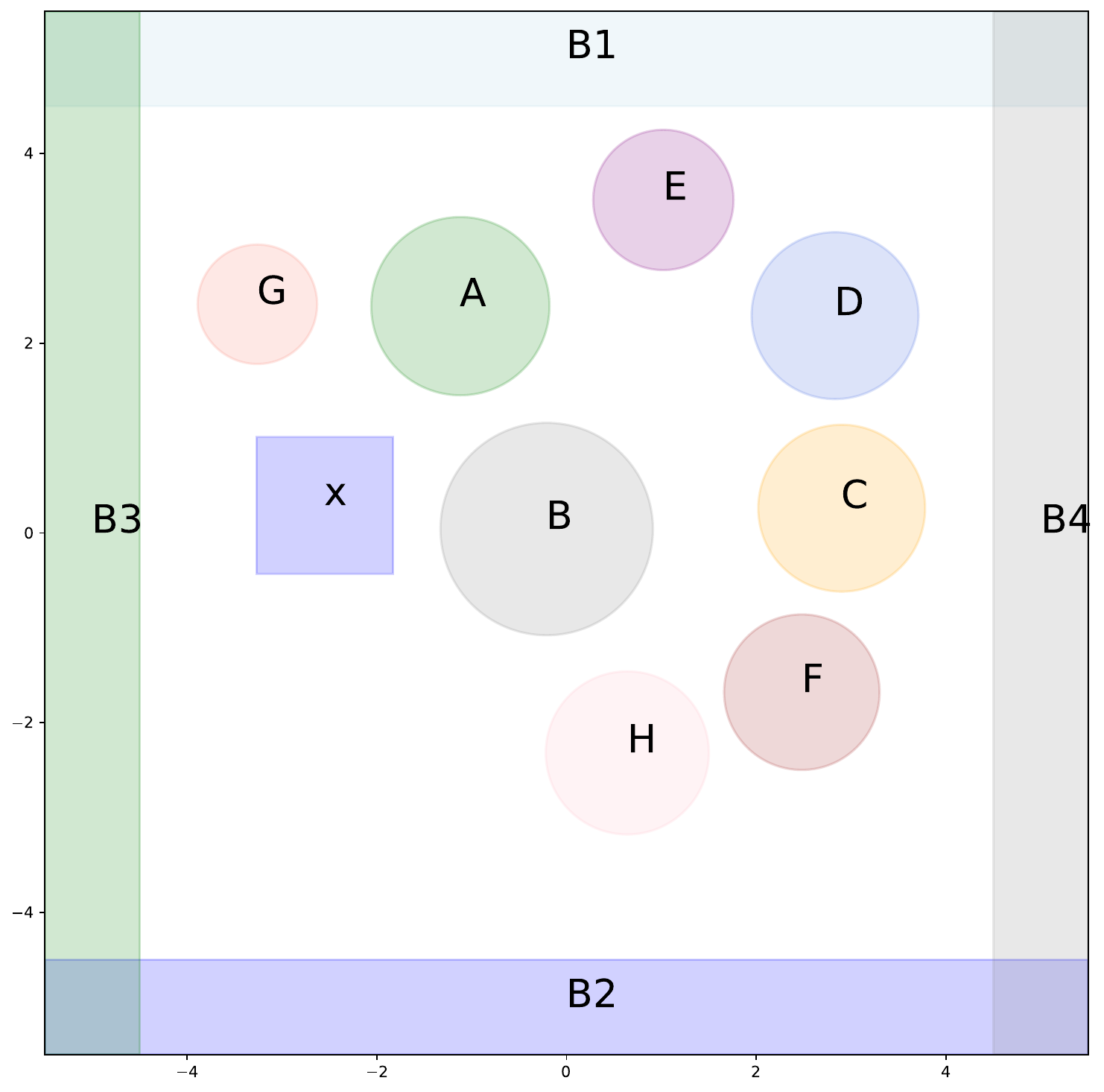}
      \caption{STL-05}
      \label{fig:scene-4ant-task-05}
  \end{subfigure}
  \caption{Scene for Ant: STL tasks 01 to 05}
  \label{fig:scene-cat-4ant-tasks-01-05}
\end{figure}
\noindent \textbf{STL-01 (Two-layer):} \quad $ F_{[10:14]} ( F_{[100:170]} (A) \land G_{[0:180]} (\neg B_5) \land G_{[0:180]} (\neg B_1) \land G_{[0:180]} (\neg B_2) \land G_{[0:180]} (\neg B_3) \land G_{[0:180]} (\neg B_4) ) $

\noindent \textbf{STL-02 (Two-layer):} \quad $ F_{[10:20]} ( F_{[0:100]} (A) \land G_{[120:160]} (C) \land G_{[0:180]} (\neg B_5) \land G_{[0:180]} (\neg B_1) \land G_{[0:180]} (\neg B_2) \land G_{[0:180]} (\neg B_3) \land G_{[0:180]} (\neg B_4) ) $

\noindent \textbf{STL-03 (Two-layer):} \quad $ F_{[10:20]} ( F_{[0:100]} (A) \land F_{[80:120]} (C) \land G_{[140:160]} (D) \land G_{[0:180]} (\neg B_5) \land G_{[0:180]} (\neg B_0) \land G_{[0:180]} (\neg B_1) \land G_{[0:180]} (\neg B_2) \land G_{[0:180]} (\neg B_3) \land G_{[0:180]} (\neg B_4) ) $

\noindent \textbf{STL-04 (Two-layer):} \quad $ F_{[10:20]} ( F_{[0:100]} (A) \land F_{[80:100]} (C) \land F_{[140:160]} (F) \land G_{[100:120]} (D) \land G_{[0:180]} (\neg B) \land G_{[0:180]} (\neg E) \land G_{[0:180]} (\neg B_1) \land G_{[0:180]} (\neg B_2) \land G_{[0:180]} (\neg B_3) \land G_{[0:180]} (\neg B_4) ) $

\noindent \textbf{STL-05 (Two-layer):} \quad $ F_{[10:20]} ( F_{[0:60]} (A) \land F_{[60:100]} (C) \land F_{[140:160]} (F) \land F_{[150:176]} (H) \land G_{[100:120]} (D) \land G_{[0:180]} (\neg B) \land G_{[0:180]} (\neg E) \land G_{[0:180]} (\neg G) \land G_{[0:180]} (\neg B_1) \land G_{[0:180]} (\neg B_2) \land G_{[0:180]} (\neg B_3) \land G_{[0:180]} (\neg B_4) ) $

\begin{figure}[!htbp]
  \centering
  \begin{subfigure}[b]{0.19\textwidth}
      \centering \includegraphics[width=\textwidth]{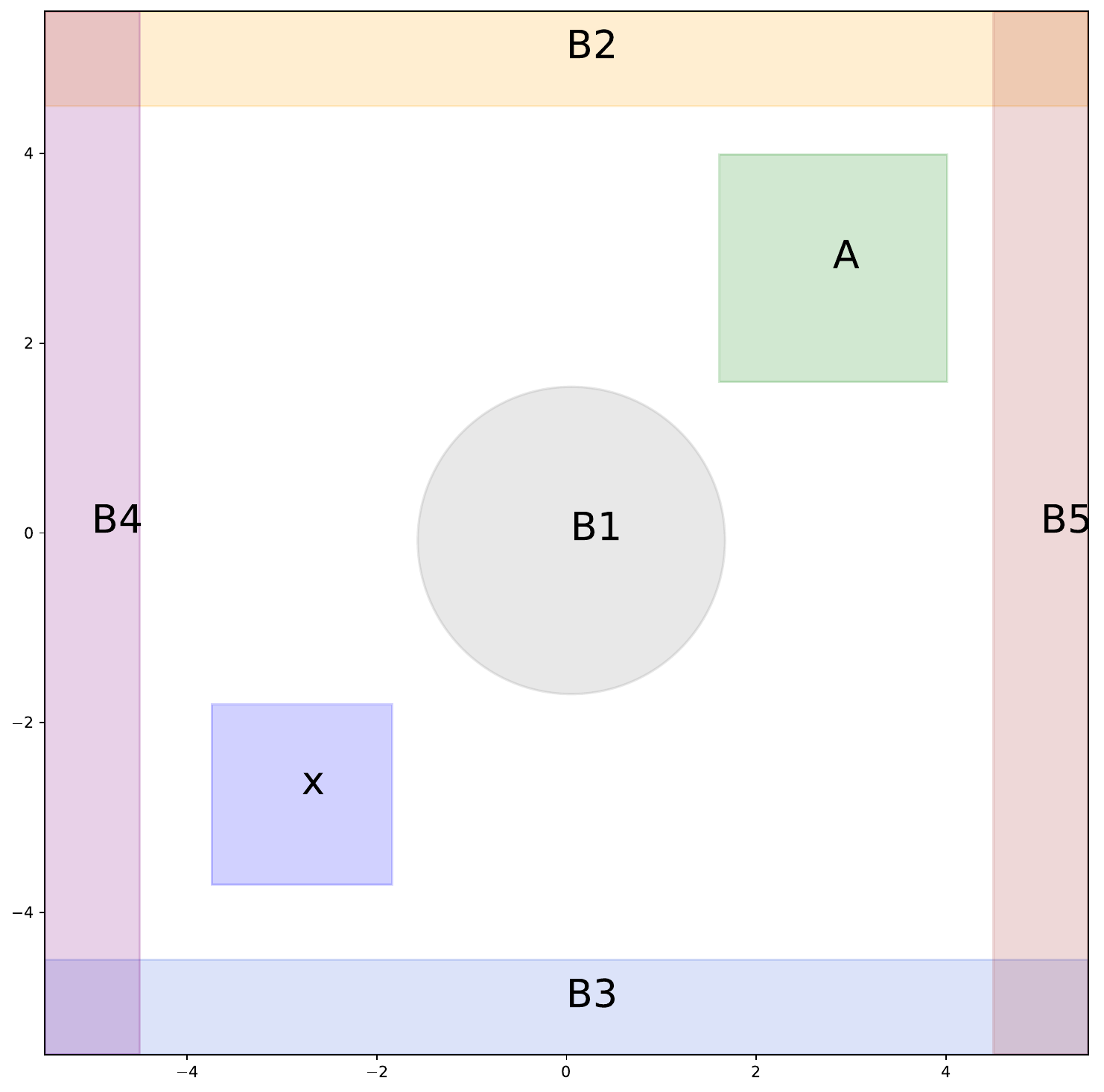}
      \caption{STL-06}
      \label{fig:scene-4ant-task-06}
  \end{subfigure}
  \begin{subfigure}[b]{0.19\textwidth}
      \centering \includegraphics[width=\textwidth]{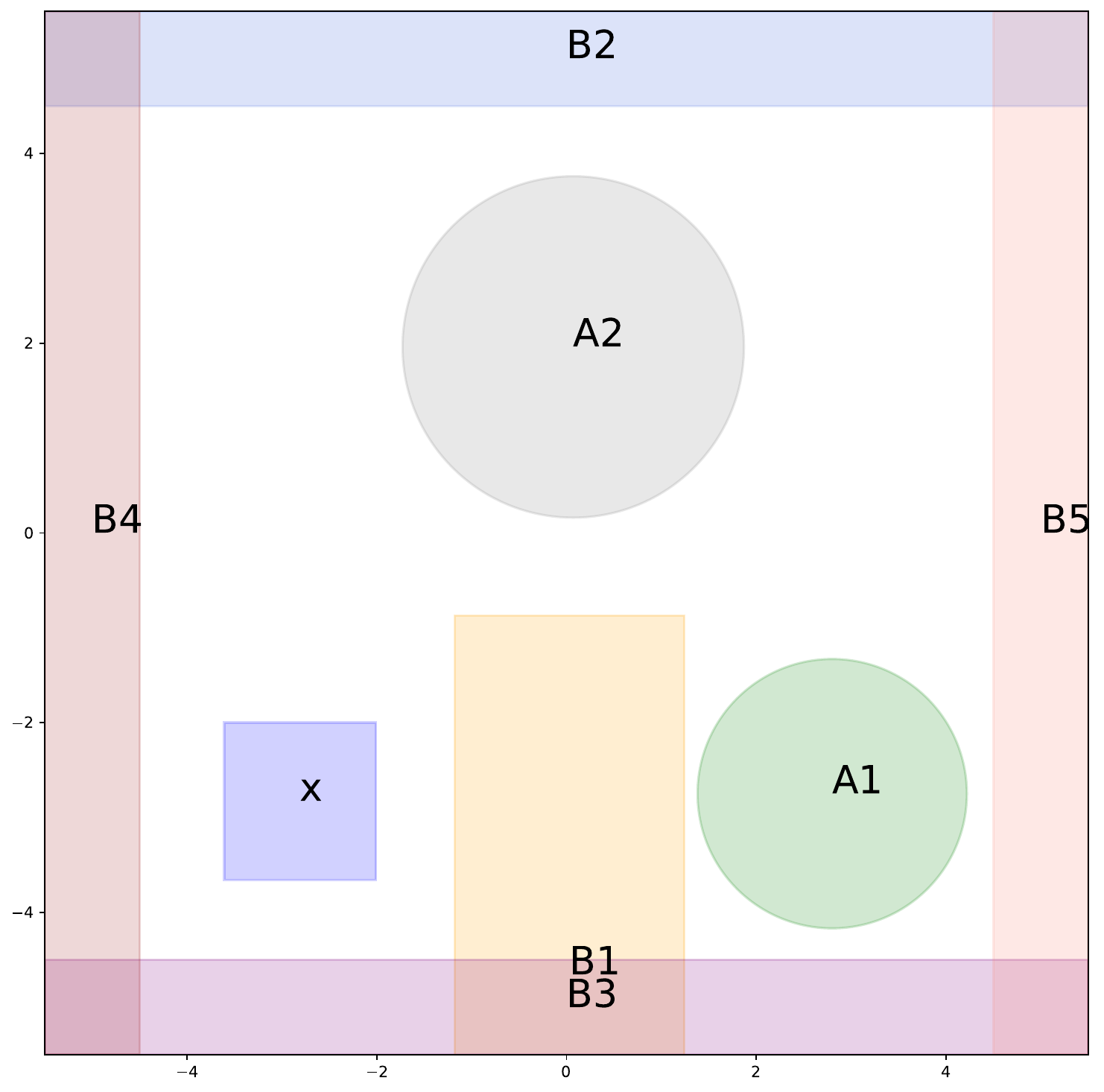}
      \caption{STL-07}
      \label{fig:scene-4ant-task-07}
  \end{subfigure}
  \begin{subfigure}[b]{0.19\textwidth}
      \centering \includegraphics[width=\textwidth]{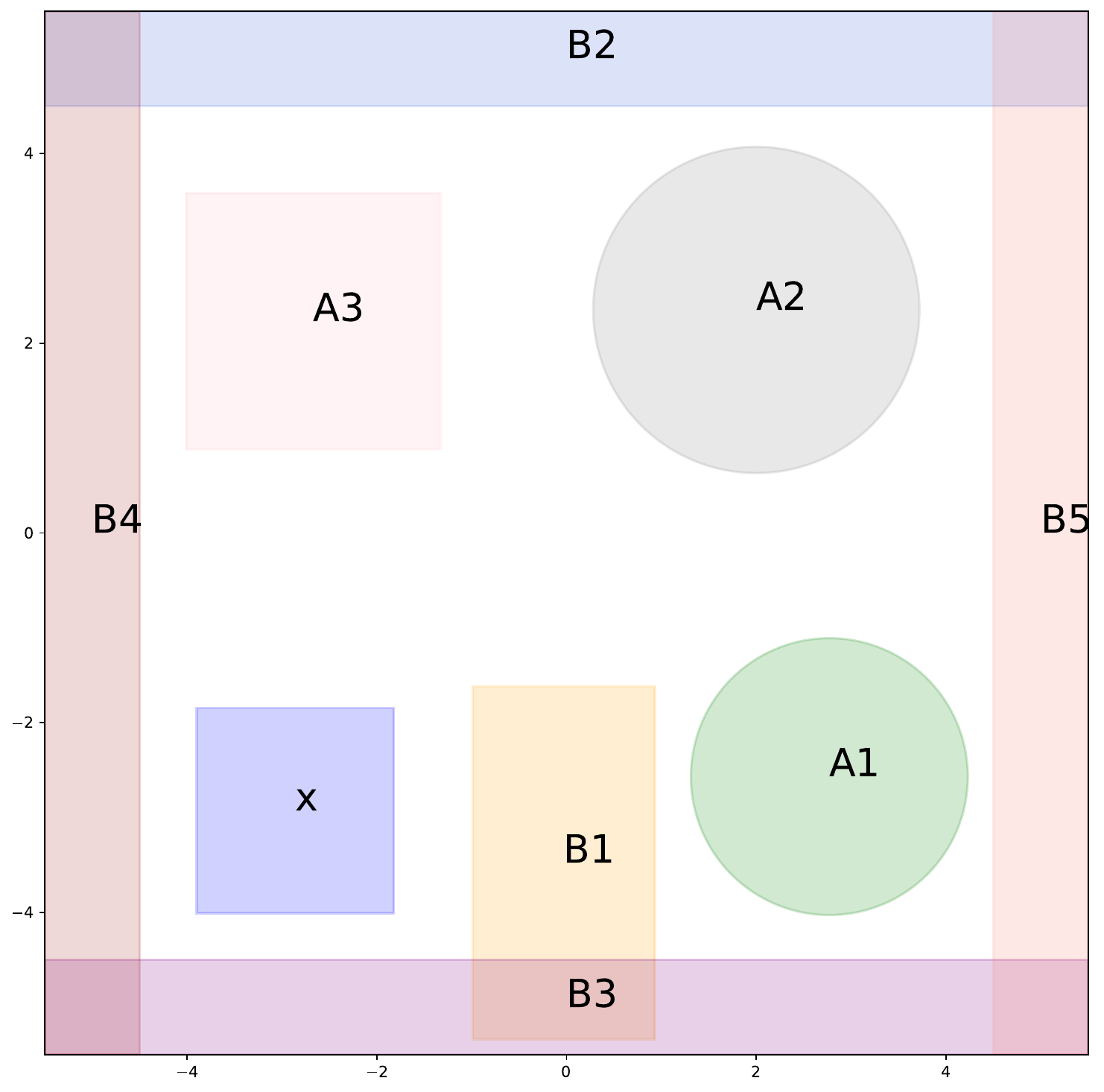}
      \caption{STL-08}
      \label{fig:scene-4ant-task-08}
  \end{subfigure}
  \begin{subfigure}[b]{0.19\textwidth}
      \centering \includegraphics[width=\textwidth]{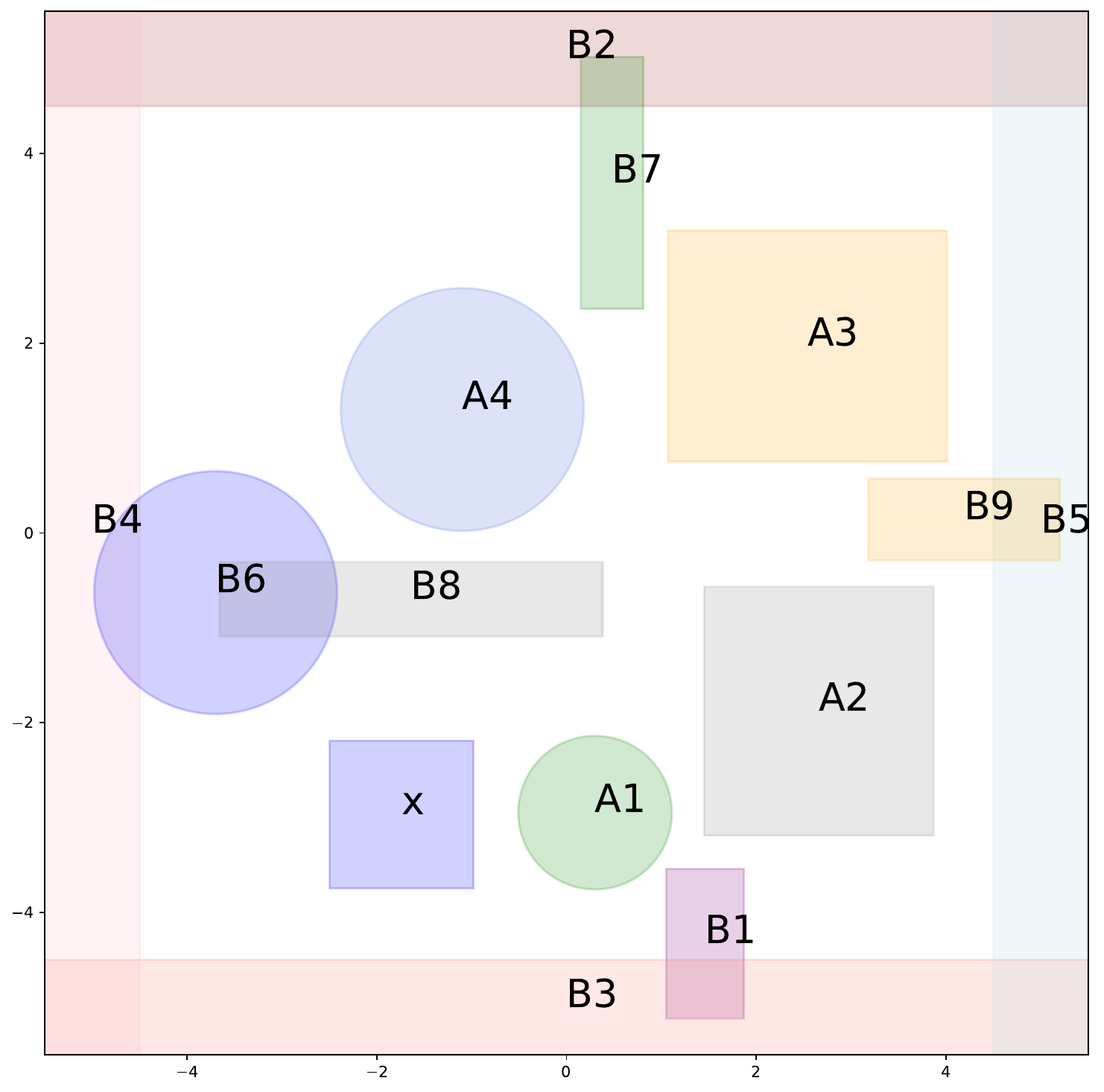}
      \caption{STL-09}
      \label{fig:scene-4ant-task-09}
  \end{subfigure}
  \begin{subfigure}[b]{0.19\textwidth}
      \centering \includegraphics[width=\textwidth]{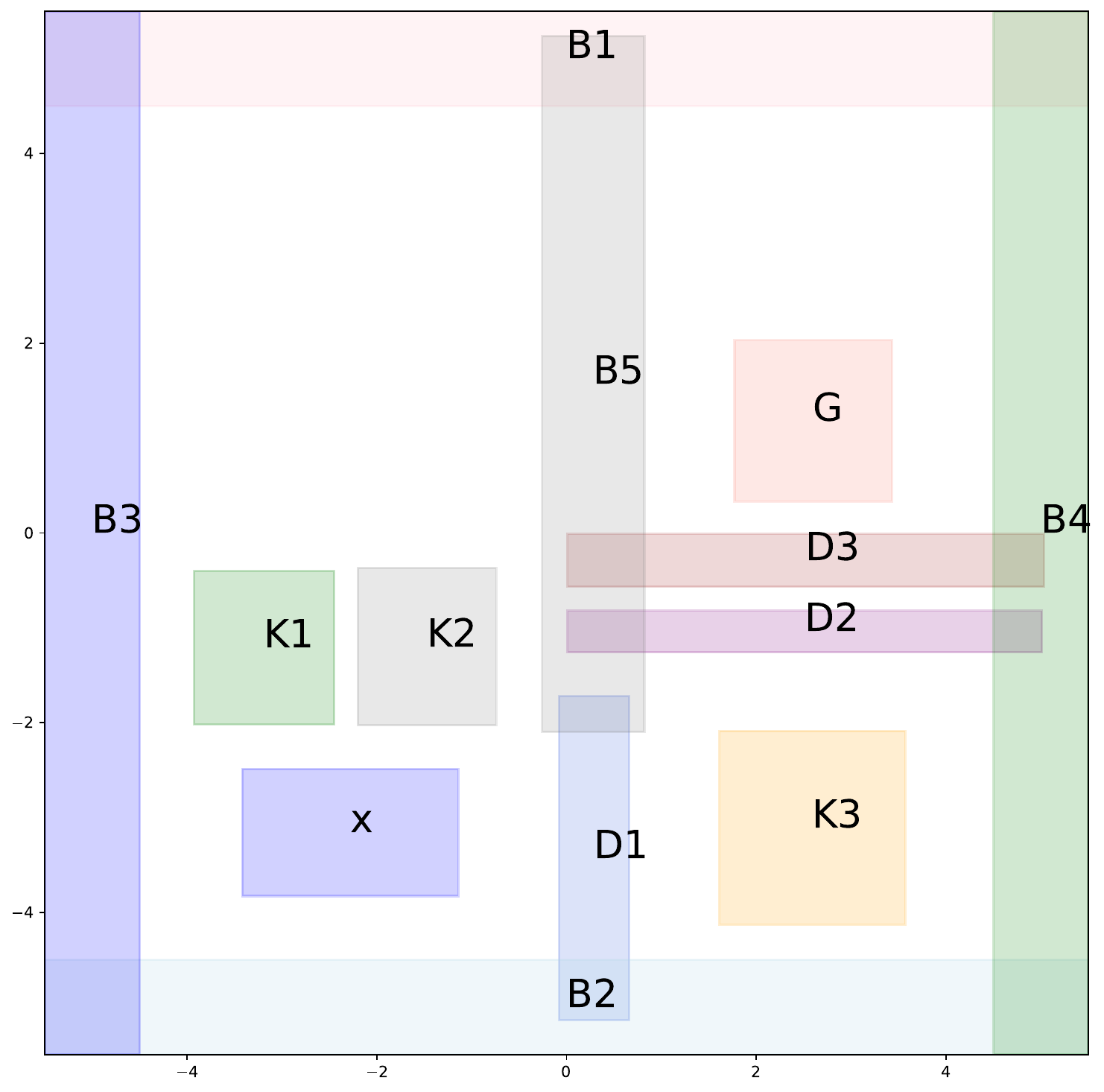}
      \caption{STL-10}
      \label{fig:scene-4ant-task-10}
  \end{subfigure}
  \caption{Scene for Ant: STL tasks 06 to 10}
  \label{fig:scene-cat-4ant-tasks-06-10}
\end{figure}
\noindent \textbf{STL-06 (Multi-layer):} \quad $  F_{[20:180]} (A) \land G_{[0:200]} (\neg B_1) \land G_{[0:200]} (\neg B_2) \land G_{[0:200]} (\neg B_3) \land G_{[0:200]} (\neg B_4) \land G_{[0:200]} (\neg B_5)  $

\noindent \textbf{STL-07 (Multi-layer):} \quad $  F_{[0:180]} (A_1) \land F_{[80:160]} (A_2) \land G_{[0:200]} (\neg B_1) \land G_{[0:200]} (\neg B_2) \land G_{[0:200]} (\neg B_3) \land G_{[0:200]} (\neg B_4) \land G_{[0:200]} (\neg B_5)  $

\noindent \textbf{STL-08 (Multi-layer):} \quad $  F_{[0:180]} (A_1) \land F_{[80:160]} ( A_2 \land F_{[20:40]} (G_{[0:20]} (A_3)) ) \land G_{[0:200]} (\neg B_1) \land G_{[0:200]} (\neg B_2) \land G_{[0:200]} (\neg B_3) \land G_{[0:200]} (\neg B_4) \land G_{[0:200]} (\neg B_5)  $

\noindent \textbf{STL-09 (Multi-layer):} \quad $  F_{[10:40]} ( A_1 \land F_{[20:40]} ( G_{[0:10]} (A_2) \land  F_{[20:60]} (G_{[0:10]} (A_3)) \land F_{[20:60]} (G_{[0:20]} (A_4))  ) ) \land G_{[0:200]} (\neg B_1) \land G_{[0:200]} (\neg B_2) \land G_{[0:200]} (\neg B_3) \land G_{[0:200]} (\neg B_4) \land G_{[0:200]} (\neg B_5) \land G_{[0:200]} (\neg B_6) \land G_{[0:200]} (\neg B_7) \land G_{[0:200]} (\neg B_8) \land G_{[0:200]} (\neg B_9)  $

\noindent \textbf{STL-10 (Multi-layer):} \quad $  (\neg D_1)U_{[0:200]}(K_1) \land (\neg D_2)U_{[0:200]}(K_2) \land (\neg D_3)U_{[0:200]}(K_3) \land F_{[160:180]} (G_{[0:10]} (G)) \land G_{[0:200]} (\neg B_1) \land G_{[0:200]} (\neg B_2) \land G_{[0:200]} (\neg B_3) \land G_{[0:200]} (\neg B_4) \land G_{[0:200]} (\neg B_5)  $

\end{document}